\documentclass{article}

\usepackage{times}
\usepackage{graphicx} 
\usepackage{url}            
\usepackage{booktabs}       
\usepackage{amsfonts}       
\usepackage{nicefrac}       
\usepackage{microtype}      
\usepackage{graphicx}
\usepackage{subfigure}
\usepackage{enumitem}
\usepackage{wrapfig}
\usepackage{booktabs} 
\usepackage{amsmath,amsfonts,amssymb,amsthm}
\usepackage[short]{optidef}
\newtheorem{proposition}{Proposition}

\newtheorem{definition}{Definition}
\newtheorem{lemma}{Lemma}

\newcommand{\ra}[1]{\renewcommand{\arraystretch}{#1}}
\usepackage{xcolor}

\usepackage{hyperref}

\usepackage[accepted]{icml2021}

\icmltitlerunning{Continual Learning of Generative Models with Limited Data}

\begin{document}

\twocolumn[
\icmltitle{Continual Learning of  Generative Models with Limited Data: From Wasserstein-1 Barycenter to Adaptive Coalescence}


\icmlsetsymbol{equal}{*}

\begin{icmlauthorlist}
\icmlauthor{Mehmet~Dedeoglu}{asu}
\icmlauthor{Sen~Lin}{asu}
\icmlauthor{Zhaofeng~Zhang}{asu}
\icmlauthor{Junshan~Zhang}{asu}
\end{icmlauthorlist}

\icmlaffiliation{asu}{School of Electrical, Computer and Energy Engineering, Arizona State University, Tempe, AZ, USA}

\icmlcorrespondingauthor{Mehmet~Dedeoglu}{mdedeogl@asu.edu}

\icmlkeywords{Continual Learning, Wasserstein Barycenter, Edge Learning, Wasserstein Generative Adversarial Networks}

\vskip 0.3in
]



\printAffiliationsAndNotice{}  

    
	
	\begin{abstract}
		Learning generative models is challenging for a network edge node with limited data and computing power. Since tasks in similar environments share model similarity, it is plausible to leverage pre-trained generative models from the cloud or other edge nodes. Appealing to optimal transport theory tailored towards Wasserstein-1 generative adversarial networks (WGAN), this study aims to develop a framework which systematically optimizes continual learning of  generative models  using local data at the edge node while exploiting adaptive coalescence of pre-trained generative models. Specifically, by treating the knowledge transfer from other nodes as Wasserstein balls centered around their pre-trained models,  continual  learning of generative models is cast as a constrained optimization problem, which is further reduced to a Wasserstein-1 barycenter problem. A two-stage approach is devised accordingly: 1) The barycenters among the pre-trained models are computed offline, where displacement interpolation is used as the theoretic foundation for finding adaptive barycenters via a ``recursive'' WGAN configuration; 2) the barycenter computed offline is used as meta-model initialization for continual learning and then fast adaptation is carried out to find the generative model using the local samples at the target edge node. Finally, a weight ternarization method, based on joint optimization of weights and threshold for quantization, is developed to compress the  generative model further. 
	\end{abstract}
	
	\section{Introduction}
	\label{intro}
	The past few years have witnessed an explosive growth of Internet of Things (IoT) devices at the network edge.
	On the grounds that the cloud has abundant computing resources, the conventional method for AI at the network edge is that the cloud trains the AI models with the data uploaded from edge devices, and then pushes the models back to the edge for on-device inference (e.g., Google Edge TPU).
	However, an emerging view is that this approach suffers from overwhelming communication overhead incurred by the data transmission from edge devices to the cloud, as well as potential privacy leakage.
	It is therefore of great interest to obtain generative models for the edge data, because they require a  smaller number of parameters than the data volume and it is much more parsimonious compared to sending the edge data to the cloud,
and further they can also help to preserve data privacy. Taking a forward-looking view, this study focuses on continual learning of generative models at edge nodes. 


	There are a variety of edge devices and edge servers, ranging from self-driving cars to robots, from 5G base station servers to mobile phones. Many edge AI applications (e.g., autonomous driving, smart robots, safety-critical health applications, and augmented/virtual reality) require edge intelligence and continual learning capability via fast adaptation with local data samples so as to adapt to dynamic application environments. Although deep generative models can parametrize high dimensional data samples at edge nodes effectively, it is often not feasible for a single edge server to train a deep generative model from scratch, which would otherwise require humongous training data and high computational power \cite{yonetani2019decentralized, wang2018transferring, wang2020transformation}. 
	A general consensus is that learning tasks across different edge nodes often share some model similarity. For instance, different robots may perform similar coordination behaviors according to the environment changes. With this sight, we advocate that the pre-trained generative models from other edge nodes are utilized to speed up the learning at a given edge node, and seek to answer the following critical questions:
	(1) {\em``What is the right abstraction of knowledge from multiple pre-trained models for continual learning?"} and (2) {\em``How can an edge server leverage this knowledge for continual learning of a generative model?"}
	
	The key to answering the first question lies in efficient model fusion of multiple pre-trained generative models. A common approach is the \emph{ensemble method} \cite{breiman1996bagging,schapire1999brief} where the outputs of different models are aggregated to improve the prediction performance. However, this requires the edge server to maintain all the pre-trained models and run each of them, which would outweigh the resources available at edge servers. Another way for model fusion is direct \emph{weight averaging} \cite{smith2017investigation, leontev2020non}. Because the weights in neural networks are highly redundant and no one-to-one correspondence exists between the weights of two different neural networks, this method is known to yield poor performance even if the networks represent the same function of the input. 
	As for the second question, Transfer Learning is a promising learning paradigm where an edge node incorporates the knowledge from the cloud or another node with its local training samples. \cite{wang2018transferring, wang2020transformation, yonetani2019decentralized}. 
	Notably, recent work on Transferring GANs \cite{wang2018transferring} proposed several transfer configurations to leverage  pre-trained GANs to accelerate the learning process. However, since  the transferred GAN is used only as initialization, Transferring GANs suffers from catastrophic forgetting. 

	
	\begin{figure}
		\centering
		\includegraphics[scale=0.35]{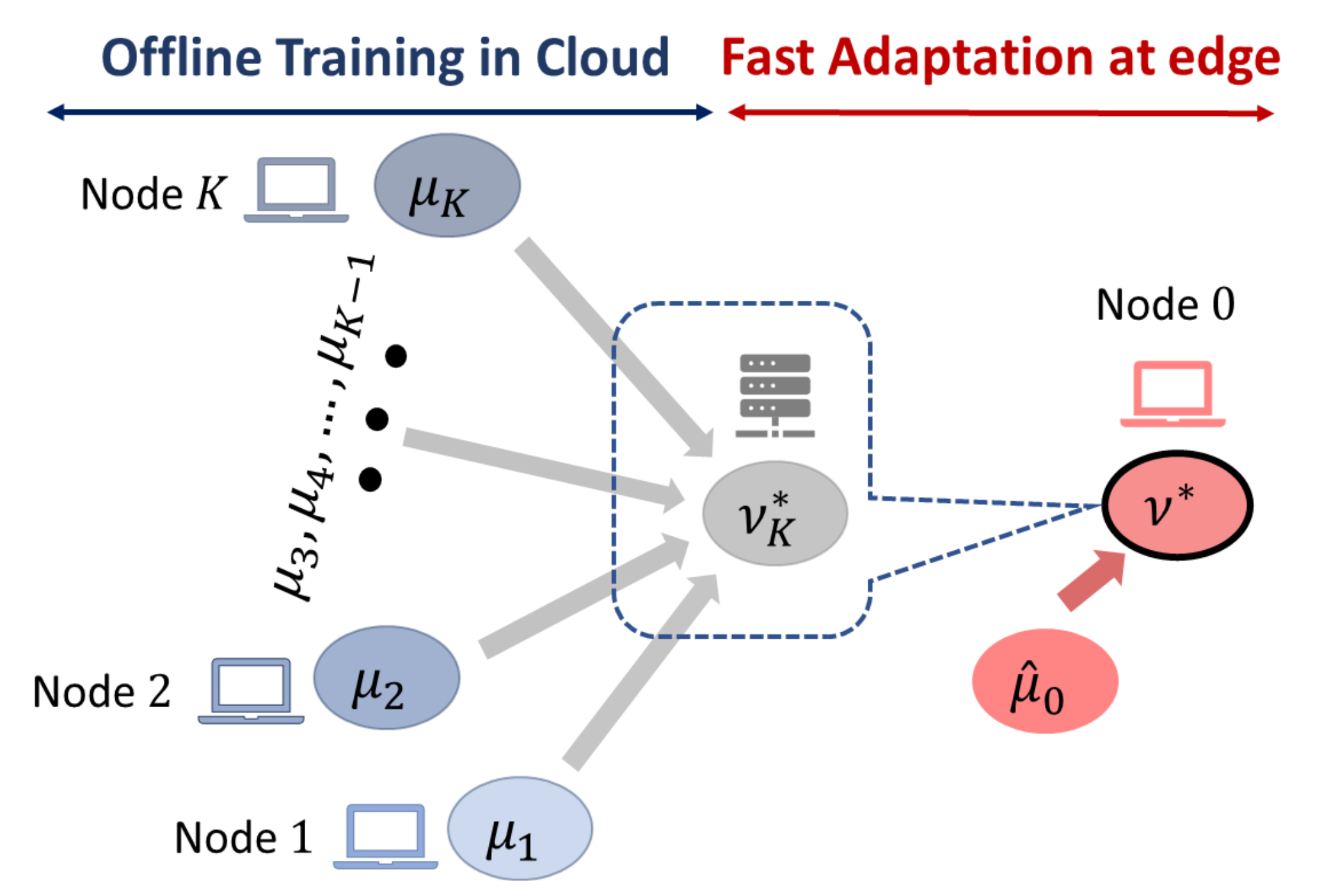}
		\caption{Continual learning of generative models based on  coalescence of pre-trained generative models \(\{\mu_k, k=1, \ldots, K\}\) and local dataset at Node 0 (denoted by \(\hat{\mu}_0\)).}
		\label{f1_a}
	\end{figure} 
	To tackle these challenges, this work aims to develop a framework which explicitly optimizes the continual learning of generative models for the edge, based on the adaptive coalescence of pre-trained generative models from other edge nodes, using optimal transport theory tailored towards GANs. 
 To mitigate the mode collapse problem due to the vanishing gradients, multiple GAN configurations have been proposed  based on the Wasserstein-\(p\) metric \(W_p\), including Wasserstein-1 distance \cite{wgan} and Wasserstein-2 distance \cite{w2gan1,w2gan2}. 
	Despite Wasserstein-2 GANs are analytically tractable, the corresponding implementation often requires regularization and is often outperformed by the Wasserstein-1 GAN (W1GAN). With this insight, in this paper we focus on the W1GAN (WGAN refers to W1GAN throughout).

Specifically, we consider a setting where an edge node, denoted Node 0, aims to learn a generative model. It has been shown that training a WGAN is intimately related to finding a distribution  minimizing the Wasserstein distance from the underlying distribution \(\mu_0\) \cite{arora2017generalization}.  
	In practice, an edge node has only a limited number of samples with empirical distribution \(\hat{\mu}_0\), which is distant from \(\mu_0\). A naive approach is to train a WGAN based on the limited local samples only, which can be captured via the optimization problem given by 
\(	\min\nolimits_{\nu\in\mathcal{P}}~W_1(\nu,\hat{\mu}_{0}) \),
with \(W_1(\cdot,\cdot)\) being the Wasserstein-\(1\) distance between two distributions. The best possible outcome of solving this optimization problem can generate a distribution very close to \(\hat{\mu}_0\), which however could still be far away from the true distribution \({\mu}_0\). Clearly, training a WGAN simply based on the limited local samples at an edge node  would not work well.

	As alluded to earlier, learning tasks across different edge nodes may share model similarity.
	To facilitate the continual learning at Node 0, pre-trained generative models from other related edge nodes can be leveraged via knowledge transfer. Without loss of generality, we assume that there are a set \(\mathcal{K}\) of \(K\) edge nodes with pre-trained generative models. Since one of the most appealing benefits of WGANs is the ability to continuously estimate the Wasserstein distance during training \cite{wgan}, we assume that the
	knowledge transfer from Node \(k\) to Node 0 is in the form of a Wasserstein ball with radius \(\eta_k\) centered around its pre-trained generative model \(\mu_k\) at Node \(k\), for \(k=1, \ldots, K\). 
	Intuitively, radius \(\eta_k\) represents the relevance (hence utility) of the knowledge transfer, and the smaller it is, the more informative the corresponding Wasserstein ball is. 
	Building on this knowledge transfer model, we treat the continual learning problem at Node 0 as the coalescence of $K$ generative models and empirical distribution \(\hat{\mu}_0\) (Figure \ref{f1_a}), and cast it as the following constrained optimization problem:
\begin{equation}\label{op:constrained}
\underset{\nu\in\mathcal{P}}{\min}~W_1(\nu,\hat{\mu}_{0}), ~\text{s.t.}~~W_1(\nu,\mu_k)\leq\eta_k, \forall k\in\mathcal{K}.
\end{equation}
	Observe that the constraints in problem \eqref{op:constrained} dictate that the optimal coalesced generative model,  denoted by \(\nu^*\), lies within the intersection of \(K\) Wasserstein balls (centered around \(\{\mu_k \}\)), exploiting the knowledge transfer systematically. It is worth noting that the optimization problem \eqref{op:constrained} can be extended to other distance functionals, e.g., Jensen-Shannon divergence.
	
	
	The contributions of this work are summarized as follows.
	
	1) We propose a systematic framework to enable continual learning of generative models via adaptive coalescence of pre-trained generative models from other edge nodes and local samples at Node 0. In particular, by treating the knowledge transferred from each node as a Wasserstein ball centered around its local pre-trained generative model, we cast the problem as a constrained optimization problem which optimizes the continual learning of generative models.
	
	2) Applying Lagrangian relaxation to \eqref{op:constrained}, we reduce the optimization problem to finding a Wasserstein-1 barycenter of \(K+1\) probability measures, among which \(K\) of them are pre-trained generative models and the last one is the empirical distribution (not a generative model though) corresponding to local data samples at Node 0. We propose a {\em barycentric fast adaptation approach} to efficiently solve the barycenter problem, where the barycenter \(\nu_K^*\) for the \(K\) pre-trained generative models is found recursively offline in the cloud, and then the barycenter between the empirical distribution \(\hat{\mu}_0\) of Node 0 and \(\nu_K^*\) is solved via fast adaptation at Node 0. A salient feature in this proposed barycentric  approach is that generative replay, enabled by pre-trained GANs, is used to annihilate catastrophic forgetting.
	
	3) It is known that the Wasserstein-1 barycenter is notoriously difficult to analyze, partly because of the existence of infinitely many minimizers of the Monge Problem. Appealing to optimal transport theory, we  use displacement interpolation as the theoretic foundation to devise recursive algorithms for finding adaptive barycenters, which ensures the resulting  barycenters lie in the baryregion.
	
	4) From the implementation perspective, we introduce a ``recursive'' WGAN configuration, where a  2-discriminator WGAN is used per recursive step to find adaptive barycenters sequentially. 
 Then the resulting barycenter in offline training is treated as the meta-model initialization and fast adaptation is carried out to find the generative model using the local samples at Node 0. A weight ternarization method, based on joint optimization of weights and threshold for quantization, is developed to compress the generative model and enable efficient edge learning. Extensive experiments corroborate the efficacy of the proposed framework for fast edge learning of generative models.

\begin{figure}
		\centering
		\includegraphics[scale=0.6]{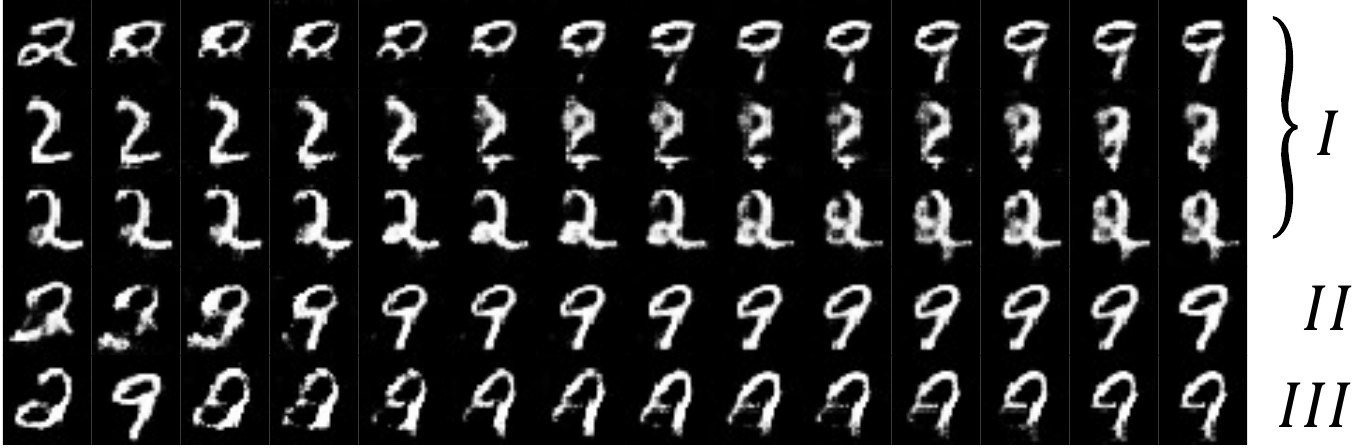}
		\vspace*{-0.25in}
		\caption{The illustrations of image morphing using 3 different techniques: \(I\)) Barycentric fast adaptation, \(II\)) Transferring GANs and \(III\)) Ensemble method. \(5000\) samples from classes ``2'' and ``9'' in MNIST are used in experiments and horizontal axis represents training iterations.}
		\label{fig_morphing}
\end{figure}
	
The proposed barycentric fast adaptation approach is useful for many applications, including  image morphing \cite{simon2020barycenters}, clustering \cite{cuturi2014}, super resolution \cite{ledig2017photorealistic} and privacy-aware synthetic data generation \cite{shrivastava2017learning} at  edge nodes. To get a more concrete sense,
Figure \ref{fig_morphing} illustrates a comparison of image morphing  using three methods, namely \emph{barycentric fast adaptation}, \emph{Transferring GANs} and \emph{ensemble}. Observe that Transferring GANs quickly morphs images from class ``2'' to class ``9'', but forgetting the previous knowledge. In contrast, barycentric fast adaptation morphs class ``2''  to a barycenter model between the two classes ``2'' and ``9,'' because it uses generative replay in the training (we will elaborate further on this in the WGAN configuration),  thus mitigating catastrophic forgetting. 
The ensemble method learns both classes ``2'' and ``9'' at the end, but its morphing process takes longer. 
	
\section{Related Work}

Optimal transport theory has recently been studied for deep learning applications (see, e.g., \cite{Brenier,ambrosio2008,villani2008}). \cite{agueh2011} has developed an analytical solution to the Wasserstein barycenter problem. Aiming to solve the Wasserstein barycenter problem, \cite{cuturi2013,cuturi2014, cuturi2016} proposed smoothing through entropy regularization for the discrete setting, based on linear programming. \cite{cevher2015} employed posterior sampling algorithms in studying Wasserstein barycenters, and \cite{anderes2016} characterized Wasserstein barycenters for the discrete setting (cf. \ \cite{staib2017, ye2017, singh2019model}). 
GANs \cite{goodfellow2014} have recently emerged as a powerful deep learning tool for obtaining  generative models. Recent work \cite{wgan} has introduced Wasserstein metric in GANs, which can help mitigate the vanishing gradient issue to avoid mode collapse. Though gradient clipping is applied to ensure \(1\)-Lipschitz conditions, it may still lead to non-convergence. \cite{wgangp} proposed to use gradient penalty to overcome the shortcomings due to weight clipping. Using optimal transport theory, recent advances of Wasserstein GANs have shed light on understanding generative models.  Recent works \cite{w2gan1,w2gan2} proposed two distinct transport theory based GANs using 2-Wasserstein distance. Furthermore, \cite{Gu} devised an computationally efficient method for computing the generator when the cost function is convex. In contrast, for the Wasserstein-1  GAN, the corresponding discriminator may constitute one of infinitely many optimal maps from underlying empirical data distribution to the generative model \cite{ambrosio2008,villani2008}, and it remains open to decipher the relation between the model training and the optimal transport maps. Along a different line, a variety of techniques have been proposed for more robust training of GANs \cite{qi2019ke, yonetani2019decentralized, durugkar2016, simon2020barycenters}. 

Pushing the AI frontier to the network edge  for achieving  edge intelligence has recently emerged as the marriage of AI and edge computing \cite{zhou2019edge}. Yet, the field of edge intelligence is still in its infancy stage and there are significant challenges since  AI model training generally requires tremendous resources that greatly outweigh the capability of resource-limited edge nodes. To address this, various approaches have been proposed in the literature, including model compression \cite{shafiee2017squishednets,yang2017designing,wang2019private}, knowledge transfer learning \cite{osia2020hybrid,wang2018not}, hardware acceleration \cite{venkataramani2017scaledeep,wang2017energy}, collaboration-based methods \cite{lin2020collaborative,zhang2020data}, etc. Different from these existing studies, {\em this work focuses on  continual learning of generative models  at the edge node}. Rather than learning the new model from scratch, continual learning aims to design algorithms leveraging knowledge transfer from pre-trained models to the new learning task \cite{thrun1995lifelong}, assuming that the
training data of previous tasks are  unavailable for the newly coming task.  Clearly, continual learning fits naturally in edge learning applications. Notably, the elastic weight consolidation  method \cite{kirkpatrick2017overcoming, zenke2017continual} estimates importance of all neural network parameters and encodes it into the Fisher information matrix, and changes of important parameters are penalized during the training of later tasks.  Generative replay is gaining more attention where synthetic samples corresponding to earlier tasks are obtained with a generative model and replayed in model training for the  new task to mitigate forgetting \cite{rolnick2019experience,rebuffi2017icarl}.
In this work, by learning generative models via the adaptive coalescence of pre-trained generative models from other nodes, the proposed ``recursive" WGAN configuration   facilitates fast edge learning in a continual manner, which can be viewed as an innovative integration of a few key ideas in continual learning, including  the replay method \cite{shin2017continual,wu2018memory,ostapenko2019learning,riemer2019scalable} which generates pseudo-samples using generative models, and the regularization-based methods \cite{kirkpatrick2017overcoming,lee2017overcoming,schwarz2018progress,dhar2019learning} which sets the regularization for the model learning based on the learned knowledge from previous tasks, in continual learning \cite{de2019continual}.

	\section{Adaptive Coalescence of Generative Models: A Wasserstein-1 Barycenter Approach}

	In what follows, we first recast problem \eqref{op:constrained} as a variant of the Wasserstein barycenter problem. Then, we propose a two-stage recursive algorithm, characterize the geometric properties of geodesic curves therein and use displacement interpolation as the foundation to devise recursive algorithms for finding adaptive barycenters.
	

	\subsection{A Wasserstein-1 barycenter formulation via Lagrangian relaxation}
Observe that the Lagrangian for \eqref{op:constrained} is given as follows:\vspace{-0.08in}
\begin{small}\begin{equation}\label{q8_e}
	\mathcal{L}(\{\lambda_k\}, \nu) = W_1(\nu,\hat{\mu}_{0})+\sum\limits_{k=1}^{K} \lambda_kW_1(\nu,\mu_k) -\sum\limits_{k=1}^{K}\lambda_k\eta_k,\vspace{-0.08in}
\end{equation}\end{small}
where \(\{\lambda_k\geq 0\}_{1:K}\) are the Lagrange multipliers.  Based on \cite{volpi2018generalizing}, problem \eqref{op:constrained} can be solved by using the following Lagrangian relaxation with \(\lambda_k= \frac{1}{\eta_k}, \forall k\in\mathcal{K}\), and \(\lambda_{0}=1\):\vspace{-0.08in}
\begin{equation}\label{q8}
	\underset{\nu\in\mathcal{P}}{\min}~\sum\limits_{k=1}^{K} \frac{1}{\eta_k} W_1(\nu,\mu_k)+ W_1(\nu,\hat{\mu}_{0}).\vspace{-0.08in}
\end{equation}
It is shown in \cite{sinha2017certifying}  that the selection \(\lambda_k= \frac{1}{\eta_k}, \forall k\in\mathcal{K}\) ensures the same levels of robustness for \eqref{q8} and \eqref{op:constrained}. Intuitively, such a selection of \(\{\lambda_k \}_{0:K}\) strikes a right balance, in the sense that larger weights are assigned to the knowledge transfer models (based on the pre-trained generative models \(\{\mu_k\}\)) from the nodes with higher relevance, captured by smaller Wasserstein-1 ball radii.  For given \(\{\lambda_k\geq 0\}\), \eqref{q8} turns out to be a Wasserstein-1 barycenter problem (cf.\ \cite{agueh2011,cevher2015}), with the new complication that \(\hat{\mu}_0\) is an empirical distribution corresponding to local samples at Node 0. Since \(\hat{\mu}_0\) is not a generative model per se, its coalescence with other $K$ general models is challenging.

\subsection{A Two-stage adaptive coalescence approach for Wasserstein-1 barycenter problem}
Based on (\ref{q8}), we take a two-stage approach to enable efficient learning of the generative model at edge Node 0. The primary objective of Stage I is to find the barycenter for \(K\) pre-trained generative models \(\{\mu_1, \ldots, \mu_K \}\). Clearly, the ensemble method would not work well due to required memory and computational resources. With this insight, we develop a recursive algorithm for adaptive coalescence of pre-trained generative models. In Stage II, the resulting barycenter solution in Stage I is treated as the model initialization, and is further trained using the local samples at Node 0.
We propose that the offline training in Stage I is asynchronously performed 
in the cloud, and the fast adaptation in Stage II is carried out at the edge server (in the same spirit as the model update of Google EDGE TPU), as outlined below:
	
\textit{Stage I: Find the barycenter of \(K\) pre-trained generative models across \(K\) edge nodes offline.} 
Mathematically, this entails the solution of the following problem:\vspace{-0.08in}
\begin{align}
\min_{\nu\in\mathcal{P}}~\sum_{k=1}^{K}\frac{1}{\eta_k}W_1(\nu,\mu_k).
\label{op:offline}
\end{align}
To reduce computational complexity, we propose the following recursive algorithm: Take $\mu_1$ as an initial point, i.e., \(\nu^*_1=\mu_1\), and let \(\nu^*_{k-1}\) denote the barycenter of \(\{\mu_i\}_{1:k-1}\) obtained at iteration \(k-1\) for \(k=2, \ldots, K\). Then, at each iteration \(k\), a new barycenter \(\nu^*_k\) is solved between the barycenter \(\nu^*_{k-1}\) and the pre-trained generative model \(\mu_k\). (Details are in Algorithm 1 in the appendix.)
	
\textit{Stage II: Fast adaptation to find the barycenter between \(\nu_K^*\) and the local dataset at Node 0.} Given the solution \(\nu_K^*\) obtained in Stage I, we subsequently solve the following problem:
\(\min_{\nu\in\mathcal{P}}~W_1(\nu,\hat{\mu}_{0})+W_1(\nu,\nu_K^*).
\label{op:fastadaptation}\)   
By taking \(\nu_K^*\) as the model initialization, fast adaptation based on local samples is used to learn the generative model at Node 0. (See Algorithm 2 in the appendix.)
	
\subsection{From Displacement Interpolation to Adaptive Barycenters}

	
As noted above, in practical implementation, the W1-GAN often outperforms Wasserstein-\(p\) GANs (\(p>1\)). However,  the Wasserstein-1 barycenter is notoriously difficult to analyze due to the non-uniqueness of the minimizer to the Monge Problem \cite{villani2008}. Appealing to optimal transport theory, we next characterize the performance of the proposed two-stage recursive algorithm for finding the Wasserstein-1 barycenter of pre-trained generative models \(\{\mu_k, k=1, \ldots, K\}\) and the local dataset at Node 0, by examining the existence of the barycenter and characterizing its geometric properties based on geodesic curves.
	
The seminal work \cite{mccann1997convexity} has established the existence of geodesic curves between any two distribution functions \(\sigma_0\) and \(\sigma_1\) in the \(p\)-Wasserstein space, \(\mathcal{P}_p\), for \(p\geq 1\). It is shown in \cite{villani2008} that there are infinitely many minimal geodesic curves between  \(\sigma_0\) and \(\sigma_1\), when \(p=1\). This is best illustrated in \(N\) dimensional Cartesian space, where the minimal geodesic curves between \(\varsigma_0\in \mathbb{R}^N\) and \(\varsigma_1\in \mathbb{R}^N\) can be parametrized as follows: \(\varsigma_t = \varsigma_0+s(t)(\varsigma_1-\varsigma_0), \label{GC10} \)
where \(s(t)\) is an arbitrary function of \(t\), indicating that there are infinitely many minimal geodesic curves between  \(\varsigma_0\) and \(\varsigma_1\). This is in stark contrast to the case \(p>1\) where there is a unique geodesic between \(\varsigma_0\) and \(\varsigma_1\). In a similar fashion, there exists infinitely many transport maps, \(T^1_0\), from \(\sigma_0\) to \(\sigma_1\) when \(p= 1\). For convenience, let \(C(\sigma_0,\sigma_1)\)
denote an appropriate transport cost function quantifying the minimum cost to move a unit mass from \(\sigma_0\) to \(\sigma_1\). 
It has been shown in \cite{villani2008} that when \(p=1\), two interpolated distribution functions on two distinct minimal curves may have a non-zero distance, i.e., \(C(\hat{T}^1_0\#\sigma_0,\tilde{T}^1_0\#\sigma_0)\geq 0\), where \(\#\) denotes the push-forward operator, thus yielding multiple minimizers to \eqref{op:offline}. For convenience, define \(\mathcal{F}:=\hat{\mu}_0\cup\{\mu_k\}_{1:K}\).

\begin{definition}(\textbf{Baryregion})\label{baryregion}
	Let \(g_t(\mu_k,\mu_\ell)_{0\leq t\leq 1}\) denote any minimal geodesic curve between any pair \(\mu_k,\mu_\ell\in\mathcal{F}\), and define the union \(\mathcal{R}:=\bigcup_{k=1}^{K}\bigcup_{\ell=k+1}^{K+1} g_t(\mu_k,\mu_\ell)_{0\leq t\leq 1}\). Then, the baryregion \(\mathcal{B}_\mathcal{R}\) is given by
	\(\mathcal{B}_\mathcal{R}=\bigcup_{\sigma\in\mathcal{R}}\bigcup_{\varpi\in\mathcal{R},\varpi\not=\sigma} g_t(\sigma,\varpi)_{0\leq t\leq 1}\). 		
\end{definition}

Intuitively, \(\mathcal{B}_\mathcal{R}\) encapsulates all possible interpolations through  distinct geodesics between any two distributions in \(\mathcal{R}\) or \(\mathcal{F}\). Since each geodesic has finite length, \(\mathcal{B}_\mathcal{R}\) defines a bounded set in \(\mathcal{P}_1\).
Next we restate in Lemma \ref{thm2_} the renowned \emph{Displacement Interpolation} result \cite{mccann1997convexity}, which sets the foundation for each recursive step in finding a barycenter in our proposed two-stage algorithm. In particular, Lemma \ref{thm2_} leads to the fact that the barycenter \(\nu^*\)  resides in \(\mathcal{B}_\mathcal{R}\).
	
\begin{lemma}(\textbf{Displacement Interpolation, \cite{villani2003}})\label{thm2_}
Let \(C(\sigma_0, \sigma_1)\) denote the minimum transport cost between \(\sigma_0\) and \(\sigma_1\), and suppose \(C(\sigma_0,\sigma_1)\) is finite for \(\sigma_0,\sigma_1\in\mathcal{P}(\mathcal{X})\). Assume that \(C(\sigma_s,\sigma_t)\), the minimum transport cost between \(\sigma_s\) and \(\sigma_t\) for any \(0\leq s\leq t\leq 1\), is continuous. Then, the following holds true for any given  continuous path \(g_t(\sigma_0,\sigma_1)_{0\leq t\leq 1}\):
\small
\begin{equation}\label{eqgeo}
	C(\sigma_{t_1},\sigma_{t_2})\hspace{-0.03in}+\hspace{-0.03in}C(\sigma_{t_2},\sigma_{t_3}) \hspace{-0.03in}=\hspace{-0.03in}C(\sigma_{t_1},\sigma_{t_3}),~~ 0\hspace{-0.03in}\leq\hspace{-0.03in} t_1\hspace{-0.03in}\leq\hspace{-0.03in} t_2\hspace{-0.03in}\leq\hspace{-0.03in} t_3\hspace{-0.03in}\leq\hspace{-0.03in} 1.
\end{equation}
\normalsize
\end{lemma}

In the adaptive coalescence algorithm, the \(k\)th recursion defines a baryregion, \(\mathcal{B}_{\{\nu_{k_1}^*,\mu_k\}}\), consisting of geodesics between the barycenter \(\nu^*_{k-1}\) found in \((k-1)\)th recursion and generative model \(\mu_k\). Clearly,  \(\mathcal{B}_{\{\nu_k^*,\mu_k\}}\subset\mathcal{B}_\mathcal{R}\). Viewing each recursive step in the above two-stage algorithm as adaptive displacement interpolation, we have the following main result on  the  geodesics and the geometric properties regarding \(\nu^*\)  and \(\{\nu_k^*\}_{1:K}\).  
\begin{proposition}\label{existence}
		
(\textbf{Displacement interpolation for adaptive barycenters}) The adaptive barycenter, \(\nu_k^*\), obtained at the output of \(k\)th recursive step in Stage I, is a displacement interpolation between \(\nu_{k-1}^*\) and \(\mu_k\) and resides inside \(\mathcal{B}_\mathcal{R}\). Further, the final barycenter \(\nu^*\) resulting from Stage II of the recursive algorithm resides inside  \(\mathcal{B}_\mathcal{R}\).
\end{proposition}

\section{Recursive WGAN Configuration for Adaptive Coalescence and Continual Learning}

Based on the above theoretic results on adaptive coalescence via Wasserstein-1 barycenters, we next turn attention to the implementation of computing adaptive barycenters. Notably, assuming the knowledge of accurate empirical distribution models on discrete support, \cite{cuturi2014} introduces a powerful linear program (LP) to compute Wasserstein-\(p\) barycenters, but the computational complexity of this approach is excessive. In light of this, we propose a WGAN-based configuration for finding the Wasserstein-1 barycenter, which in turn enables fast learning of generative models based on the coalescence of pre-trained models. Specifically, \eqref{q8} can be rewritten as:
\begin{align}
	\label{cost_q8}
	\underset{G}{\min} &\underset{
		\{\varphi_k\}_{0:K} }{\max}  \left\{\mathbb{E}_{{x}\sim\hat{\mu}_0}[\varphi_0({x})] -\mathbb{E}_{{z}\sim\vartheta}[\varphi_0(G({z}))]\right\}\nonumber\\			&+\sum_{k=1}^{K}\frac{1}{\eta_k}\left\{\mathbb{E}_{{x}\sim\mu_k}[\varphi_k({x})] -\mathbb{E}_{{z}\sim\vartheta}[\varphi_k(G({z}))]\right\}, 
\end{align}
where \(G\)  represents the generator and \(\{\varphi_k\}_{0:K}\) are $1-$Lipschitz functions for discriminator models, respectively. Observe that the optimal generator DNN \(G^*\) facilitates the barycenter distribution \(\nu^*\) at its output. We note that the multi-discriminator WGAN configuration have recently been developed \cite{durugkar2016,hardy2019,behnam}, by using a common latent space to train multiple discriminators so as to improve stability. In stark contrast, in this work distinct generative models from multiple nodes are exploited to train different discriminators, aiming to learn distinct transport plans among generative models.
\begin{figure}
	\centering
	\hspace{0\columnwidth}
	\includegraphics[width=0.98\columnwidth]{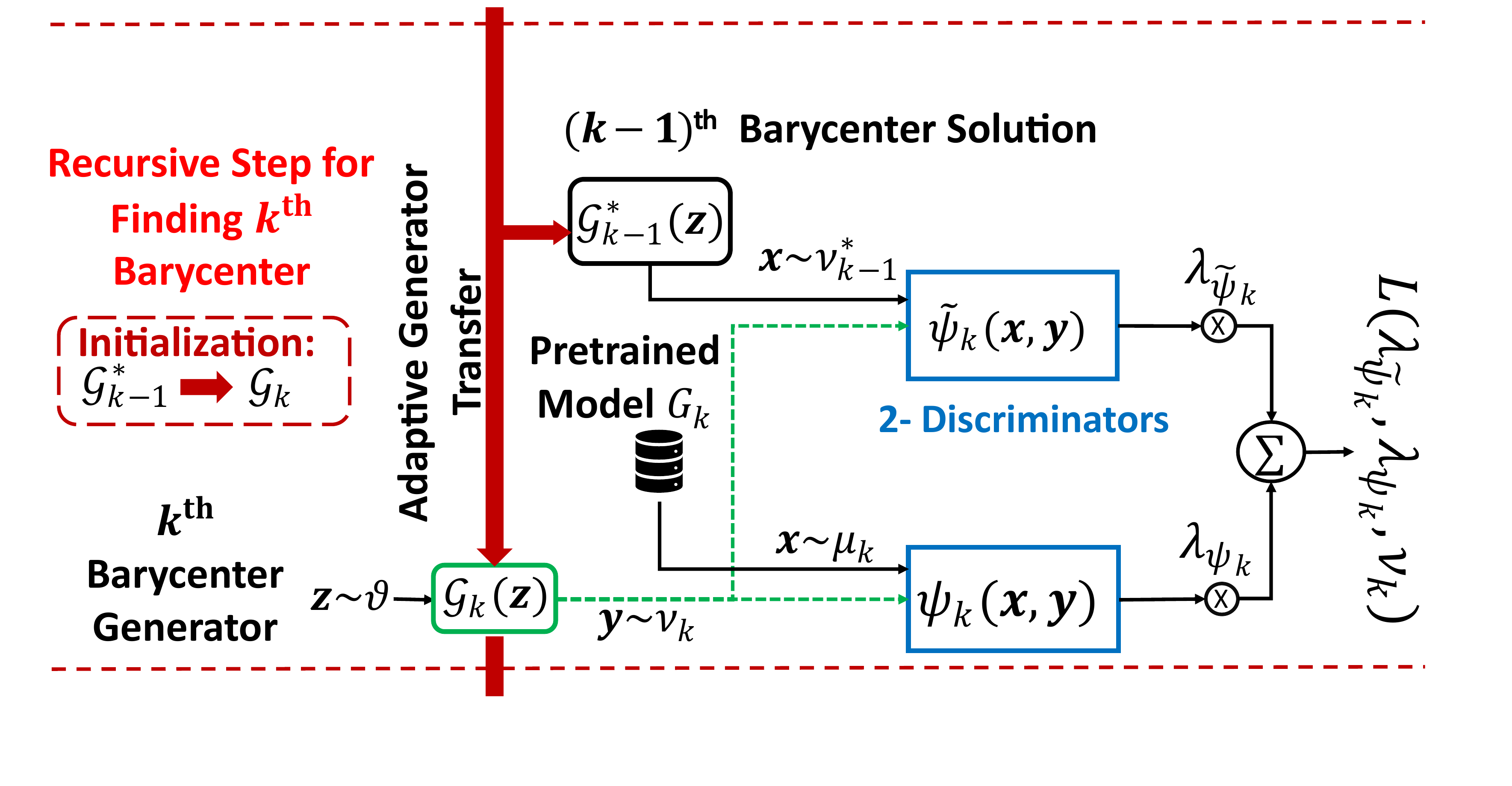}\label{fig3a}
	\caption{A 2-discriminator WGAN for efficient learning of \(k\)th barycenter generator in offline training, where \({x}\) denotes the synthetic data generated from pretrained models.}
	\label{fig3} 
\end{figure}

\begin{figure}
	\centering
		\includegraphics[width=0.95\columnwidth]{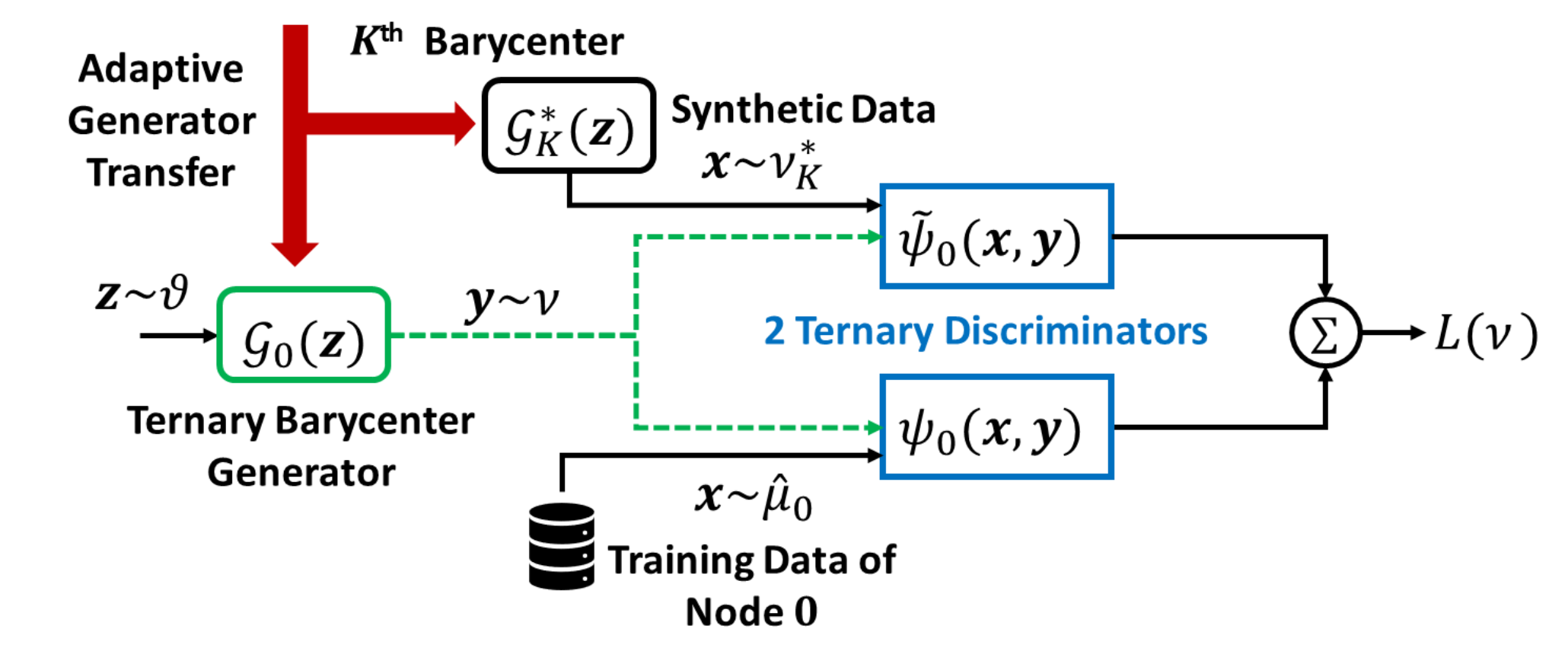}
		\caption{Fast adaptation for learning  generative model at Node 0.}
		\label{f3}
\end{figure}

A naive approach is to implement the above multi-discriminator WGAN in a one-shot manner where the generator and \(K+1\) discriminators are trained simultaneously, which however would require overwhelming computation power and memory. To enable efficient training, we use the proposed two-stage algorithm and develop a ``recursive'' WGAN configuration to sequentially compute 1) the barycenter \(\nu_K^*\) for the offline training in the cloud, as shown in Figure \ref{fig3}; and 2) the barycenter \(\nu^*\) for the fast adaptation at the target edge node, as shown in Figure \ref{f3}. The analytical relation between \emph{one-shot} and \emph{recursive} barycenters has been studied for Wasserstein-2 distance, and sufficient conditions for their equivalence is presented in \cite{boissard2015distribution}, which, would not suffice for Wasserstein-1 distance, because of the existence of multiple Wasserstein-1 barycenters. Proposition \ref{existence} shows that any barycenter solution to recursive algorithm resides inside a baryregion, which can be viewed as the counterpart for the \emph{one-shot} solution. We highlight a few important advantages of the ``recursive'' WGAN configuration for the barycentric fast adaptation algorithm.

1) \emph{A 2-discriminator WGAN implementation per recursive step to enable efficient training.} At each recursive step \(k\), we aim to find the barycenter \(\nu_k^*\) between pre-trained model \(\mu_k\) and the barycenter \(\nu_{k-1}^*\) from last round, which is achieved by training a 2-discriminator WGAN as follows:
\begin{align}\label{wcost}
	\underset{\mathcal{G}_k}{\min}&\underset{\psi_k, \tilde{\psi}_k}{\max} \lambda_{\psi_k}\big\{\mathbb{E}_{{x}\sim\mu_k}[\psi_k({x})]- \mathbb{E}_{{z}\sim\vartheta}[{\psi}_k(\mathcal{G}_k({z}))]\big\}\nonumber\\
	&+\lambda_{\tilde{\psi}_k}\big\{\mathbb{E}_{{x}\sim\nu_{k-1}^*}[\tilde{\psi}_k({x})]- \mathbb{E}_{{z}\sim\vartheta}[\tilde{\psi}_k(\mathcal{G}_k({z}))]\big\},
\end{align}
where \(\psi\) and \(\tilde{\psi}\) denote the corresponding discriminators for pre-trained model \(G_k\) and barycenter model \(\mathcal{G}_{k-1}^*\) from the previous recursive step, respectively.

2) \emph{Model initialization in each recursive step.} For the initialization of the generator \(\mathcal{G}_k\), we use the trained generator \(\mathcal{G}_{k-1}^*\) in last step. \(\mathcal{G}_{k-1}^*\) corresponds to the barycenter \(\nu_{k-1}^*\), and using it as the initialization the displacement interpolation would move along the  geodesic curve from \(\nu_{k-1}^*\) to \(\mu_k\) \cite{w2gan1}. It has been shown  that training GANs with such initializations would accelerate the convergence  compared with training from scratch \cite{wang2018transferring}. Finally, \(\nu_K^*\) is adopted as initialization to enable fast  adaptation at the target edge node. With the barycenter \(\nu_K^*\) solved via offline training, a new barycenter \(\nu^*\) between local dataset (represented by \(\hat{\mu}_0\)) and \(\nu_K^*\),  can be obtained by training a  2-discriminator WGAN, and fine-tuning the generator \(\mathcal{G}_{0}\) from \(\mathcal{G}_{K}^*\) would be notably \emph{faster and more accurate} than learning the generative model from local data only.

3) \emph{Fast adaptation for training ternary WGAN  at Node 0.} As outlined in Algorithm 2, fast adaptation is used to find the barycenter between \(\nu_K^*\) and the local dataset at Node 0. To further enhance  edge learning, we adopt the weight ternarization method to compress the WGAN model during training. The weight ternarization method not only replaces computationally-expensive multiplication operations with efficient addition/subtraction operations, but also enables the sparsity in model parameters \cite{han2015deep}. Specifically, the ternarization process is formulated as:
\begin{small}
\begin{equation}\label{ternary}
	w_{l}^{\prime}\hspace{-0.03in}=\hspace{-0.03in}S_{l} \cdot Tern\left(w_{l}, \Delta_{l}^{\pm}\right)=S_{l} \cdot\left\{\begin{array}{ll}+1 & w_{l}>\Delta_{l}^{+} \\ 0 & \Delta_{l}^{-} \leq w_{l} \leq \Delta_{l}^{+} \\ -1 & w_{l}<\Delta_{l}^{-}\end{array}\right.
\end{equation}
\end{small}
where \(\{w_{l} \}\) are the full precision weights for \(l\)th layer, \(\{w_{l}^{\prime}\}\) are the weight after ternarization, \(\{S_{l}\}\) is the layer-wise weight scaling coefficient and \(\Delta_{l}^{\pm}\) are the layer-wise thresholds. Since the fixed weight thresholds may lead to accuracy degradation, \(S_{l}\) is approximated as a differentiable closed-form function of \(\Delta_{l}^{\pm}\) so that both weights and thresholds can be optimized simultaneously through back-propagation \cite{he2019simultaneously}. Let the generator and the discriminators of WGAN at Node 0 be denoted by \(\mathcal{G}_0\), $\tilde{\psi}_0$ and \(\psi_0\), which are parametrized by the ternarized weights \(\{\mathbf{w}^\prime_{l_\mathcal{G}}\}_{l_\mathcal{G}=1}^{L_\mathcal{G}}\), \(\{\mathbf{w}^\prime_{l_{\tilde{\psi}}}\}_{l_{\tilde{\psi}}=1}^{L_{\tilde{\psi}}}\) and \(\{\mathbf{w}^\prime_{l_\psi}\}_{l_\psi=1}^{L_\psi}\), respectively. The barycenter \(\nu^*\) at Node 0, captured by \(\mathcal{G}_0^*\), can be obtained by training the ternary WGAN via  iterative updates of both weights and thresholds:
\begin{align}\label{ternarycost}
	\underset{\mathcal{G}_0}{\min}\underset{\psi_0, \tilde{\psi}_0}{\max} &\mathbb{E}_{{x}\sim\hat{\mu}_0}[\psi_0({x})]- \mathbb{E}_{{z}\sim\vartheta}[{\psi}_0(\mathcal{G}_0({z}))]\nonumber\\
	&+\mathbb{E}_{{x}\sim\nu_{K}^*}[\tilde{\psi}_0({x})]- \mathbb{E}_{{z}\sim\vartheta}[\tilde{\psi}_0(\mathcal{G}_0({z}))],
\end{align}
which takes three steps in each iteration: a) calculating the scaling coefficients and the ternary weights for  \(\mathcal{G}_0\), $\tilde{\psi}_0$ and \(\psi_0\), b) calculating the loss function using the ternary weights via forward-propagation and c) updating the full precision weights and the thresholds via back-propagation.

\section{Experiments}	

\textbf{Datasets, Models and Evaluation.} We extensively examine the performance of learning a generative model, using the barycentric fast adaptation algorithm, on a variety of widely adapted datasets in the GAN literature, including CIFAR10, CIFAR100, LSUN and MNIST. In experiments, we used various DCGAN-based architectures \cite{radford2015unsupervised} depending on the dataset as different datasets vary in image size, feature diversity and in sample size, e.g., image samples in MNIST has less diversity compared to the rest of the datasets, while LSUN contains the largest number of samples with larger image sizes. Further, we used the weight ternarization method \cite{he2019simultaneously} to jointly optimize weights and quantizers of the generative model at the target edge node, reducing the memory burden of generative models in memory-limited edge devices. Details on the characteristics of datasets and network architectures used in experiments are relegated to the appendix. 

The Frechet-Inception Distance (FID) score \cite{fid} is used for evaluating the performance of the two-stage adaptive coalescence algorithm and all baseline algorithms. The FID score is widely adopted for evaluating the performance of GAN models in the literature \cite{chong2019fid, wang2018transferring, fid2}, since it provides a quantitative assessment of the similarity of a dataset to another reference dataset. In all experiments, we use the entire dataset as the reference dataset. We here emphasize that a smaller FID score of a GAN indicates that it has better performance. A more comprehensive discussion of FID score is relegated to the appendix.

To demonstrate the improvements by using the proposed framework based on \emph{barycentric fast adaptation}, we conduct extensive experiments and compare performance with \(3\) distinct baselines: 1) \emph{transferring GANs} \cite{wang2018transferring}: a  pre-trained GAN model is used as initialization at Node 0 for training a new WGAN model by using local data samples. 2) \emph{Ensemble method}: The model initialization,  obtained by using pre-trained GANs at other edge nodes,  is further trained using both local data from Node 0 and synthetic data samples. 3) \emph{Edge-Only}: only local dataset  at node 0 is used in WGAN training. 


Following \cite{heusel2017gans, wang2018transferring}, we use the FID score to quantify the image quality. Due to the lack of sample diversity  at the target edge node, the   WGAN model trained using local data only is not expected to attain a small FID score. In stark contrast,  the WGAN model  trained using the proposed two-stage adaptive coalescence algorithm, inherits diversity from pre-trained models at other edge nodes, and can result in  lower FID scores than its counterparts. We note that if the entire dataset were available  at Node \(0\), then the minimum FID score would be achieved (see the appendix).

\begin{figure*}[t]
    \centering
		\hspace{0\columnwidth}
		\subfigure[Comparison of convergence over MNIST: The non-overlapping case. ]{\includegraphics[width=0.67\columnwidth]{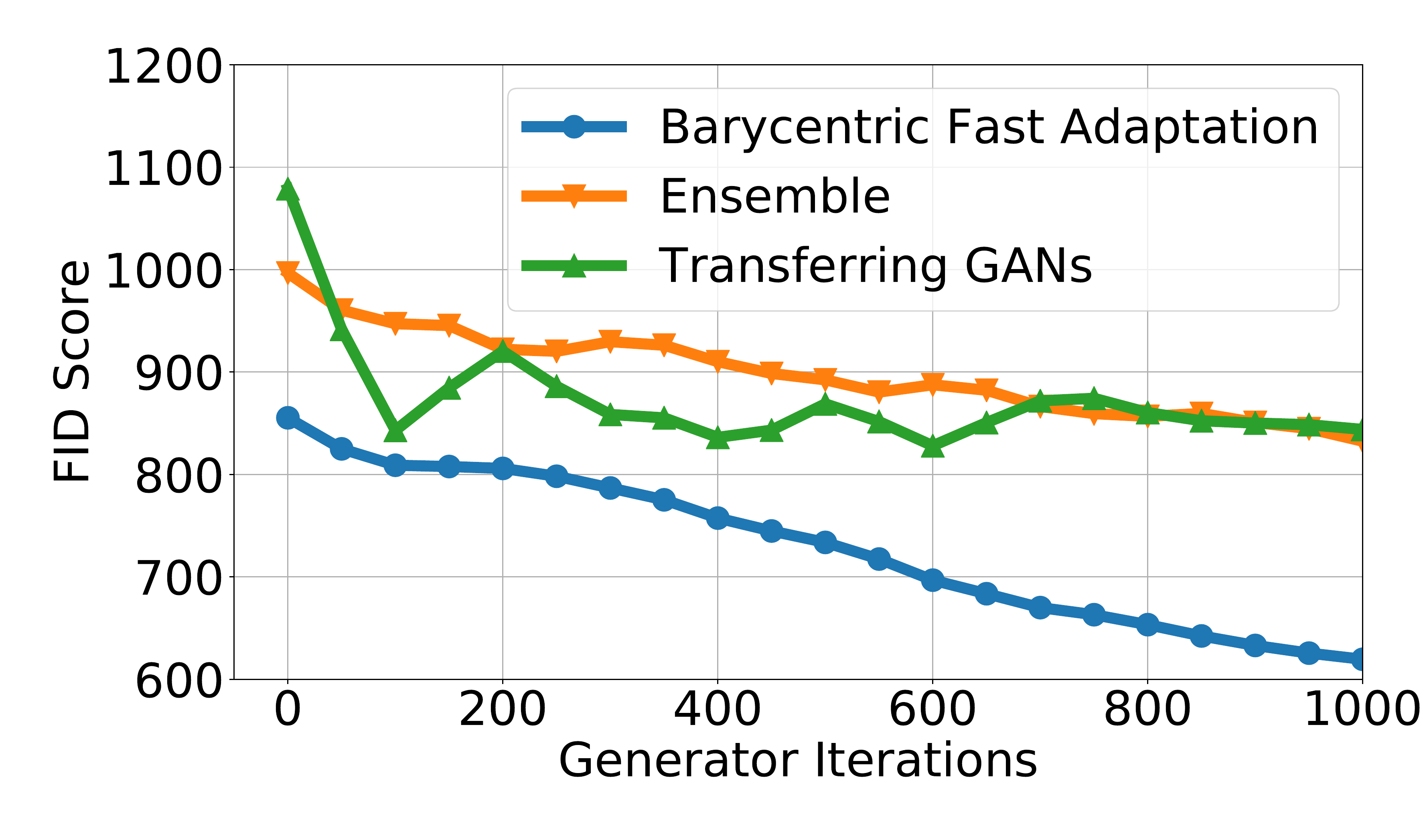}\label{sample4}}
		\hspace{0.01\columnwidth}
		\subfigure[Comparison of convergence over MNIST: The overlapping case. ]{\includegraphics[width=0.66\columnwidth]{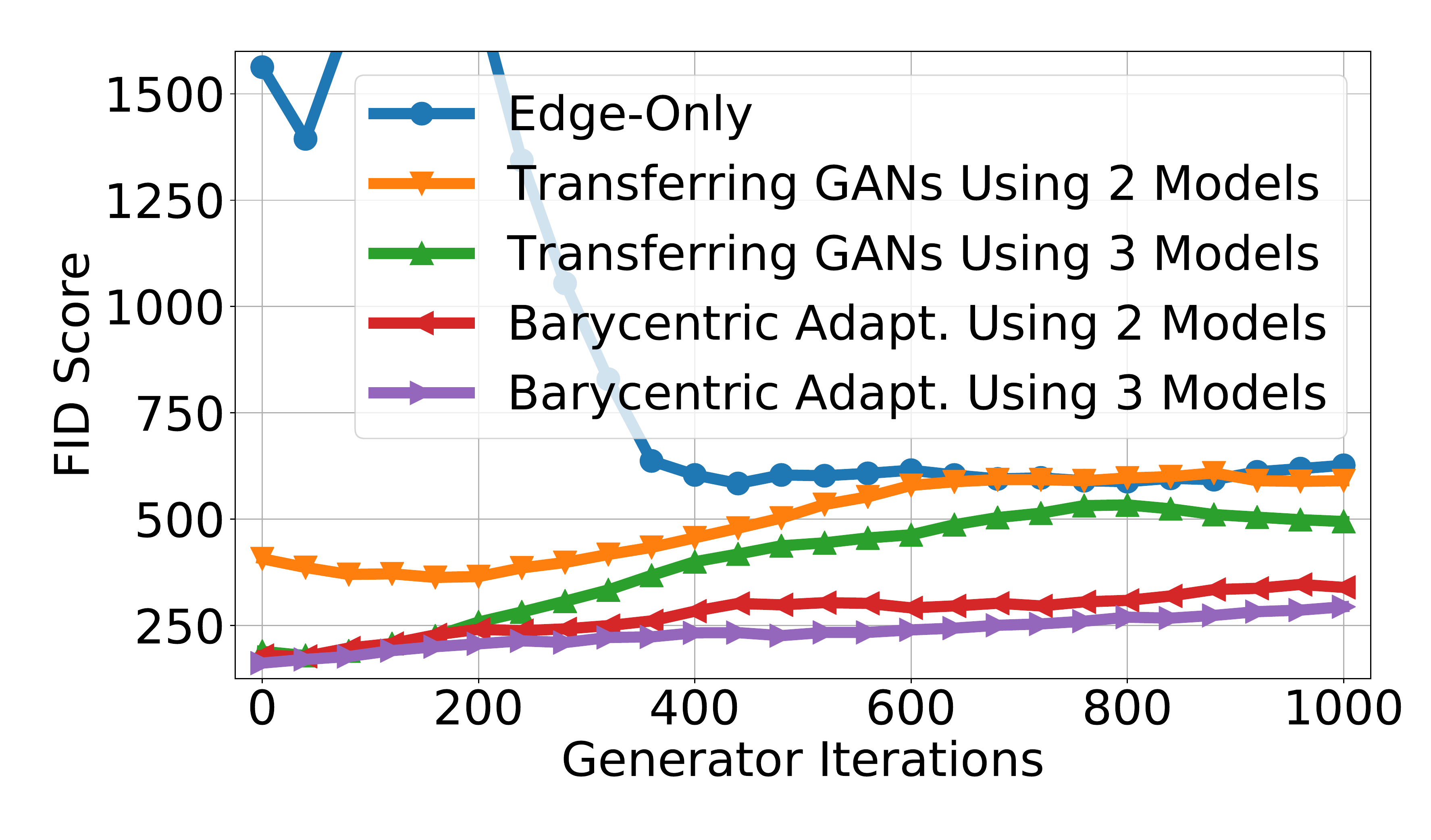}\label{trvsbaMNIST}}%
		\hspace{0.01\columnwidth}
		\subfigure[Comparison of convergence over CIFAR10: The overlapping case. ]{\includegraphics[width=0.67\columnwidth]{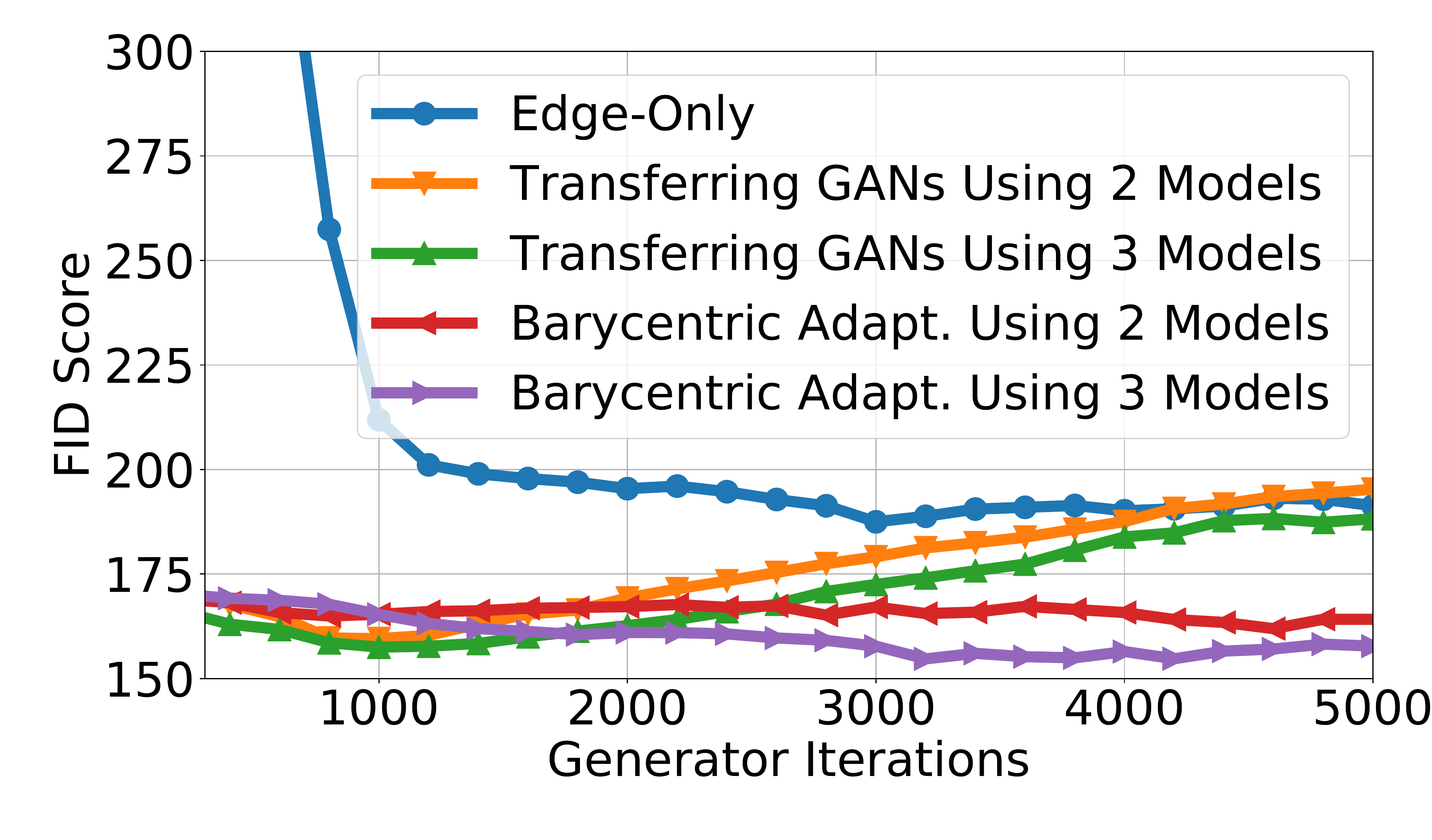}\label{trvsbaCIFAR}}%
		\hspace{0.01\columnwidth}
		\subfigure[Comparison of convergence over CIFAR100: The overlapping case. ]{\includegraphics[width=0.67\columnwidth]{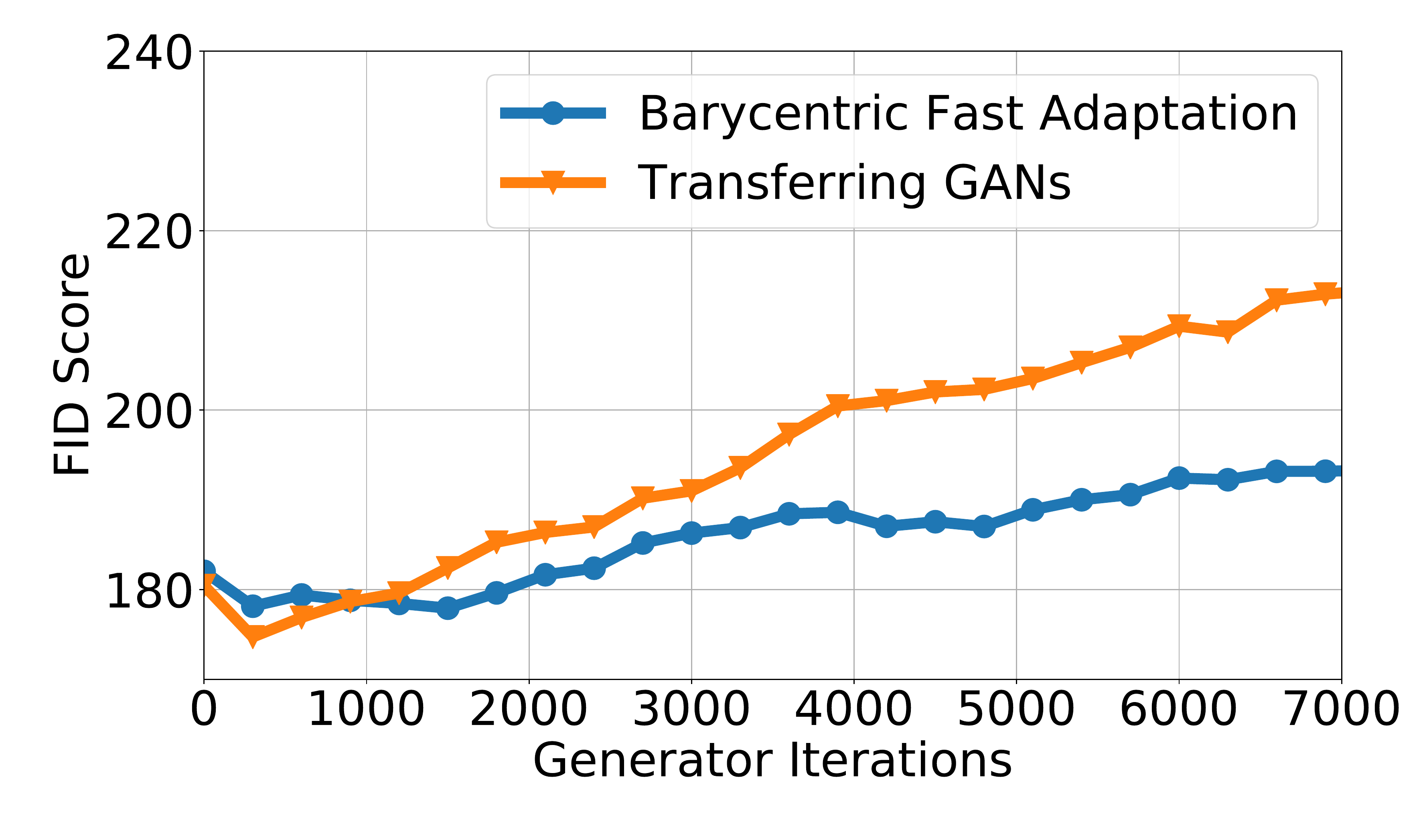}\label{samp2}}%
		\hspace{0.01\columnwidth}
		\subfigure[Comparison of convergence over CIFAR10: The overlapping case. ]{\includegraphics[width=0.67\columnwidth]{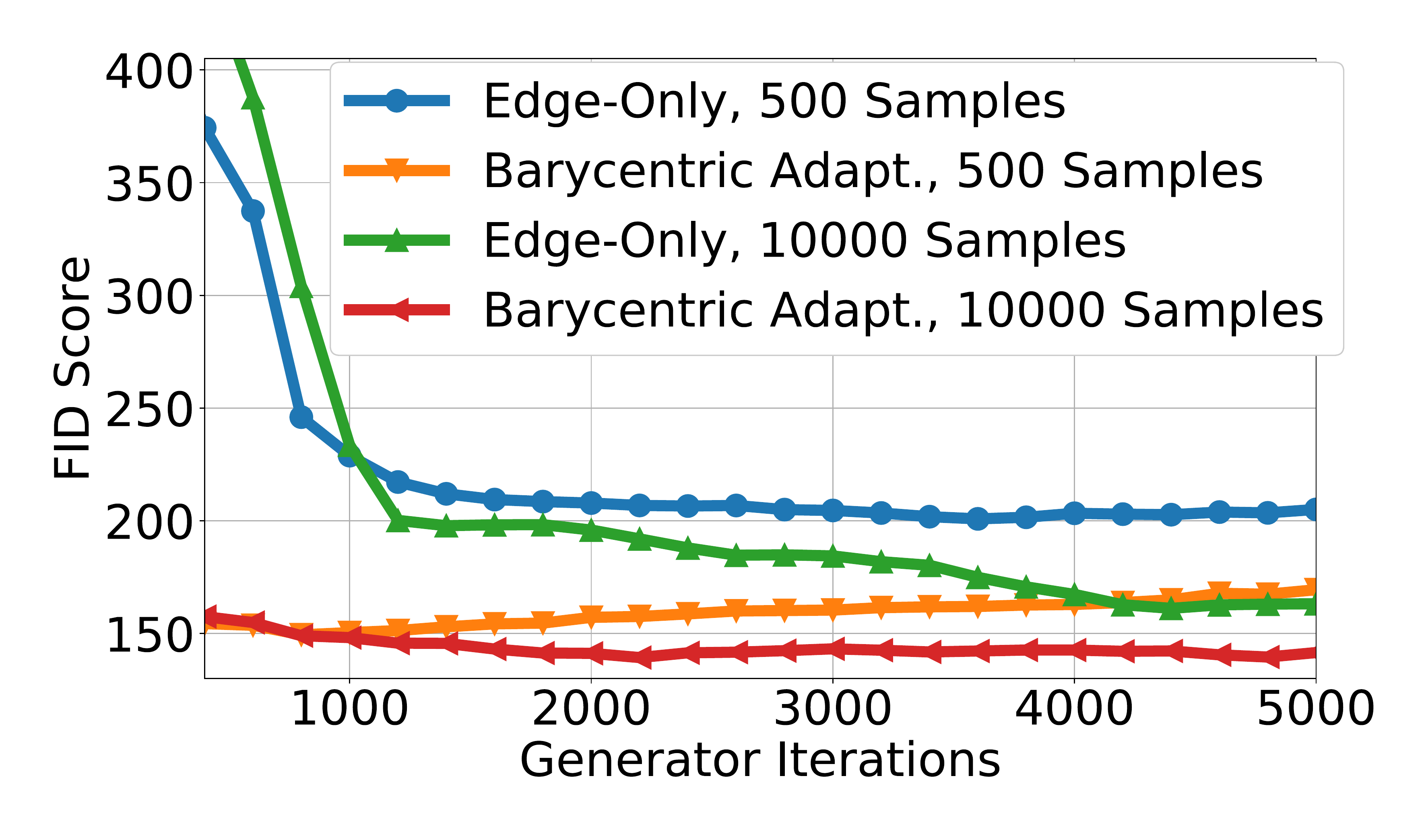}\label{sample1}}%
		\hspace{0.01\columnwidth}
		\subfigure[Comparison of convergence over LSUN: The overlapping case. ]{\includegraphics[width=0.67\columnwidth]{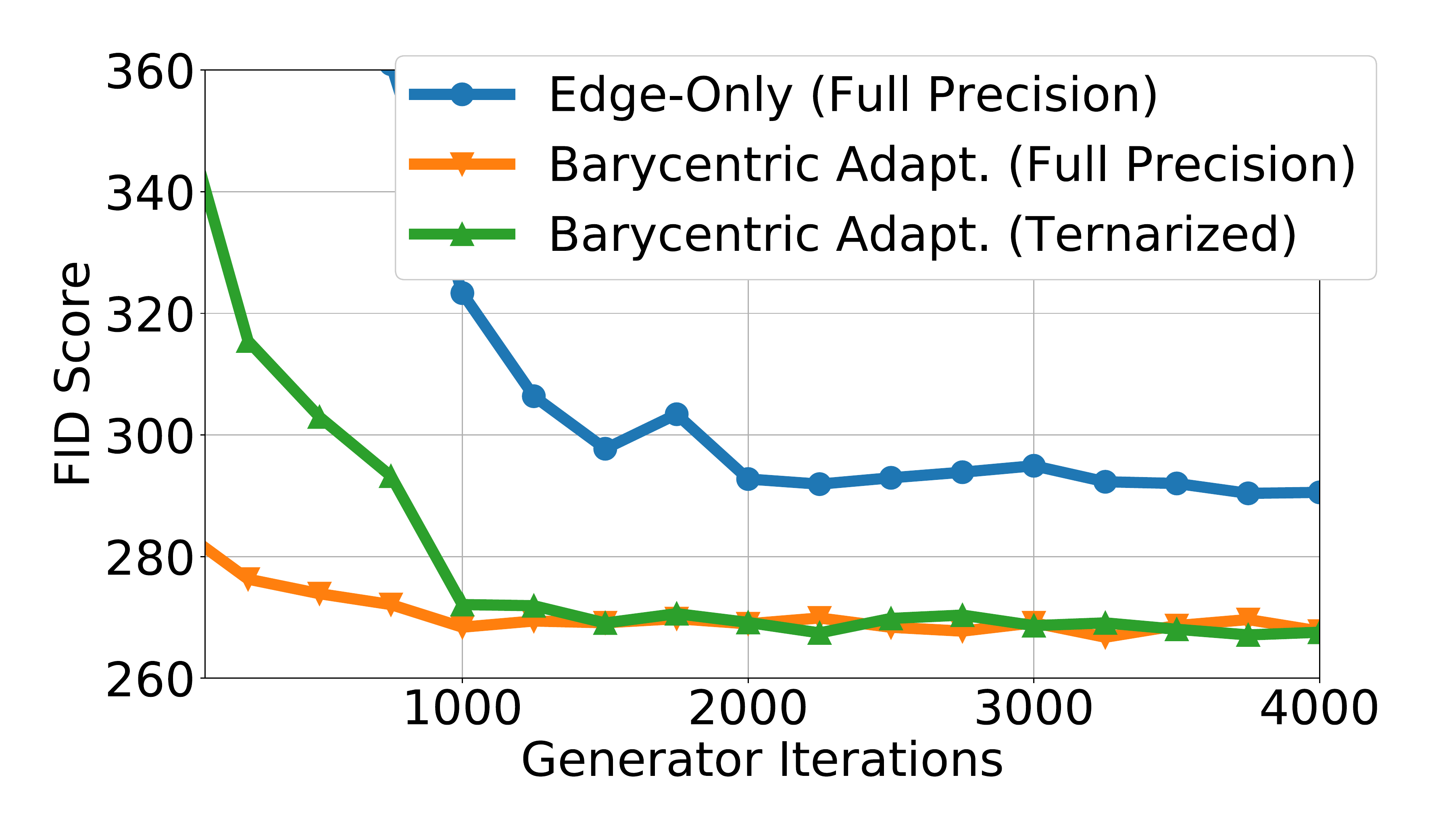}\label{ternaryperformanceLSUN}}%
		\hspace{0\columnwidth}
	\caption{Performance comparison of barycentric fast adaptation with various baselines. }\label{sampleNumber}
\end{figure*}

\begin{figure*}[t]
    \begin{center}
		\includegraphics[width=1.8\columnwidth]{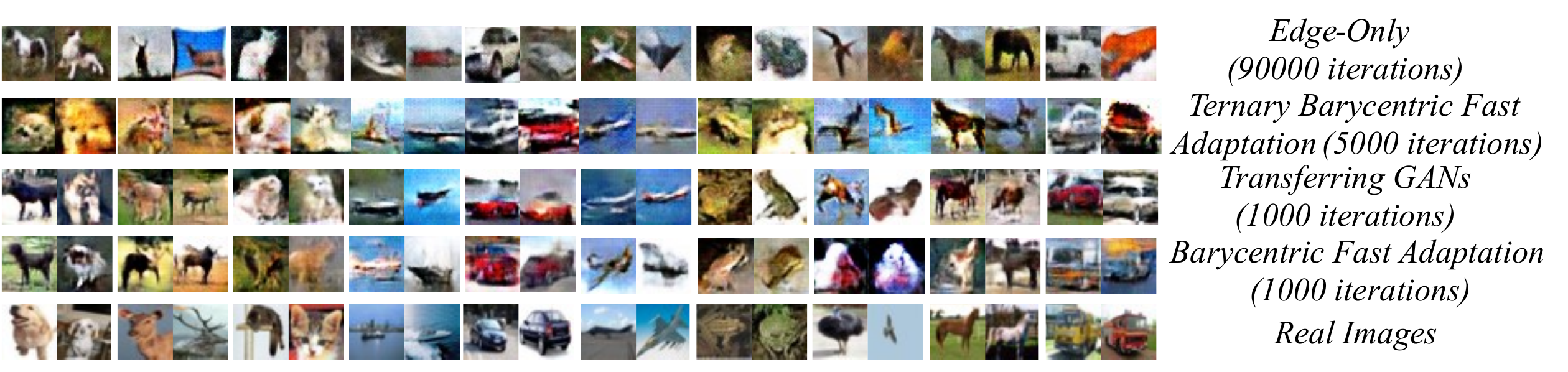}
	\end{center}
	\caption{Image samples from \(3\) different approaches for CIFAR10.}\label{images}
\end{figure*}

\textbf{Fine-tuning via fast adaptation.} We investigate the convergence and the generated image quality of various training scenarios on CIFAR100 and MNIST datasets. Specifically, we consider the following two scenarios: 1) \emph{The overlapping case}: the classes of the data samples at other edge nodes and at Node 0 overlap; 2) \emph{The non-overlapping case}: the classes of the data samples at other edge nodes and at Node 0 are mutually exclusive.
 As illustrated in Figure \ref{sampleNumber} and \ref{images}, \emph{barycentric fast adaptation} clearly outperforms all baselines. Transferring GANs suffers from catastrophic forgetting, because the continual learning is  performed over local data samples at Node 0 only. On the contrary, the barycentric fast adaptation and the ensemble method leverage generative replay, which mitigates the negative effects of catastrophic forgetting. Further, observe that the ensemble method suffers because of  the limited data samples at Node 0, which are significantly outnumbered by  synthetic data samples from pre-trained GANs, and this imbalance degrades the applicability of the ensemble method  for  continual learning. On the other hand, the barycentric fast adaptation can obtain the barycenter between the local data samples at Node 0 and the  barycenter model trained offline, and hence can effectively leverage the abundance of data samples from edge nodes and the accuracy of local data samples at Node 0 for better continual learning. 

\textbf{Impact of number of pre-trained generative models.} To quantify the impact of cumulative model knowledge from pre-trained generative models on the learning performance at the target node, we consider the scenario where \(10\) classes in CIFAR10/MNIST are split into \(3\) subsets, e.g., the first pre-train model has classes $\{0, 1, 2\}$, the second pre-trained model has classes $\{2, 3, 4\}$ and the third pre-trained model has the remaining classes. 
One barycenter model is trained  offline by using the first two pre-trained models and the second barycenter model is trained using all \(3\)  pre-trained models, respectively, based on which we evaluate the performance of  barycentric fast adaptation with \(1000\) data samples at the target node.
Figure \ref{trvsbaMNIST} and \ref{trvsbaCIFAR} showcase that the more model knowledge is accumulated in the barycenter computed offline, the higher image quality is achieved at Node 0. As expected, more model knowledge can help new edge nodes in training higher-quality generative models. In both figures, the barycentric fast adaptation outperforms Transferring GANs.

\textbf{Impact of the Number of Data Samples at Node 0.} 
Figure \ref{sample1} further illustrates the convergence across different number of data samples at the target node on CIFAR10 dataset. As expected, the FID score gap between barycentric fast adaptation and \emph{edge-only} method decreases as the number of data samples at the target node increases, simply because the empirical distribution becomes more `accurate'. In particular, the significant gap of FID scores between \emph{edge-only} and the \emph{barycentric fast adaptation} approaches in the initial stages indicates that the barycenter found via offline training and adopted as the model initialization for fast adaptation, is indeed close to the underlying model at the target node, hence enabling faster and more accurate edge learning than \emph{edge-only}. 

\textbf{Impact of Wasserstein ball radii.} 
Intuitively, the Wasserstein ball radius \(\eta_k\) for pre-trained model $k$ represents the relevance (and hence utility) of the knowledge transfer which is also intimately related to the capability to generalize beyond the pre-trained generative models, and the smaller it is, the more informative the corresponding Wasserstein ball is. Hence, larger weights $\lambda_k =1/\eta_k $ would be assigned to the nodes with higher relevance. We  note that the weights are determined by the  constraints and thus are fixed. Since we introduce the recursive WGAN configuration, the order of coalescence (each corresponding to a geodesic curve) may impact the final barycentric WGAN model, and hence the performance of barycentric fast adaptation. To this end, we compute the coalescence of models of nodes with higher relevance at  latter recursions to ensure that the final barycentric model is closer to the models of nodes with higher relevance. 

\textbf{Ternary WGAN based barycentric fast adaptation.} With the model initialization in the form of a full-precision barycenter model computed in offline training, we next train a ternary WGAN with 2 discriminators for the target node to compress the generative model further.
In particular, we use the same split of classes as the experiment illustrated in Figure \ref{sample1}, and compare the image quality obtained by Ternary WGAN-based fast adaptation against both full precision counterpart and \emph{Edge-Only}. It can be seen from the FID scores (Figure \ref{ternaryperformanceLSUN}), the ternary WGAN-based barycentric fast adaptation results in negligible performance degradation compared to its full precision counterpart, and is still much better compared to the \emph{Edge-Only} approach.

\section{Conclusions}

In this work, we propose a systematic framework for continual learning of generative models via adaptive coalescence of pre-trained models from other edge nodes. Particularly, we cast the  continual learning problem as a constrained optimization problem that can be reduced to a Wasserstein-1 barycenter problem. Appealing to optimal transport theory, we characterize the geometric properties of geodesic curves therein and use displacement interpolation as the foundation to devise recursive algorithms for finding adaptive barycenters. Next, we take a two-stage approach to efficiently solve the barycenter problem, where the barycenter of the pre-trained models is first computed offline in the cloud via a "recursive'' WGAN configuration based on displacement interpolation. Then,  the resulting barycenter is treated as the meta-model initialization and fast adaptation is used to find the generative model using the local samples at the target edge node. A weight ternarization method, based on  joint optimization of weights and threshold for quantization, is developed to compress the edge generative model further. Extensive experimental studies corroborate the efficacy of the proposed  framework.

\bibliography{example_paper}
\bibliographystyle{icml2021}
	
	%
	%
	%
	%
	%

	\newpage
	\appendix
	\onecolumn
	\section*{Appendices}
	\section{A Preliminary Review on Optimal Transport Theory and Wasserstein GANs}
	
	This section provides a brief overview of optimal transport  theory and Wasserstein GAN, which serves as the theoretic foundation for the proposed two-stage adaptive coalescence algorithm for fast edge learning of generative models. In particular, it is known that the Wasserstein-1 barycenter is  difficult to analyze,  because of the existence of infinitely many minimizers of the Monge Problem. We will review related geometric properties of geodesic curves therein and  introduce displacement interpolation.
	
	
	\subsection{Monge Problem and Optimal Transport Plan}
	
	Optimal transport theory has been extensively utilized in  economics for decades, and 
has recently garnered much interest in deep learning applications (see, e.g., \cite{Brenier,ambrosio2008,villani2008}). Simply put, optimal transport theory aims to find the most efficient transport map from one probability distribution to another with respect to a predefined cost function $c(x,y)$. The optimal distribution preserving the transport map can be obtained by solving the Monge problem.
	\begin{definition}\label{def1}(\textbf{Monge Problem})
		Let $(\mathcal{X},d)$ and $\mathcal{P}(\mathcal{X})$ denote a complete and separable metric space, i.e., a Polish space, and the set of probability distributions on $\mathcal{X}$, respectively. Given $\mu\in\mathcal{P}(\mathcal{X})$ and $\nu\in\mathcal{P}(\mathcal{Y})$  defined on two Polish spaces which are connected with a Borel map $T$, the Monge problem is defined as:
		\begin{align}
			&\underset{T:T\#\mu=\nu}{\inf}\int_{\mathcal{X}} c(x,T(x))d\mu(x).
		\end{align}
	\end{definition}
	In Definition \ref{def1}, $T$ is referred as the distribution preserving \emph{transport map} and $\#$ denotes the push-forward operator. In lieu of the strict constraint, there may not exist an optimal transport map for the Monge problem. A relaxation of the Monge problem leads to Kantorovich's optimal transport problem.
	
	\begin{definition}\label{def2}(\textbf{Kantorovich Problem}) Given $\mu\in\mathcal{P}(\mathcal{X})$ and $\nu\in\mathcal{P}(\mathcal{Y})$ are two probability distributions defined on two Polish spaces,  the Kantorovich problem is defined as:
		\begin{align}
			&\underset{\gamma\in\Pi(\mu,\nu)}{\inf}\int_{\mathcal{X}\times \mathcal{Y}} c(x,y)d\gamma(x,y),
		\end{align}
		where $\Pi(\mu,\nu)$ is the admissible set with its elements satisfying:
		\begin{align}
			\pi_\mu\#\gamma=\mu,&&\pi_\nu\#\gamma=\nu,
		\end{align}
		where $\pi_\mu$ and $\pi_\nu$ are two projector transport maps.
	\end{definition}
	In Definition \ref{def2}, $\gamma$ is referred as the \emph{transference plan} and the admissible set $\Pi$ is a relaxation to $T\#\mu=\nu$. A transference plan can leverage \emph{mass splitting} in contrast to transport maps, and hence can result in a solution under the semi-continuity assumptions. Mass splitting further enables the reputed Kantorovich duality, as shown in the following lemma, facilitating an alternative and convenient representation of the Kantorovich problem.
	
	\begin{lemma}\label{defkdual}(\textbf{Kantorovich Duality, \cite{villani2003}})
		Let $\mu\in\mathcal{P}(\mathcal{X})$ and $\nu\in\mathcal{P}(\mathcal{Y})$ be two probability distributions defined on Polish spaces $\mathcal{X}$ and $\mathcal{Y}$, respectively, and let $c(x,y)$ be a lower semi-continuous cost function. Further, define $\Phi_c$ as the set of all measurable functions $(\varphi, \psi)\in L^1(d\mu)\times L^1(d\nu)$ satisfying:
		\begin{align}
			\varphi(x)+\psi(y)\leq c(x,y),
		\end{align}
		for $d\mu$-almost all $x\in \mathcal{X}$ and $d\nu$-almost all $y\in \mathcal{Y}$. Then, the following strong duality holds for $c$-concave function $\varphi$:
		\begin{align}\label{kantordual}
			\underset{\gamma\in\Pi(\mu,\nu)}{\inf}\int_{\mathcal{X}\times \mathcal{Y}}c(x,y)d\gamma(x,y)=\underset{(\varphi,\psi)\in\Phi_c}{\sup} \int_\mathcal{X} \varphi(x)d\mu(x)+\int_\mathcal{Y} \psi(y)d\nu(y).
		\end{align}
	\end{lemma}
	As the right hand side of \eqref{kantordual} is an optimization over two functions, efficient gradient algorithms can be employed to learn the optimal solution. \eqref{kantordual} can be further simplified using $c$-transform \cite{villani2008}, in which $\psi(y)$ can be replaced by the $c$-transform $\varphi^c(y)={\inf}_{{x\in \mathcal{X}}}c(x,y)-\varphi(x)$, and $\varphi$ is referred as the Kantorovich potential. The following lemma establishes the existence of a Kantorovich potential that can also represent the Monge problem.
	
	\begin{lemma}(\textbf{Existence of Optimal Transport Plan, \cite{ambrosio2003lecture}})\label{thm1}
		For a lower semi-continuous cost function $c(x,y)$ defined on $\mathcal{X}\times \mathcal{Y}$, there exists at least one $\gamma\in\Pi(\mu,\nu)$ solving the Kantorovich problem. Furthermore, if $c(x,y)$ is continuous and real-valued, and $\mu$ has no atoms, then the minimums to both Monge and Kantorovich problems are equivalent, i.e.,
		\begin{align}
			\underset{T:T\#\mu=\nu}{\inf}\int_{\mathcal{X}} c(x,T(x))d\mu(x)=\underset{\gamma\in\Pi(\mu,\nu)}{\inf}\int_{\mathcal{X}\times \mathcal{Y}} c(x,y)d\gamma(x,y).
		\end{align}
	\end{lemma}
	Lemma \ref{thm1} indicates that there exists at least one transport map which are solutions to the Kantorovich problem. We here remark that not all transference plans are necessarily  transport maps. Lemma \ref{thm1} further facilitates a connection between dataset interpolation and the proposed Wasserstein GAN configuration in this study, along with the McCann's celebrated displacement interpolation result \cite{mccann1997convexity}.

	\subsection{From Vanilla Generative Adversarial Networks (GAN) to Wasserstein-1 GAN}
	
	A  generative adversarial network is comprised of a generator and discriminator neural networks. Random noise samples are fed into the generator to generate data samples of certain structure at the output of the generator. The generated (or fake) samples are then fed into the discriminator along with real-world samples taken from the dataset. The discriminator acts as a classifier and incurs a loss when mislabeling takes place. From a game theoretic point of view, the generator and the discriminator play a zero-sum game, in which the generator seeks to manipulate the discriminator to classify fake samples as real by generating  samples similar to the real-world dataset. In principle, GAN training is equivalent to solving for the following optimization problem:
	\begin{align}\label{q1}
		\underset{G}{\min}~\underset{D}{\max}~V(D,G)&=\underset{G}{\min}~\underset{D}{\max}~ \mathbb{E}_{\mathbf{x}\sim \mu}[\log D(\mathbf{x})]+\mathbb{E}_{\mathbf{z}\sim \vartheta}[\log(1-D(G(\mathbf{z})))]\nonumber\\
		&=\underset{G}{\min}~\underset{D}{\max}~ \mathbb{E}_{\mathbf{x}\sim \mu}[\log D(\mathbf{x})]+\mathbb{E}_{\mathbf{y}\sim \nu}[\log(1-D(\mathbf{y}))],
	\end{align}
	where $D$ and $G$ represent the discriminator and generator networks, respectively. Let $\mu$, $\nu$ and $\vartheta$ denote the distributions from empirical data, at generator output and at generator input, respectively. The latent distribution $\vartheta$ is often selected to be uniform or Gaussian. The output of the generator, denoted $\mathbf{y}=G(\mathbf{z},\theta_G)\sim \nu$, is composed by propagating $\mathbf{z}$ through a nonlinear transformation, represented by neural network parameter $\theta_G$. Model parameter $\theta_G$ entails $\nu$ to reside in a parametric probability distribution space $\mathcal{Q}_G$, constructed by passing $\vartheta$ through $G$. It has been shown in \cite{goodfellow2014}  that the solution to \eqref{q1} can be expressed as an optimization problem over $\nu$ as:
	\begin{align}\label{q2}
		\underset{\nu\in\mathcal{Q}_G}{\min}~-\log(4)+2\cdot \text{JSD}(\mu||\nu),
	\end{align}
	where $\text{JSD}$ denotes Jensen-Shannon divergence. Clearly, the solution to \eqref{q2} can be achieved at $\nu^*=\mu$, and the corresponding $\theta_G^*$ is the optimal generator model parameter. 

	The vanilla GAN training process suffers from the mode collapse issue that is often caused by vanishing gradients during the training process of GANs \cite{wgan}. In contrast to $\text{JSD}$, under mild conditions the Wasserstein distance does not incur vanishing gradients, and hence exhibits more useful gradient properties for preventing mode collapse. The training process of Wasserstein-1 distance based GAN can be expressed as solving an optimization problem ${\min}_{\nu\in\mathcal{Q}_G}~W_1(\nu,\mu)$.
	Since the $c$-transform of the Kantorovich potential admits a simpler and more convenient form for $W_1$, i.e., $\varphi^c=-\varphi$, the Wasserstein-1 GAN cost function can be rewritten as:
	\begin{align}\label{q5}
		W_1(\nu,\mu)=\underset{||\varphi||_L\leq 1}{\sup}\left\{\mathbb{E}_{\mathbf{x}\sim \mu}[\varphi(\mathbf{x})]-\mathbb{E}_{\mathbf{x}\sim \nu}[\varphi(\mathbf{y})]\right\},
	\end{align}
	where $\varphi$ is constrained to be a $1$-Lipschitz function. Following the same line as in the vanilla GAN, $\varphi$ in \eqref{q5} can be characterized by a neural network, which is parametrized by  model parameter $\theta_D$. Consequently, training a Wasserstein-1 GAN is equivalent to solve the following non-convex optimization problem through training the generator and discriminator neural networks:
	\begin{align}
		\underset{G}{\min}\underset{||\varphi||_L\leq 1}{\max}\left\{\mathbb{E}_{\mathbf{x}\sim \mu}[\varphi(\mathbf{x})]-\mathbb{E}_{\mathbf{y}\sim \nu}[\varphi(\mathbf{y})]\right\}=\underset{G}{\min}\underset{||\varphi||_L\leq 1}{\max}\left\{\mathbb{E}_{\mathbf{x}\sim \mu}[\varphi(\mathbf{x})]-\mathbb{E}_{\mathbf{z}\sim \vartheta}[\varphi(G(\mathbf{z}))]\right\}.
	\end{align}
	We here note that $\varphi$ must be selected from a family of $1$-Lipschitz functions. To this end, various training schemes have been proposed in the literature.

\subsection{From Wasserstein-1 Barycenter to Multi-Discriminator GAN Cost}

Problem \eqref{q8} can be expressed in terms of Kantorovich potentials by applying Kantorovich's Duality as:
\begin{align}\label{eq9}
\underset{\nu\in\mathcal{P}}{\min}\sum\limits_{k=1}^{K} \frac{1}{\eta_k} W_1(\nu,\mu_k)+ W_1(\nu,\hat{\mu}_{0})\hspace*{-0.03in}=\hspace*{-0.03in} \underset{\nu\in\mathcal{P}}{\min}~\sum\limits_{k=1}^{K}\frac{1}{\eta_k}&\underset{(\varphi_k,\psi_k)\in\Phi_c}{\sup} \left\{\int_\mathcal{X} \varphi_k(x)d\mu_k(x)+\int_\mathcal{Y} \psi_k(y)d\nu(y)\right\}\nonumber\\
+& \underset{(\varphi_0,\psi_0)\in\Phi_c}{\sup} \left\{\int_\mathcal{X} \varphi_0(x)d\hat{\mu}_0(x)+\int_\mathcal{Y} \psi_0(y)d\nu(y)\right\}.
\end{align}
By using \(c\)-transformation, we have \(\psi_k(y) = \varphi_k^c(y)\). In particular, for the Wasserstein-1 distance, we have that \(\varphi_k^c(y) = -\varphi_k(y)\), and hence \eqref{eq9} is further simplified as:
\begin{align}\label{multi_Gan}
    \underset{\nu\in\mathcal{P}}{\min}\sum\limits_{k=1}^{K} \frac{1}{\eta_k} &W_1(\nu,\mu_k)+ W_1(\nu,\hat{\mu}_{0}) = \underset{\nu\in\mathcal{P}}{\min}~\sum\limits_{k=1}^{K}\frac{1}{\eta_k} \underset{\lVert\varphi_k\rVert_L\leq 1}{\max}\left\{\mathbb{E}_{\mathbf{x}\sim\mu_k}[\varphi_k(\mathbf{x})]-\mathbb{E}_{\mathbf{y}\sim\nu}[\varphi_k(\mathbf{y})]\right\}\nonumber\\
    +& \underset{\lVert\varphi_0\rVert_L\leq 1}{\max}\left\{\mathbb{E}_{\mathbf{x}\sim\hat{\mu}_0}[\varphi_0(\mathbf{x})]-\mathbb{E}_{\mathbf{y}\sim\nu}[\varphi_0(\mathbf{y})]\right\} \nonumber\\ =&\underset{G}{\min} \underset{
		\{\lVert\varphi_k\rVert_L\leq 1\}_{0:K} }{\max}  \left\{\mathbb{E}_{\mathbf{x}\sim\hat{\mu}_0}[\varphi_0(\mathbf{x})] -\mathbb{E}_{{\mathbf{z}}\sim\vartheta}[\varphi_0(G(\mathbf{z}))]\right\}+\sum_{k=1}^{K}\frac{1}{\eta_k}\left\{\mathbb{E}_{\mathbf{x}\sim\mu_k}[\varphi_k(\mathbf{x})] -\mathbb{E}_{{\mathbf{z}}\sim\vartheta}[\varphi_k(G(\mathbf{z}))]\right\}.
\end{align}
Therefore, a barycenter of \(K\) distributions can  be obtained by minimizing the cost in \eqref{multi_Gan} through a specially designed GAN configuration.

\subsection{A Discussion on the Relationship between One-Shot and Recursive Configurations}

Even though the multi-discriminator GAN  can lead to a Wasserstein-1 barycenter in principle, training a multi-discriminator GAN in a one-shot manner is overwhelming for memory-limited edge nodes. The proposed 2-stage recursive configuration is designed to address the memory problem by converting the one-shot formulation to a nested Wasserstein barycenter problem. In a nutshell, a 2-discriminator GAN configuration suffices to obtain a shape-preserving interpolation of all distributions. As discussed above, the Wasserstein-1 barycenter problem not necessarily constitutes a unique solution due to the non-uniqueness of geodesic curves between distributions in the probability space. Proposition \ref{existence} asserts that any solution to each pairwise Wasserstein-1 barycenter problem, referred as a barycenter in this study, resides inside the baryregion formed by \(\{\mu_k\}_{1:K}\). Consequently, the final barycenter \(\nu^*\), obtained at the end of all recursions, also resides inside the baryregion. However, the 2-stage recursive configuration may not obtain the same barycenter solution to Wasserstein-1 barycenter problem. Through the intuition that the Wasserstein ball radius \(\eta_k=\frac{1}{\lambda_{\psi_k}}\) for pre-trained model $k$ represents the relevance (and hence utility) of the distribution \(k\), larger weights $\lambda_k =1/\eta_k $ would be assigned to the nodes with higher relevance. Since we introduce the recursive WGAN configuration, the order of coalescence (each corresponding to a geodesic curve) may impact the final barycentric WGAN model, and hence the performance of barycentric fast adaptation. To this end, we compute the coalescence of models of nodes with higher relevance at latter recursions to ensure that the final barycentric model is closer to the models of nodes with higher relevance.


\subsection{Refined Forming Set}
The following definition identifies a more compact forming set for baryregions when they exist.

	\begin{definition}(\textbf{Refined Forming Set})\label{def3}
		Let $\{\mu_k\}_{k\in\kappa}$ be a subset of the forming set $\{\mu_k\}_{1:K}$ for a set $\kappa\subset\mathcal{K}$, and let $\mathcal{B}_\mathcal{R}(\kappa)$ represent the baryregion facilitated by $\{\mu_k\}_{k\in\kappa}$. The smallest subset $\{\mu_k\}_{k\in\kappa^*}$, satisfying $\mathcal{B}_\mathcal{R}(\kappa^*)\supseteq\mathcal{B}_\mathcal{R}$, is defined as the refined forming set of $\mathcal{B}_\mathcal{R}$.
	\end{definition}
	A refined forming set can characterize a baryregion as complete as the original forming set, but can better capture the geometric properties of the barycenter problem. In particular, a refined forming set $\kappa^*$ dictates that $\{\mu_k\}_{k\in\kappa^*}$ engenders exactly the same geodesic curves as in $\mathcal{B}_\mathcal{R}$. 
	\begin{proposition}(\textbf{Non-uniqueness})\label{refined}
		A refined forming set of $\{\mu_k\}_{1:K}$ is not necessarily unique.
	\end{proposition}

	\begin{proof}
	To prove Proposition 1, it suffices to construct a counter example. Consider a forming set $\{\mu_k\}_{1:4}$ with the probability measures $\mu_1=\delta_{(0,0)}$, $\mu_2=\delta_{(1,0)}$, $\mu_3=\delta_{(0,1)}$, and $\mu_4=\delta_{(1,1)}$, where $\delta_{(\mathbf{a},\mathbf{b})}$ is the delta function with value $1$ at $(x,y)=(\mathbf{a},\mathbf{b})$ and $0$ otherwise. Further, let $\{\mu_k\}_{k\in\{1,4\}}$ and $\{\mu_k\}_{k\in\{2,3\}}$ be two subsets of the forming set. Then, the length of the minimal geodesic curve between $\mu_1$ and $\mu_4$ can be computed as:
	\begin{align}\label{ap1}
		W_1(\mu_1(x),\mu_4(y))&=\underset{\gamma\in\Pi(\mu_1,\mu_4)}{\inf} \int_{\mathcal{X}\times \mathcal{Y}}d(x,y)d\gamma(x,y)\nonumber\\ &=\int_{\mathcal{Y}}\int_{\mathcal{X}}d([0,0]^T,[1,1]^T)\delta_{([0,0]^T,[1,1]^T)}dxdy= 2.
	\end{align}
	By recalling that there  exist infinitely many minimal geodesics satisfying \eqref{ap1}, we check the lengths of two other sets of geodesics that traverse through $\mu_2$ and $\mu_3$, respectively. First, for $\mu_2$, 
	\begin{align}
		W_1(&\mu_1(x),\mu_4(y))\leq W_1(\mu_1(x),\mu_2(z))+W_1(\mu_2(z),\mu_4(y))\nonumber\\
		&=\underset{\gamma\in\Pi(\mu_1,\mu_2)}{\inf} \int_{\mathcal{X}\times \mathcal{Z}}d(x,z)d\gamma(x,z)+\underset{\gamma\in\Pi(\mu_2,\mu_4)}{\inf} \int_{\mathcal{Z}\times \mathcal{Y}}d(z,y)d\gamma(z,y)\nonumber\\
		&=\int_{\mathcal{Z}}\int_{\mathcal{X}}d([0,0]^T,[1,0]^T)\delta_{([0,0]^T,[1,0]^T)}dxdz+\int_{\mathcal{Y}}\int_{\mathcal{Z}}d([1,0]^T,[1,1]^T)\delta_{([1,0]^T,[1,1]^T)}dzdy\nonumber\\
		&=2\leq W_1(\mu_1(x),\mu_4(y)),
	\end{align}
	based on the triangle inequality and the definition of first-type Wasserstein distance. Similarly for $\mu_3$, we can show that
	\begin{align}
		W_1(\mu_1(x),\mu_4(y))\leq W_1(\mu_1(x),\mu_3(z))+W_1(\mu_3(z),\mu_4(y))\leq W_1(\mu_1(x),\mu_4(y)),
	\end{align}
	through the selections $\gamma(x,z)=\delta_{([0,0]^T,[0,1]^T)}$ and $\gamma(z,y)=\delta_{([0,1]^T,[1,1]^T)}$. As a result,  there exists at least a single minimal geodesic between $\mu_1$ and $\mu_4$ passing through $\mu_\ell$ for $\ell\in\{2,3\}$, indicating that $\mu_2,\mu_3\in\mathcal{R}(\{\mu_k\}_{k\in\{1,4\}})$ and  $\mathcal{B}_{\mathcal{R}}(\{\mu_k\}_{k\in\{1,4\}})\supseteq \mathcal{B}_{\mathcal{R}}$. Observing that there exists no smaller forming set than $\{\mu_k\}_{k\in\{1,4\}}$,  we conclude that $\{\mu_k\}_{k\in\{1,4\}}$ is a refined forming set.


	Following the same line, we can have that $\{\mu_k\}_{k\in\{2,3\}}$ is another refined forming set of $\{\mu_k\}_{1:4}$ by first showing the following three inequalities:
	\begin{align}
		W_1(\mu_2(x),\mu_3(y))&= \int_{\mathcal{Y}}\int_{\mathcal{X}}d([1,0]^T,[0,1]^T)\delta_{([1,0]^T,[0,1]^T)}dxdy=2,\\
		W_1(\mu_2(x),\mu_3(y))&\leq W_1(\mu_2(x),\mu_1(z))+W_1(\mu_1(z),\mu_3(y))\leq W_1(\mu_2(x),\mu_3(y)),\label{ap27}\\
		W_1(\mu_2(x),\mu_3(y))&\leq W_1(\mu_2(x),\mu_4(z))+W_1(\mu_4(z),\mu_3(y))\leq W_1(\mu_2(x),\mu_3(y)),\label{ap28}
	\end{align}
	where the transport maps $\gamma(x,z)=\delta_{([1,0]^T,[0,0]^T)}$ and $\gamma(z,y)=\delta_{([0,0]^T,[0,1]^T)}$ for \eqref{ap27}, and $\gamma(x,z)=\delta_{([1,0]^T,[1,1]^T)}$ and $\gamma(z,y)=\delta_{([1,1]^T,[0,1]^T)}$ for \eqref{ap28}. Consequently, there exists at least a single minimal geodesic between $\mu_2$ and $\mu_3$ passing through $\mu_\ell$ for $\ell\in\{1,4\}$, indicating that $\mu_1,\mu_4\in\mathcal{R}(\{\mu_k\}_{k\in\{2,3\}})$ and  $\mathcal{B}_{\mathcal{R}}(\{\mu_k\}_{k\in\{2,3\}})\supseteq \mathcal{B}_{\mathcal{R}}$. Since there exists no smaller forming set than $\{\mu_k\}_{k\in\{2,3\}}$, we have that  $\{\mu_k\}_{k\in\{2,3\}}$ is another refined forming set,  thereby 
	completing the proof of non-uniqueness.
	\end{proof}

\section{Proof of Proposition~\ref{existence}}
\begin{proof} Let $\{\mu_k\}_{1:K}$ be any set of probability measures on a refined forming set and $\nu_k^*$ denote a continuous probability measure with no atoms, which minimizes the problem $\underset{\nu_k}{\min}~W_1(\mu_{k},\nu_k)+W_1(\nu_{k-1}^*,\nu_k)$ \cite{ambrosio2008}. By Proposition \ref{refined}, there exists multiple refined forming sets, and the proceeding proof holds true for any refined forming set induced by the original set of probability distributions. The proceeding proof utilizes the geodesic property and the existence of a barycenter in Wasserstein-1 space, for which the details can be found in \cite{villani2003,ambrosio2008} and \cite{le2017existence}, respectively. Suppose  that $\alpha\not\in \mathcal{B}_\mathcal{R}$ is a distribution satisfying
	\begin{align}\label{prop_wrong}
		W_1(\mu_2,\nu_2^*)+W_1(\mu_1,\nu_2^*)= W_1(\mu_2,\alpha)+W_1(\mu_1,\alpha).
	\end{align}
Let $\nu_2^*=\alpha$. Note that if $\alpha\not\in \mathcal{B}_\mathcal{R}$, $\alpha$ cannot reside on the geodesic curve $g_t(\mu_1, \mu_2)_{0\leq t\leq 1}$ since $g_t(\mu_1, \mu_2)_{0\leq t\leq 1}\in\mathcal{B}_\mathcal{R}$. Subsequently, by considering another distribution $\beta$ which resides on geodesic curve $g_t(\mu_1,\mu_2)$, we can also show that:
	\begin{align}
		W_1(\mu_1,\beta)+W_1(\mu_2,\beta)&=W_1(\mu_1,\beta)+W_1(\beta,\mu_2) =W_1(\mu_1,\mu_2)\nonumber\\
		&<W_1(\mu_1,\alpha)+W_1(\alpha,\mu_2)= W_1(\mu_2,\nu_2^*)+W_1(\mu_1,\nu_2^*),
	\end{align}
	indicating that $\beta$ attains a lower cost than the minimizer $\nu_2^*$, which is a contradiction, indicating that $\nu_2^*$ must reside in $\mathcal{B}_\mathcal{R}$. Similarly, $\nu_3^*$ must also reside in $\mathcal{B}_\mathcal{R}$:
	\begin{align}
		W_1(\mu_3,\beta)+W_1(\nu_2^*,\beta)&=W_1(\mu_3,\beta)+W_1(\beta,\nu_2^*) =W_1(\mu_3,\nu_2^*)\nonumber\\
		&<W_1(\mu_3,\alpha)+W_1(\alpha,\nu_2^*).
	\end{align}
	By induction, $\beta\in\mathcal{B}_\mathcal{R}$ attains a lower cost compared with $\alpha\not\in\mathcal{B}_\mathcal{R}$ at the $k$th iteration:
	\begin{align}
		W_1(\mu_{k},\beta)+W_1(\nu_{k-1}^*,\beta)&=W_1(\mu_{k},\beta)+W_1(\beta,\nu_{k-1}^*) =W_1(\mu_{k},\nu_{k-1}^*)\nonumber\\
		&<W_1(\mu_{k},\alpha)+W_1(\alpha,\nu_{k-1}^*).
	\end{align}
	Hence, $\nu_k^*=\beta\in\mathcal{B}_\mathcal{R}$. Consequently, all barycenters to at each iteration must reside in the baryregion $\mathcal{B}_\mathcal{R}$. 
	
	Similarly, we can show that for stage $\mathrm{II}$ the following holds:
	\begin{align}
		W_1(\mu_0,\beta)+W_1(\nu_K^*,\beta)&=W_1(\mu_0,\beta)+W_1(\beta,\nu_K^*) =W_1(\mu_0,\nu_K^*)\nonumber\\
		&<W_1(\mu_0,\alpha)+W_1(\alpha,\nu_K^*).
	\end{align}
	Consequently, $\nu^*$ also resides in $\mathcal{B}_\mathcal{R}$, which completes the proof.
\end{proof}

\section{Algorithms and Experiment Settings}

\subsection{Algorithms}
\begin{figure}[t]
		\begin{center}
			\centerline{\includegraphics[width=\columnwidth]{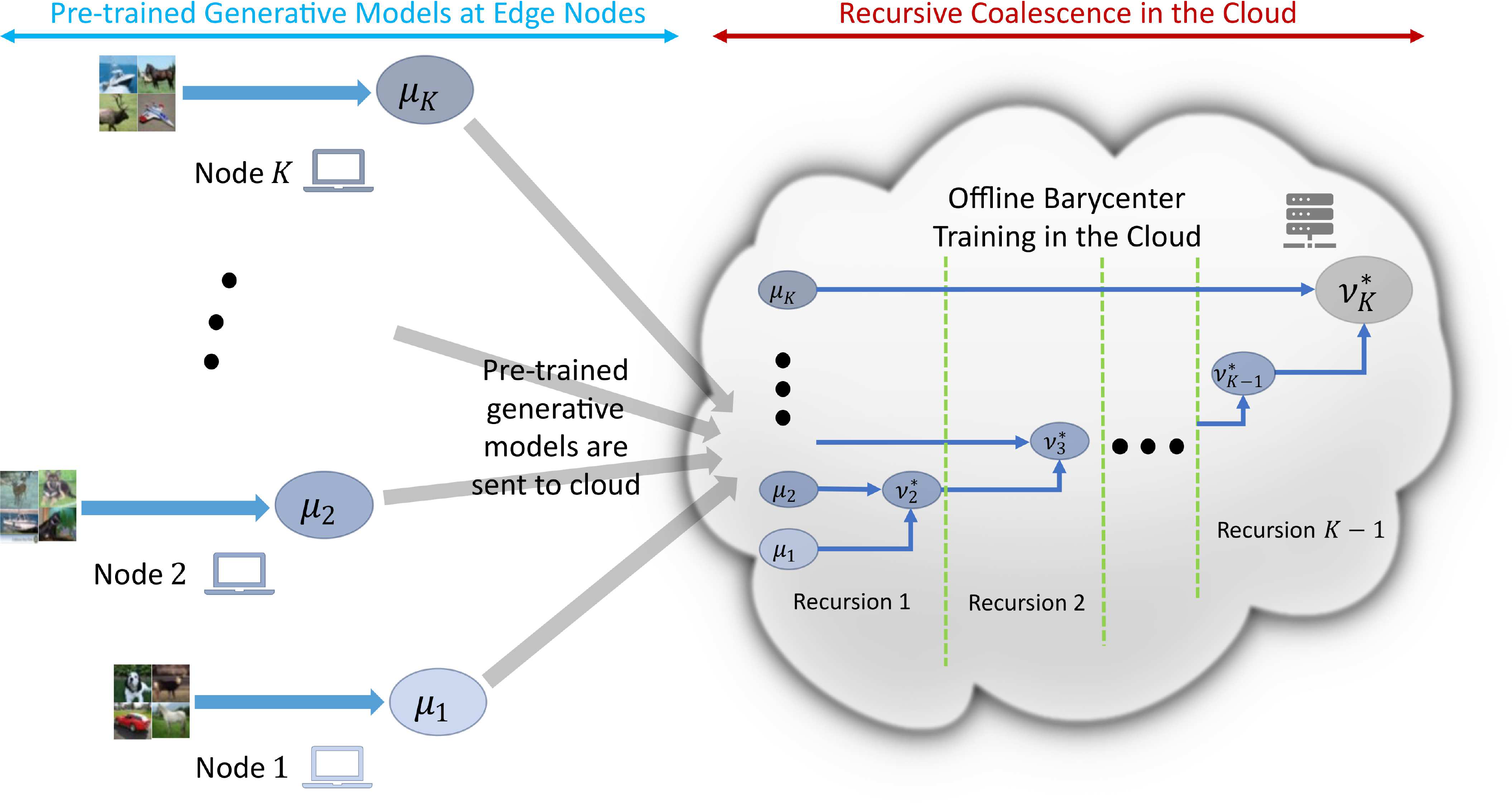}}%
			\caption{Illustration of offline training in the cloud:  Each edge node sends pre-trained generative model (instead of  datasets) to the cloud, based on which  the cloud  computes adaptive barycenters using the recursive configuration.}\label{detailedflow}
		\end{center}
		\vskip -0.2in
	\end{figure}

For the proposed two-stage adaptive coalescence algorithm,  the offline training in Stage I is done in the cloud, and the fast adaptation in Stage II is carried out at the edge server, in the same spirit as the model update of Google EDGE TPU. Particularly, as illustrated in Figure \ref{detailedflow}, each edge node sends its pre-trained generative model  (instead of its own training dataset) to the cloud. As noted before, the amount of bandwidth required to transmit data from an edge node to cloud is also significantly reduced by  transmitting only a generative model, because neural network model parameters require much smaller storage than the dataset itself. The algorithms developed in this study are summarized as follows:	

	\begin{algorithm}[H]
		\footnotesize
		\caption{Offline training to solve the barycenter of $K$ pre-trained generative models}
		\label{alg1}
		\begin{algorithmic}[1]
			\STATE\textbf{Inputs:} $K$ pre-trained generator-discriminator pairs $\{(G_k, D_k)\}_{1:K}$ of corresponding source nodes $k\in\mathcal{K}$, noise prior $\vartheta(z)$, the batch size $m$, learning rate $\alpha$\;
			\STATE\textbf{Outputs:} Generator $\mathcal{G}^*_{K}$ for barycenter $\nu_K^*$, discriminators $\tilde{\psi}_{K}^*$, $\psi_{K}^*$;
			\STATE Set $\mathcal{G}^*_{1}\leftarrow G_1$, $\tilde{\psi}^*_{1}\leftarrow D_{1}$;
			//\textbf{Barycenter initialization}
			\FOR{iteration $k= 2, ..., K$}
			\STATE Set $\mathcal{G}_{k}\leftarrow \mathcal{G}^*_{k-1}$, $\tilde{\psi}_k\leftarrow \{\tilde{\psi}_{k-1}^*,\psi_{k-1}^*\}$, $\psi_k\leftarrow D_k$ and choose  $\lambda_{\tilde{\psi}_k}$, $\lambda_{\psi_k}$;
			//\textbf{Recursion initialization}
			\WHILE{ generator $\mathcal{G}_{k}$ has not converged}
			\STATE Sample batches of prior samples $\{z^{(i)}\}_{i=1}^m$, $\{{z}^{(i)}_{\tilde{\psi}_k}\}_{i=1}^m$, $\{{z}^{(i)}_{\psi_k}\}_{i=1}^m$ independently from prior $\vartheta(z)$;
			\STATE Generate synthetic data batches $\{{x}^{(i)}_{\tilde{\psi}_k}\}_{i=1}^m\sim \nu_{k-1}^*$ and $\{{x}^{(i)}_{\psi_k}\}_{i=1}^m\sim \mu_k$ by passing $\{{z}^{(i)}_{\tilde{\psi}_k}\}_{i=1}^m$ and $\{{z}^{(i)}_{\psi_k}\}_{i=1}^m$ through $\mathcal{G}_{k-1}^*$ and $G_k$, respectively;
			\STATE Compute gradients $g_{\tilde{\psi}_k}$ and $g_{\psi_k}$: $\left\{g_{\omega}\leftarrow\lambda_\omega\nabla_{\omega}\frac{1}{m}\sum_{i=1}^{m} \left[\omega(x^{(i)}_\omega)- \omega(\mathcal{G}_k(z^{(i)})\right]\right\}_{\omega = {\tilde{\psi}_k}, {\psi_k} }$;
			\STATE Update both discriminators $\psi_k$ and $\tilde{\psi}_k$: $\left\{\omega\leftarrow \omega+\alpha\cdot \text{Adam}(\omega, g_\omega)\right\}_{\omega=\psi_k,\tilde{\psi}_k}$;
			
			\STATE Compute gradient $g_{\mathcal{G}_k}\leftarrow -\nabla_{\mathcal{G}_k}\frac{1}{m} \sum_{i=1}^{m} \left[ \lambda_{\psi_k}\psi_k(\mathcal{G}_k(z^{(i)}))+ \lambda_{\tilde{\psi}_k}\tilde{\psi}_k(\mathcal{G}_k(z^{(i)}))\right]$;
			\STATE Update generator $\mathcal{G}_{k}$: $\mathcal{G}_{k}\leftarrow \mathcal{G}_{k}-\alpha\cdot \text{Adam}(\mathcal{G}_{k}, g_{\mathcal{G}_{k}})$ until optimal generator $\mathcal{G}^*_{k}$\ is computed;
			\ENDWHILE
			\ENDFOR
			\STATE\textbf{return} generator $\mathcal{G}^*_K$, discriminators $\tilde{\psi}_{K}^*$, $\psi_{K}^*$.
		\end{algorithmic}
	\end{algorithm}

	\begin{algorithm}[H]
	  \footnotesize
		\caption{Fast adaptive learning of the ternary generative model for edge Node 0}
		\label{alg2}
	 	\begin{algorithmic}[1]
			  \STATE\textbf{Inputs:}  Training dataset $\mathcal{S}_0$, generator $\mathcal{G}^*_{K}$ for the barycenter $\nu_K^*$, offline pre-trained discriminators $\psi_{K}^*$, $\tilde{\psi}_{K}^*$, noise prior $\vartheta(z)$, the batch size $m$, learning rate $\alpha$, the number of layers $L_\mathcal{G}=L_\psi=L_{\tilde{\psi}}=L$;
			  \STATE\textbf{Outputs:} the ternary generator $\mathcal{G}_{0}$;
			  \STATE Set $\mathcal{G}_{0}\leftarrow \mathcal{G}^*_{K}$, $\tilde{\psi}_0\leftarrow\tilde{\psi}_{K}^*$ and $\psi_{0}\leftarrow\psi_{K}^*$;
			  //\textbf{Initialization}
			  \WHILE{generator $\mathcal{G}_{0}$ has not converged}
			         \FOR{$l:=1$ to $L$~//\textbf{Weight ternarization}}
			         	\STATE $\left\{\mathbf{w}^\prime_{l_{\omega}}\leftarrow Tern(\mathbf{w}_{l_{\omega}}, S_{l_{\omega}}, \Delta_{l_{\omega}}^{\pm})\right\}_{\omega = {\tilde{\psi}_k}, {\psi_k} }$;
			         	\STATE $\mathbf{w}^\prime_{l_\mathcal{G}}\leftarrow Tern(\mathbf{w}_{l_\mathcal{G}}, S_{l_\mathcal{G}}, \Delta_{l_\mathcal{G}}^{\pm})$;
			         \ENDFOR
			     \STATE Sample batches of prior samples $\{z^{(i)}\}_{i=1}^m$  from prior $\vartheta(z)$;
			         \STATE Sample batches of training samples $\{x_0^{i}\}_{i=1}^{m}$ from local dataset $\mathcal{S}_0$;
			         \FOR{$l:=L$ to $1$~//\textbf{Update the thresholds}}
			         \STATE Compute gradients $\left\{g_{\Delta_{l_\omega}^{\pm}}\right\}_{\omega = {\tilde{\psi}_k}, {\psi_k} }$: $\left\{g_{\Delta_{l_\omega}^{\pm}}\leftarrow\nabla_{\Delta_{l_\omega}^{\pm}}\frac{1}{m}\sum_{i=1}^{m} \left[\omega_{0}(x^{(i)}_0)-\omega_{0}(\mathcal{G}_0(z^{(i)})\right]\right\}_{\omega = {\tilde{\psi}_k}, {\psi_k} }$;
			         \STATE Update $\left\{\Delta_{l_\omega}^{\pm}\right\}_{\omega = {\tilde{\psi}_k}, {\psi_k} }$: $\left\{\Delta_{l_\omega}^{\pm}\leftarrow \Delta_{l_\omega}^{\pm}+\alpha\cdot g_{\Delta_{l_\omega}^{\pm}}\right\}_{\omega = {\tilde{\psi}_k}, {\psi_k} }$;
			        \STATE Compute gradient $g_{\Delta_{l_\mathcal{G}}^{\pm}}$: $g_{\Delta_{l_\mathcal{G}}^{\pm}}\leftarrow -\nabla_{\Delta_{l_\mathcal{G}}^{\pm}}\frac{1}{m} \sum_{i=1}^{m} \left[ \psi_0(\mathcal{G}_0(z^{(i)}))+ \tilde{\psi}_0(\mathcal{G}_0(z^{(i)}))\right]$;
			        \STATE Update $\Delta_{l_\mathcal{G}}^{\pm}$: $\Delta_{l_\mathcal{G}}^{\pm}\leftarrow \Delta_{l_\mathcal{G}}^{\pm}-\alpha\cdot g_{\Delta_{l_\mathcal{G}}^{\pm}}$;
			        \ENDFOR
			        \STATE Repeat step 3-5 using updated thresholds;
			        \FOR{$l:=L$ to $1$~//\textbf{Update the full-precision weights}}
			         \STATE Compute gradients $\left\{g_{\mathbf{w}_{l_\omega}}\right\}_{\omega = {\tilde{\psi}_k}, {\psi_k} }$: $\left\{g_{\mathbf{w}_{l_\omega}}\leftarrow\nabla_{\mathbf{w}_{l_\omega}}\frac{1}{m}\sum_{i=1}^{m} \left[\omega_{0}(x^{(i)}_0)-\omega_{0}(\mathcal{G}_0(z^{(i)})\right]\right\}_{\omega = {\tilde{\psi}_k}, {\psi_k} }$;
			         \STATE Update $\left\{\mathbf{w}_{l_\omega}\right\}_{\omega = {\tilde{\psi}_k}, {\psi_k} }$: $\left\{\mathbf{w}_{l_\omega}\leftarrow \mathbf{w}_{l_\omega}+\alpha\cdot \text{Adam}(\mathbf{w}_{l_\omega}, g_{\mathbf{w}_{l_\omega}})\right\}_{\omega = {\tilde{\psi}_k}, {\psi_k} }$;
			        \STATE Compute gradient $g_{\mathbf{w}_{l_\mathcal{G}}}$: $g_{\mathbf{w}_{l_\mathcal{G}}}\leftarrow -\nabla_{\mathbf{w}_{l_\mathcal{G}}}\frac{1}{m} \sum_{i=1}^{m}  \left[ \psi_0(\mathcal{G}_0(z^{(i)}))+ \tilde{\psi}_0(\mathcal{G}_0(z^{(i)}))\right]$;
			        \STATE Update $\mathbf{w}_{l_\mathcal{G}}$: $\mathbf{w}_{l_\mathcal{G}}\leftarrow \mathbf{w}_{l_\mathcal{G}}-\alpha\cdot \text{Adam}(\mathbf{w}_{l_\mathcal{G}}, g_{\mathbf{w}_{l_\mathcal{G}}})$;
			        \ENDFOR
			        \STATE Repeat step 3-5 using updated full-precision weights;
			    \ENDWHILE
			\STATE\textbf{return} the ternary generator $\mathcal{G}_{0}$.
		\end{algorithmic}
	\end{algorithm}
	
\subsection{Experiment Settings}
	
This section outlines the architecture of deep neural networks and hyper-parameters used in the experiments. 

\begin{figure}[H]
		\begin{center}
			\centerline{\includegraphics[width=0.9\columnwidth]{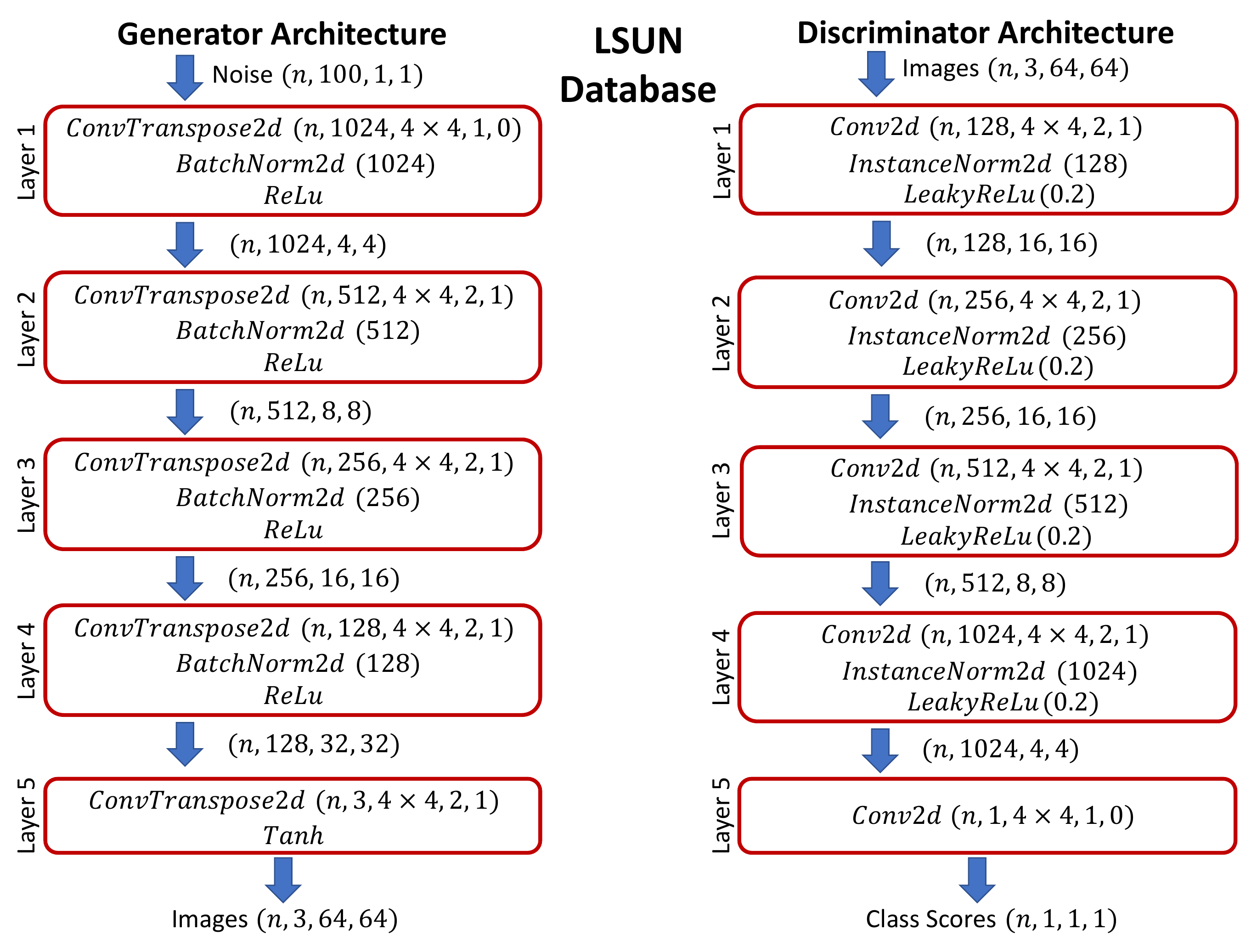}}%
			\caption{The DNN architecture used in experiments for LSUN dataset. }\label{fig_7}
		\end{center}
	\end{figure}

\begin{figure}[H]
		\begin{center}
			\centerline{\includegraphics[width=0.85\columnwidth]{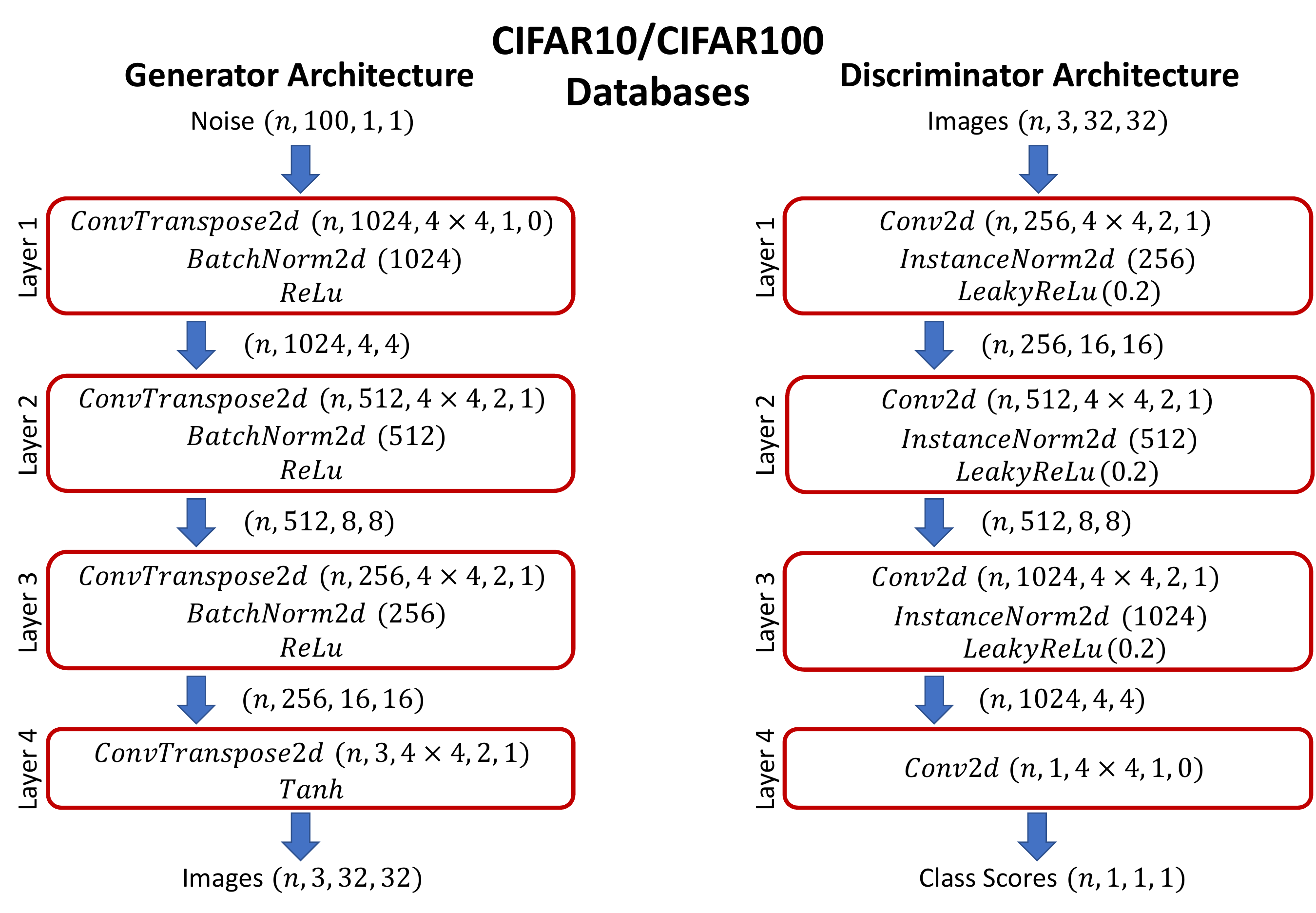}}%
			\caption{The DNN architecture used in experiments for CIFAR10 and CIFAR100 datasets. }\label{fig6}
			\vspace{-0.3in}
		\end{center}
	\end{figure}

\begin{figure}[H]
		\begin{center}
			\centerline{\includegraphics[width=0.85\columnwidth]{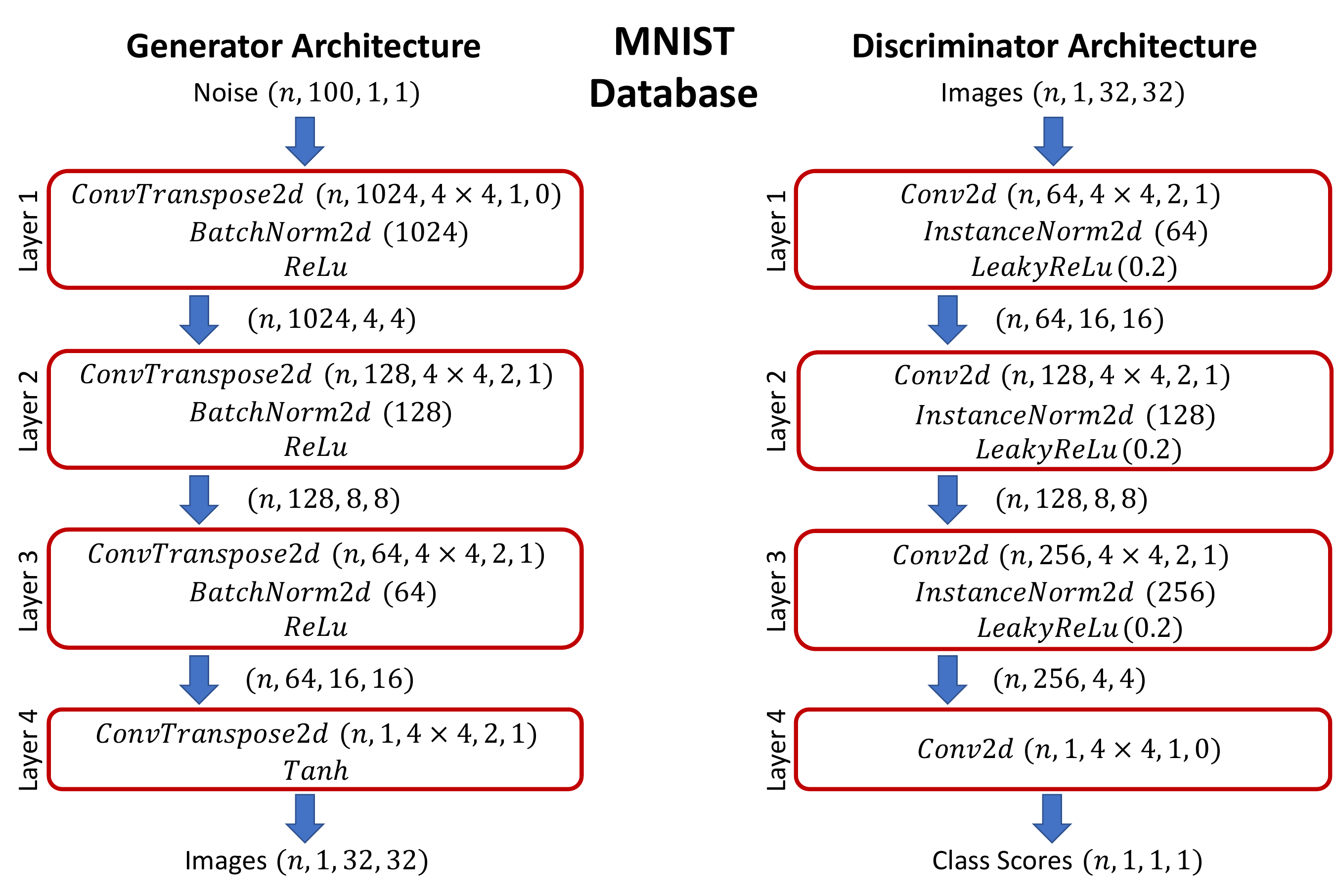}}%
			\caption{The DNN architecture used in experiments for MNIST dataset. }\label{fig7}
			\vspace{-0.3in}
		\end{center}
	\end{figure}

\noindent\textbf{Network architectures deployed in the experiments.}
Figures \ref{fig_7}, \ref{fig6} and \ref{fig7} depict the details of the DNN architecture used in our experiments; the shapes for convolution layers follow \((batch~size, number~of~filters, kernel~size, stride, padding)\); and the shapes for network inputs follow \((batch~size, number~of~channels, heights, widths)\).

\noindent\textbf{Hyper-parameters used in the experiments.}
All experiments are conducted in PyTorch on a server with RTX 2080 Ti and 64GB of memory. The selection of most parameter values, e.g., the number of generator iterations, batch size, optimizer, gradient penalty factor, and the number of discriminator iterations per generator iterations, follows \cite{wgan, wgangp, wang2018transferring}. For other parameters, we select the values giving the best performance via trial-and-error. In Table \ref{parameters} and \ref{parameters2} all hyper-parameters are listed. We have considered different ranges of values for different parameters. The number of generator iterations (fast adaptation) ranges from \(800\) up to \(100000\). For better illustration, the figures depict only the iterations until satisfactory image quality is achieved. For the number of samples at the target edge node, \(500 \sim 10000\) samples in CIFAR10, \(20 \sim 500\) samples in MNIST and \(500 \sim 1000\) samples in LSUN and CIFAR100 are used.
Each experiment is smoothed via a moving average filter for better visualization. More details and instructions to modify the hyper-parameters are available in the accompanying code, which will be publicly available on GitHub once the review process is over.

\begin{table*}\centering
\ra{1.3}
\caption{List of hyper-parameters used in all experiments.}\label{parameters}
\begin{tabular}{@{}lllllll@{}}\toprule
 &\multicolumn{3}{c}{CIFAR10} & \\\cmidrule{2-4}
Parameter\textbackslash Experiment Figure & Fig. \ref{sample1} & Fig. \ref{ternaryperformance} & Fig. \ref{trvsbaCIFAR}& \\ \midrule
Num. of Generator Iter. (Offline Training) & 100000 & 100000 & 5000 &\\
Num. of Generator Iter. (fast adaptation) & 5000 & 2000 & 3000 &\\
Batch Size & 64 & 64 & 64 &\\
Optimizer Parameters & \multicolumn{3}{c}{Adam(\(\beta_1=0.5, \beta_2=0.999\), l.r.\(=0.0001\))}& \\
Num. of Discriminator Iter. per Gen. Iter.& 5 & 5 & 5& \\
Gradient-Penalty Factor & 10 & 10 & 10&\\
Num. of Samples in Node \(0\) & Varies & 1000 & 1000 &\\
\(\{\lambda_k\}\) & Equal & Equal  & Equal& \\
Number of Training Samples  & 50000 & 50000 & 50000 &\\
Avg. Dur. of Pretraining/Transfer per Iter. & \multicolumn{3}{c}{0.755 seconds} &\\
Avg. Dur. of Training Barycenter per Iter. & \multicolumn{3}{c}{1.480 seconds} &\\

\midrule 
&\multicolumn{3}{c}{LSUN} & & \\\cmidrule{2-4}
Parameter\textbackslash Experiment Figure & Fig. \ref{sample2} & Fig. \ref{ternaryperformanceLSUN} & Fig. \ref{LSUNsamples} \\ \midrule
Num. of Generator Iter. (Offline Training) & 10000 & 10000 & 10000 \\
Num. of Generator Iter. (fast adaptation) & 10000 & 10000 & 5000\\
Batch Size & 64 & 64 & 64 \\
Optimizer Parameters & \multicolumn{4}{c}{Adam(\(\beta_1=0.5, \beta_2=0.999\), l.r.\(=0.0001\))} \\
Num. of Discriminator Iter. per Gen. Iter.& 5 & 5 & 5 \\
Gradient-Penalty Factor & 10 & 10 & 10 \\
Num. of Samples in Node \(0\) & 1000 & 1000 & Varies \\
\(\{\lambda_k\}\) & Equal & Equal & Equal \\
Number of Training Samples  & 100000 & 100000 & 100000 \\
Avg. Dur. of Pretraining/Transfer per Iter.  & \multicolumn{3}{c}{1.055 seconds} &  &  \\
Avg. Dur. of Training Barycenter per Iter.  & \multicolumn{3}{c}{2.125 seconds} &  &  \\
\midrule 
&\multicolumn{4}{c}{MNIST} & & \\\cmidrule{2-5}
Parameter\textbackslash Experiment & Fig. \ref{exp4_1}, \ref{exp4} & Fig. \ref{sample4} & Fig. \ref{exp6} & Fig. \ref{trvsbaMNIST} \\ \midrule
Num. of Generator Iter. (Offline Training) & 1000 & 3000 & 1000 & 1000\\
Num. of Generator Iter. (fast adaptation) & 800 & 1000 & 800 & 800\\
Batch Size & 256 & 256 & 256 & 256 \\
Optimizer Parameters & \multicolumn{4}{c}{Adam(\(\beta_1=0.5, \beta_2=0.999\), l.r.\(=0.0001\))} & &\\
Num. of Discriminator Iter. per Gen. Iter.& 5 & 5 & 5 & 5 \\
Gradient-Penalty Factor & 10 & 10 & 10 & 10 \\
Num. of Samples in Node \(0\) & Varies & 8 & 100 & 20\\
\(\{\lambda_k\}\) & Equal & Equal & Equal & Equal \\
Number of Training Samples  & 60000 & 60000 & 60000 & 60000 \\
Avg. Dur. of Pretraining/Transfer per Iter.  & \multicolumn{4}{c}{0.535 seconds} &  &  \\
Avg. Dur. of Training Barycenter per Iter.  & \multicolumn{4}{c}{1.020 seconds} &  &  \\
\bottomrule
\end{tabular}
\end{table*}

\begin{table*}\centering
\ra{1.3}
\caption{List of hyper-parameters used in all experiments.}\label{parameters2}
\begin{tabular}{@{}llllllll@{}}\toprule
&\multicolumn{4}{c}{CIFAR100} & & \\\cmidrule{2-5}
Parameter\textbackslash Experiment Figure & Fig. \ref{sample3} & Fig. \ref{CIFAR100Figure2} & Fig. \ref{CIFAR100Figure3} & Fig. \ref{CIFAR100Figure4} \\ \midrule
Num. of Generator Iter. (Offline Training) & 10000 & 10000 & 10000 & 10000 \\
Num. of Generator Iter. (fast adaptation) & 10000 & 10000 & 10000 & 10000\\
Batch Size & 64 & 64 & 64 & 64\\
Optimizer Parameters & \multicolumn{4}{c}{Adam(\(\beta_1=0.5, \beta_2=0.999\), l.r.\(=0.0001\))} & &\\
Num. of Discriminator Iter. per Gen. Iter.& 5 & 5 & 5 & 5\\
Gradient-Penalty Factor & 10 & 10 & 10 & 10 \\
Num. of Samples in Node \(0\) & 1000 & Varies & 500 & 500 \\
\(\{\lambda_k\}\) & Equal & Equal & Equal & Varies \\
Number of Training Samples  & 50000 & 50000 & 50000 & 50000\\
Avg. Dur. of Pretraining/Transfer per Iter.  & \multicolumn{4}{c}{0.756 seconds} &  &  \\
Avg. Dur. of Training Barycenter per Iter.  & \multicolumn{4}{c}{1.482 seconds} &  &  \\
\midrule 
&\multicolumn{3}{c}{CIFAR100}\\\cmidrule{2-5}
Parameter\textbackslash Experiment Figure & Fig. \ref{CIFAR100Figure5} & Fig. \ref{CIFAR100Figure6} & Fig. \ref{CIFAR100Figure7} & Fig. \ref{samp2} \\\midrule
Num. of Generator Iter. (Offline Training) & 5000 & 5000 & 10000 & 10000\\
Num. of Generator Iter. (fast adaptation) & 5000 & 5000 & 10000 & 7000\\
Batch Size & 64 & 64 & 64 & 64 \\
Optimizer Parameters & \multicolumn{4}{c}{Adam(\(\beta_1=0.5, \beta_2=0.999\), l.r.\(=0.0001\))}\\
Num. of Discriminator Iter. per Gen. Iter.& 5 & 5 & 5 & 5\\
Gradient-Penalty Factor & 10 & 10 & 10 & 10\\
Num. of Samples in Node \(0\) & 1000 & 1000 & 5000 & 400\\
\(\{\lambda_k\}\) & Equal & Equal & Equal & Equal\\
Number of Training Samples  & 50000 & 50000 & 50000 & 50000\\
Avg. Dur. of Pretraining/Transfer per Iter.  & \multicolumn{4}{c}{0.756 seconds}\\
Avg. Dur. of Training Barycenter per Iter.  & \multicolumn{4}{c}{1.482 seconds}\\
\bottomrule
\end{tabular}
\end{table*}
\begin{figure}[t]
	\begin{center}
		\centerline{\includegraphics[width=0.85\columnwidth]{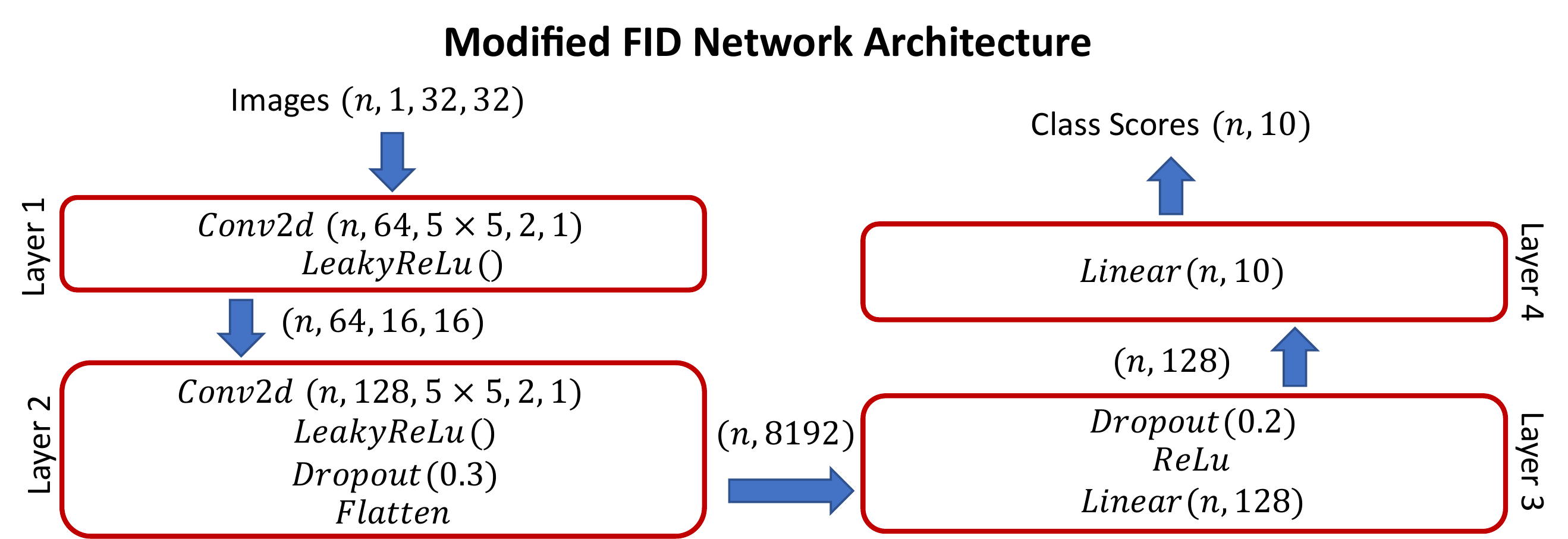}}%
		\caption{The DNN architecture used for extracting features in MNIST images and computing the modified FID scores. }\label{modifiednetwork}
	\end{center}
	\vskip -0.2in
\end{figure}
\begin{figure}[t]
		\hspace{0\columnwidth}
		\subfigure[Evolution of image quality on MNIST using FID score under different number of samples at target edge node.]{\includegraphics[width=0.47\columnwidth]{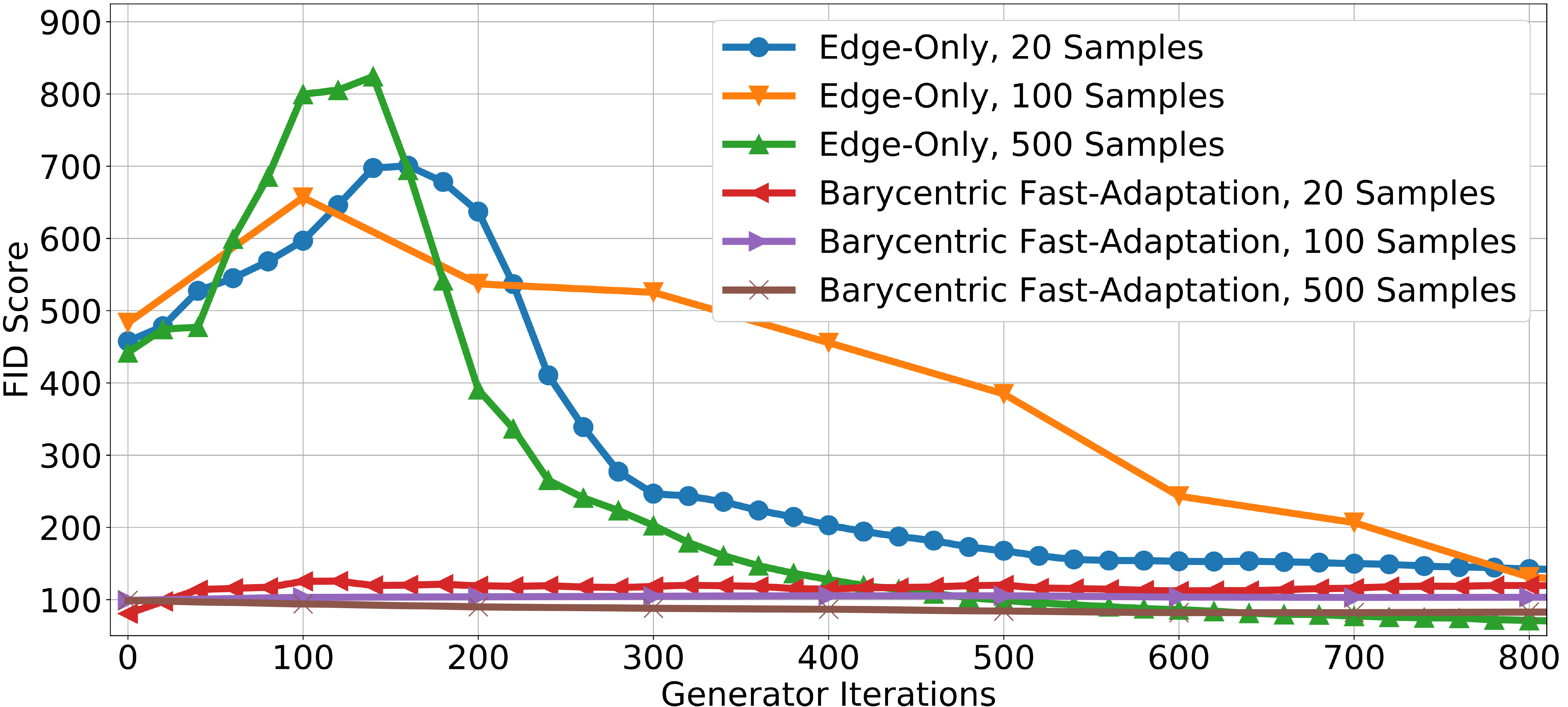}\label{exp4_1}}%
		\hspace{0.01\columnwidth}
		\subfigure[Evolution of image quality on MNIST using modified FID score under different number of samples at target edge node.]{\includegraphics[width=0.49\columnwidth]{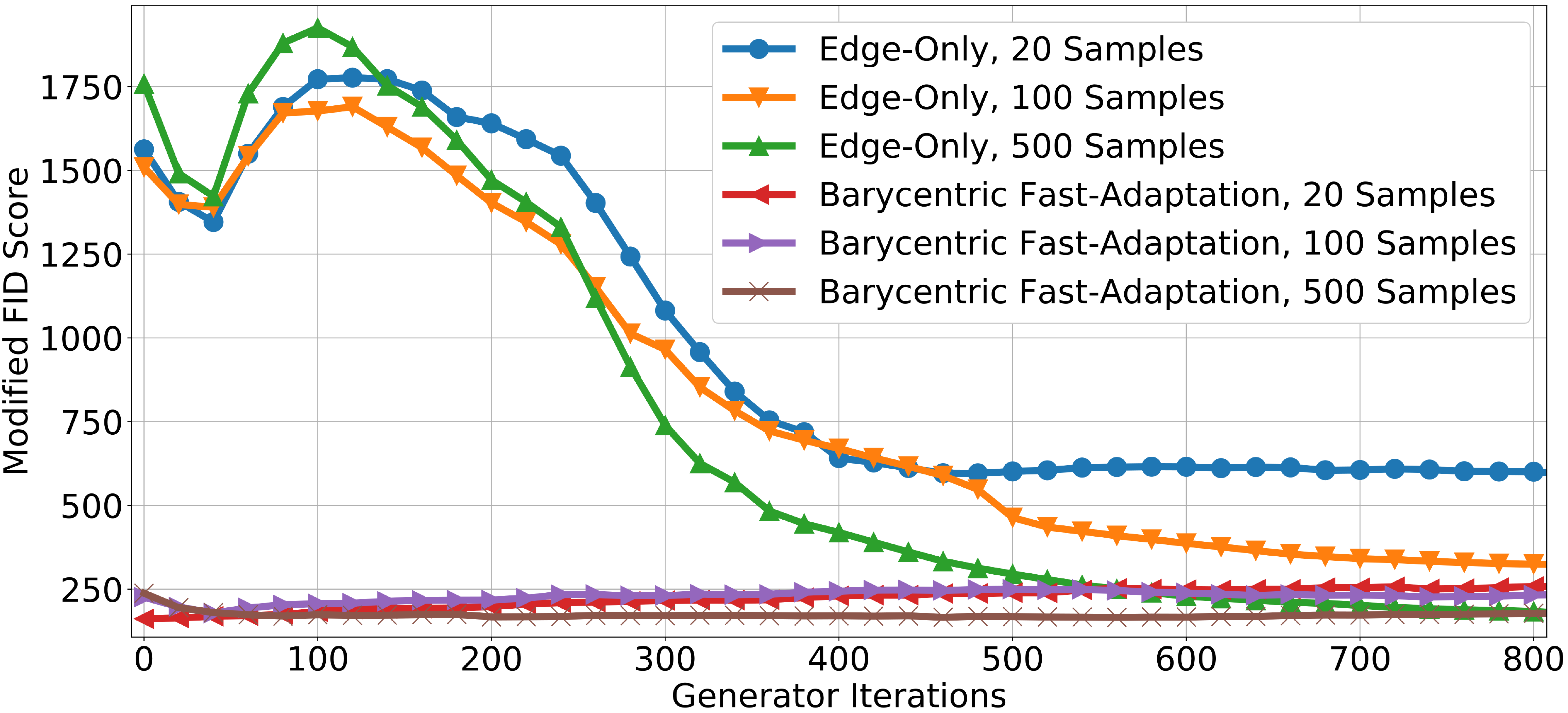} \label{exp4}}%
		\caption{Image quality performance of two stage adaptive coalescence algorithm in various scenarios.}
\end{figure}

\section{Experiments and Further Discussion}\label{appC}

\subsection{Frechet Inception Distance Score}
\textbf{An overview of FID score.}
Quantifying the quality of images is an important problem for performance comparison in the  literature on GANs. A variety of metrics have been proposed in the literature to  quantify image quality with the consideration of over-fitting and mode collapse issues. This study adopts the FID score \cite{fid}, which has been shown to be able to accurately evaluate image quality and over-fitting, independent of the number of classes. Since most of the datasets considered in this study (CIFAR10, LSUN and MNIST) only contain \(10\) classes and they are further split into subsets, using a metric independent of classes is essential for our study, and the metrics highly dependent on the number of classes, e.g., Inception score (IS), may not be appropriate here.

Similar to IS, a pre-trained `Inception' network is utilized to extract useful features for obtaining the FID score, such that the features of real and fake images can then be used to compute correlations between these images so as to evaluate the quality of images. A perfect score of \(1\) can be obtained only if the features of both real and fake datasets are the same, i.e., fake images span every image in the real datasets.
Consequently, if a generative model is trained only on a subset of the real-world dataset, the model would over-fit the corresponding subset and does not capture the features of the remaining real samples, thus yielding a bad FID score.

\noindent\textbf{Modified FID score for MNIST dataset.} Since the `Inception' network is pre-trained on `ILSVRC 2012' dataset, both IS and FID scores are most suitable for RGB images (e.g., CIFAR), which however cannot accurately capture the valuable features in MNIST images, simply because `ILSVRC 2012' dataset does not contain MNIST classes.

To resolve these issues, we particularly train a new neural network to extract useful features for MNIST dataset. The network architecture of the corresponding DNN is shown in Figure \ref{modifiednetwork}. Fully trained network achieves an accuracy rate of \(99.23\%\) for classifying the images in MNIST. Though the corresponding architecture is much simpler in comparison to the `Inception' network, the high classification accuracy indicates that the network can extract the most valuable features in MNIST dataset.

To further demonstrate the difference between FID and modified FID scores, we evaluate the results of Experiment \(4\) using both approaches, as shown in Figure \ref{exp4_1} and \ref{exp4}, respectively. 
It can be seen that upon convergence, the FID scores for the `Edge-Only' with different number of samples are similar, whereas the modified FID scores under different cases are more distinct from each other and correctly reflect the learning performance. 
Besides, `Edge-Only' with \(20\) samples incorrectly performs better than `Edge-Only' with \(100\) samples in the FID score, while `Edge-Only' with \(20\) and \(100\) samples perform as expected with the modified FID score. 
Hence, the modified FID score can better capture the image features compared with the FID score, 
and is a more suitable metric to evaluate the image quality in experiments with MNIST.
\begin{figure}[t]
		\hspace{0\columnwidth}
		\subfigure[Evolution of image quality on LSUN using FID score under different number of samples at target edge node.]{\includegraphics[width=0.47\columnwidth]{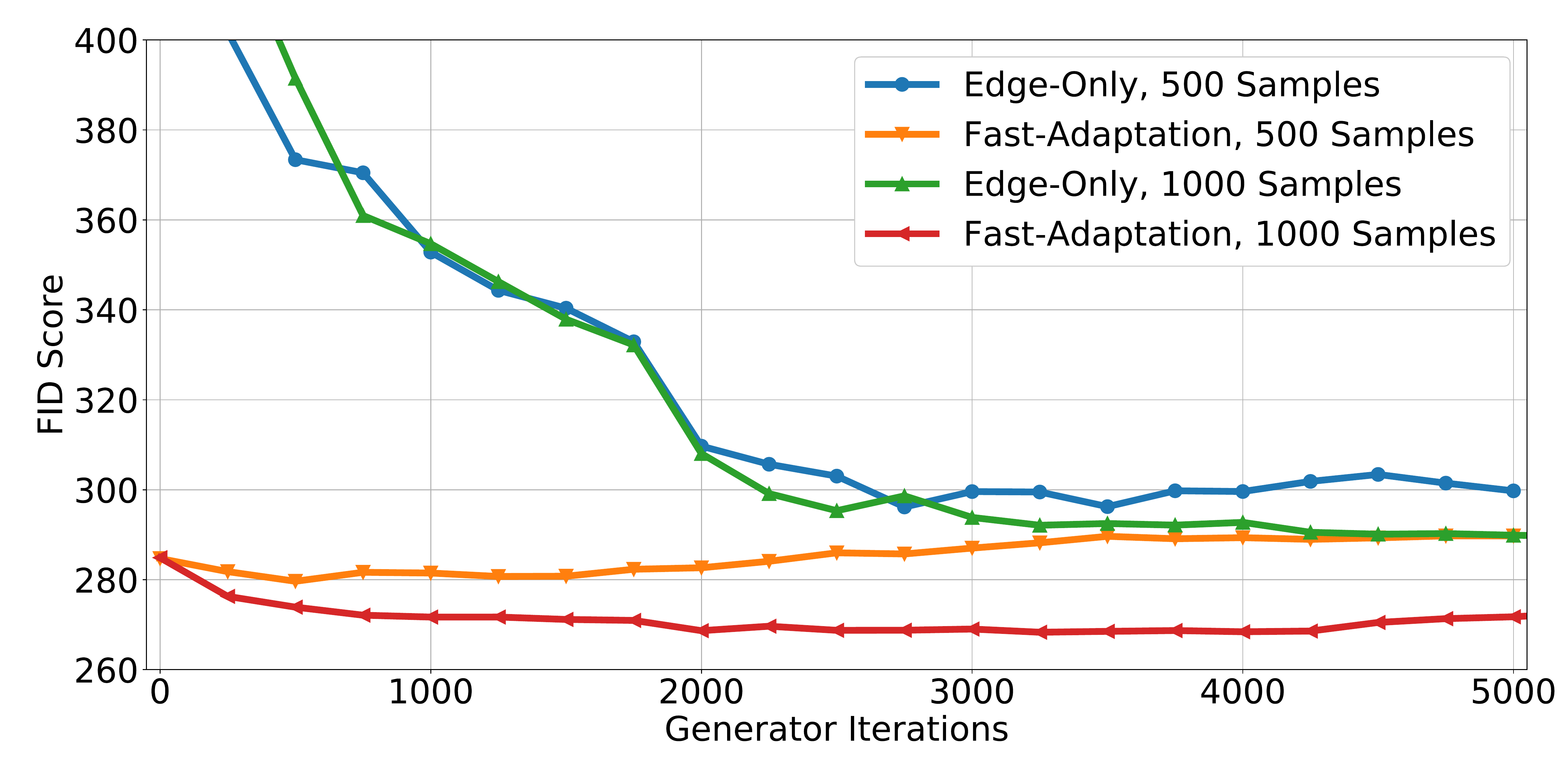}\label{LSUNsamples}}%
		\hspace{0.01\columnwidth}
		\subfigure[Evolution of image quality on CIFAR100 using FID score under different number of samples at target edge node.]{\includegraphics[width=0.49\columnwidth]{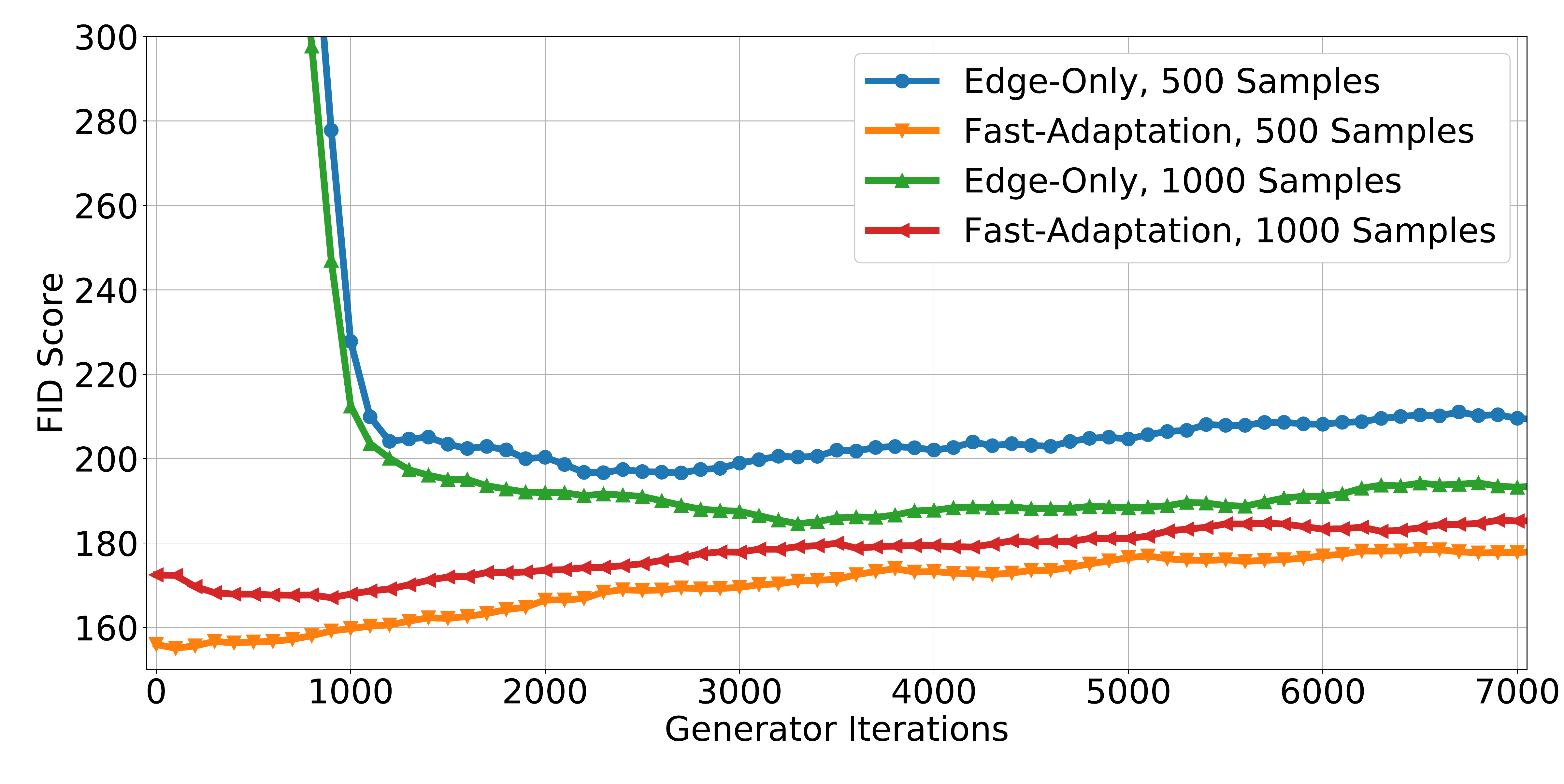} \label{CIFAR100Figure2}}%
		\hspace{0\columnwidth}
		\subfigure[Convergence of Barycentric fast-adaptation compared to 3 different baselines: Case for LSUN.]{\includegraphics[width=0.49\columnwidth]{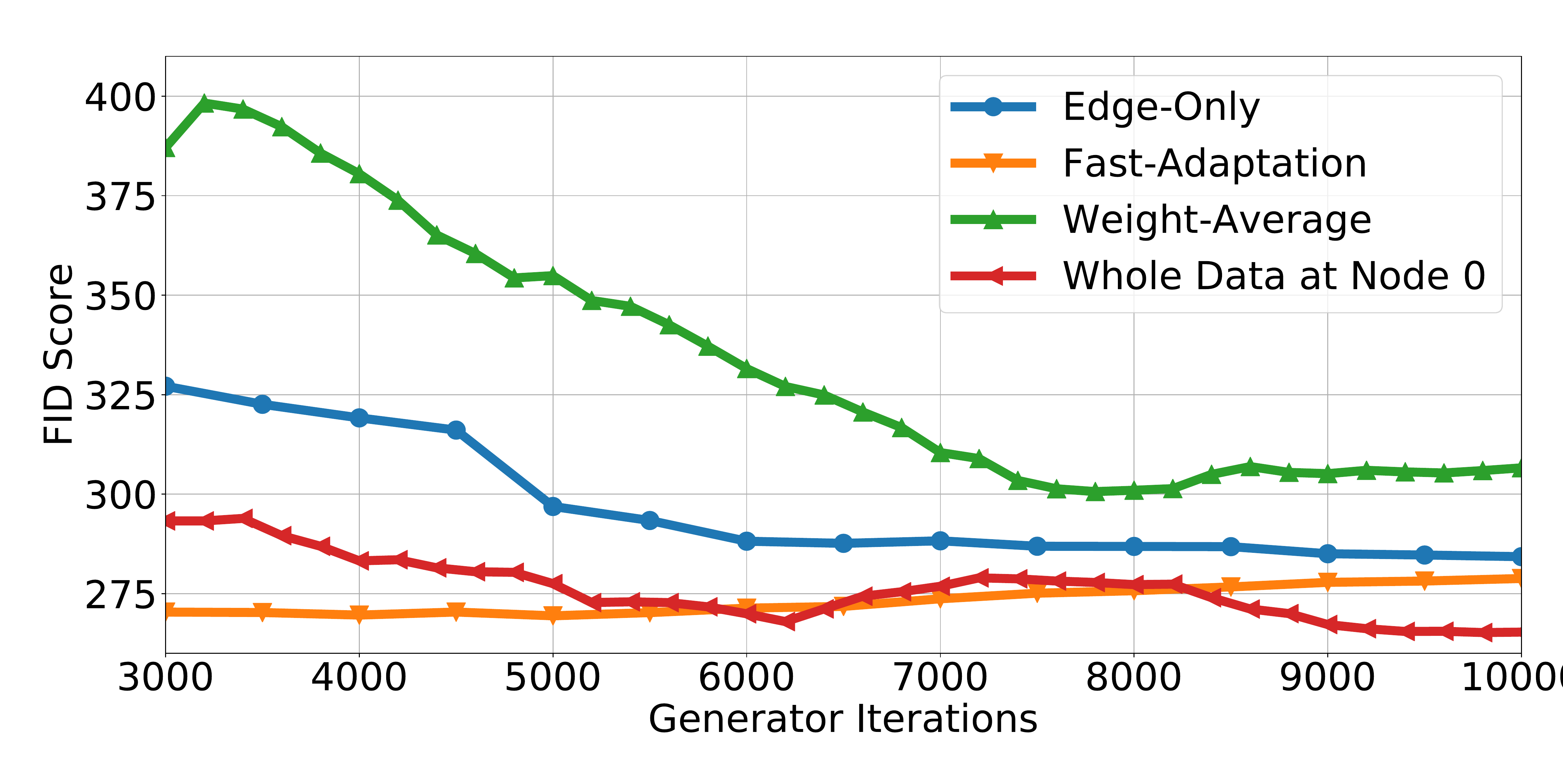}\label{sample2}}%
		\hspace{0.01\columnwidth}
		\subfigure[Convergence of Barycentric fast-adaptation compared to 3 different baselines: Case for CIFAR100. ]{\includegraphics[width=0.49\columnwidth]{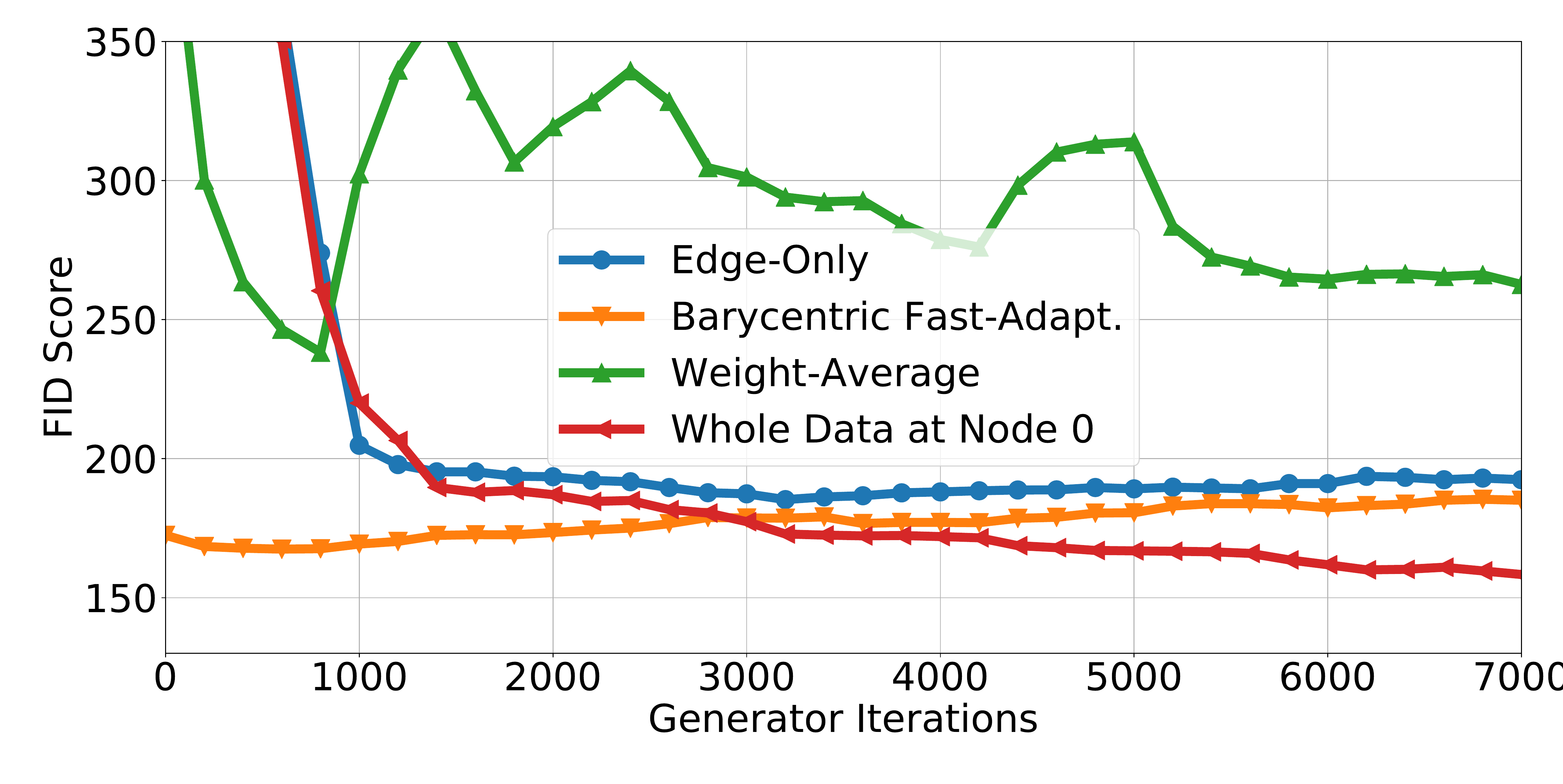} \label{sample3}}%
		\hspace{0\columnwidth}
		\caption{Image quality performance of barycentric fast adaptation in various scenarios.}\label{fig13}
\end{figure}
\subsection{Additional Experiments on MNIST, CIFAR10, CIFAR100 and LSUN}

\noindent\textbf{Fine-tuning via fast adaptation.} We investigate the convergence and the image quality of various training scenarios on MNIST, CIFAR10, CIFAR100 and LSUN datasets. To demonstrate the improvements by using the proposed framework based on \emph{Barycentric Fast-Adaptation}, we conduct extensive experiments and compare performance with \(3\) additional baselines: 1) \emph{Edge-Only}: only local dataset with few samples at the target edge node is used in WGAN training; 2) \emph{Weight-Average}: an initial model for training a WGAN model at the target edge node is computed by weight-averaging pre-trained models across other edge nodes, and then \emph{Barycentric Fast-Adaptation} is used to train a WGAN model; 3) \emph{Whole Data at Node \(0\)}: the whole dataset available across all edge nodes is used in WGAN training.

As illustrated in Figure \ref{exp4}, \ref{LSUNsamples} and \ref{CIFAR100Figure2}, \emph{barycentric fast adaptation} outperforms \emph{Edge-Only} in all scenarios with different sizes of the training set. In particular, the significant gap of modified FID scores between two approaches in the initial stages indicates that the barycenter found via offline training and adopted as the model initialization for fast adaptation, is indeed close to the underlying model at the target edge node, hence enabling faster and more accurate edge learning than \emph{Edge-Only}. Moreover, upon convergence, the \emph{barycentric fast adaptation} approach  achieves a better FID score (hence better image quality) than \emph{Edge-Only}, because the former converges to a barycenter residing between the coalesced model computed offline and the empirical model at target edge node. We further notice that \emph{barycentric fast adaptation} noticeably addresses catastrophic forgetting problem apparent in \emph{Transferring GANs} and \emph{Edge-Only}, but cannot eliminate it completely in Figure \ref{fig13}. As it will be illustrated in Figure \ref{fig15}, catastrophic forgetting can be eliminated by selecting appropriate \(\eta_k\) values. As expected, the modified FID score gap between two approaches decreases as the number of data samples at the target node increases, simply because the empirical distribution becomes more `accurate'.

\begin{figure}[t]
		\hspace{0\columnwidth}
		\subfigure[Convergence of ternarized and full precision barycentric fast adaptation methods on CIFAR100.]{\includegraphics[width=0.47\columnwidth]{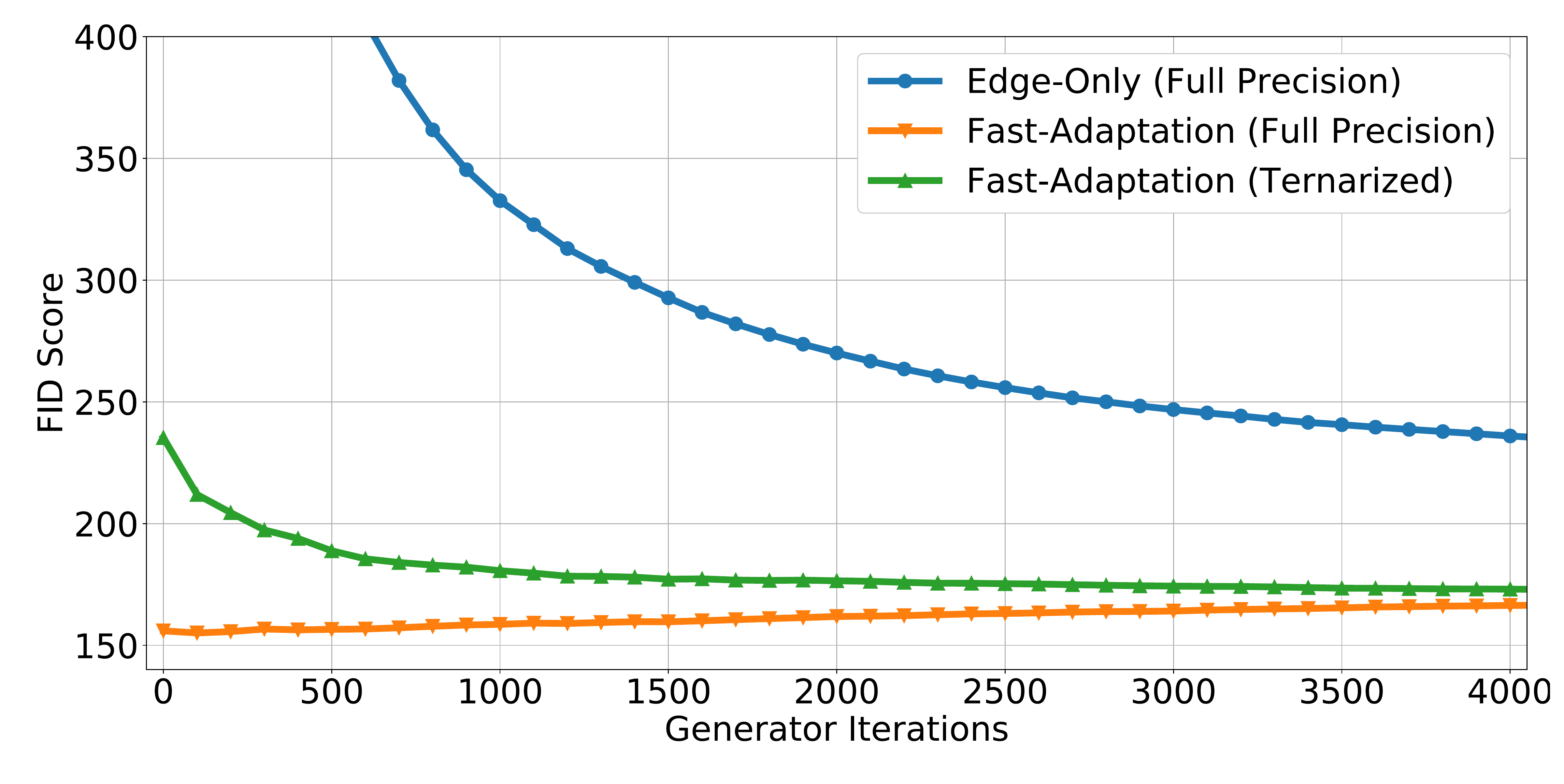}\label{CIFAR100Figure3}}%
		\hspace{0\columnwidth}
		\subfigure[Convergence of image quality of ternarized and full precision barycentric fast adaptation techniques on MNIST.]{\includegraphics[width=0.5\columnwidth]{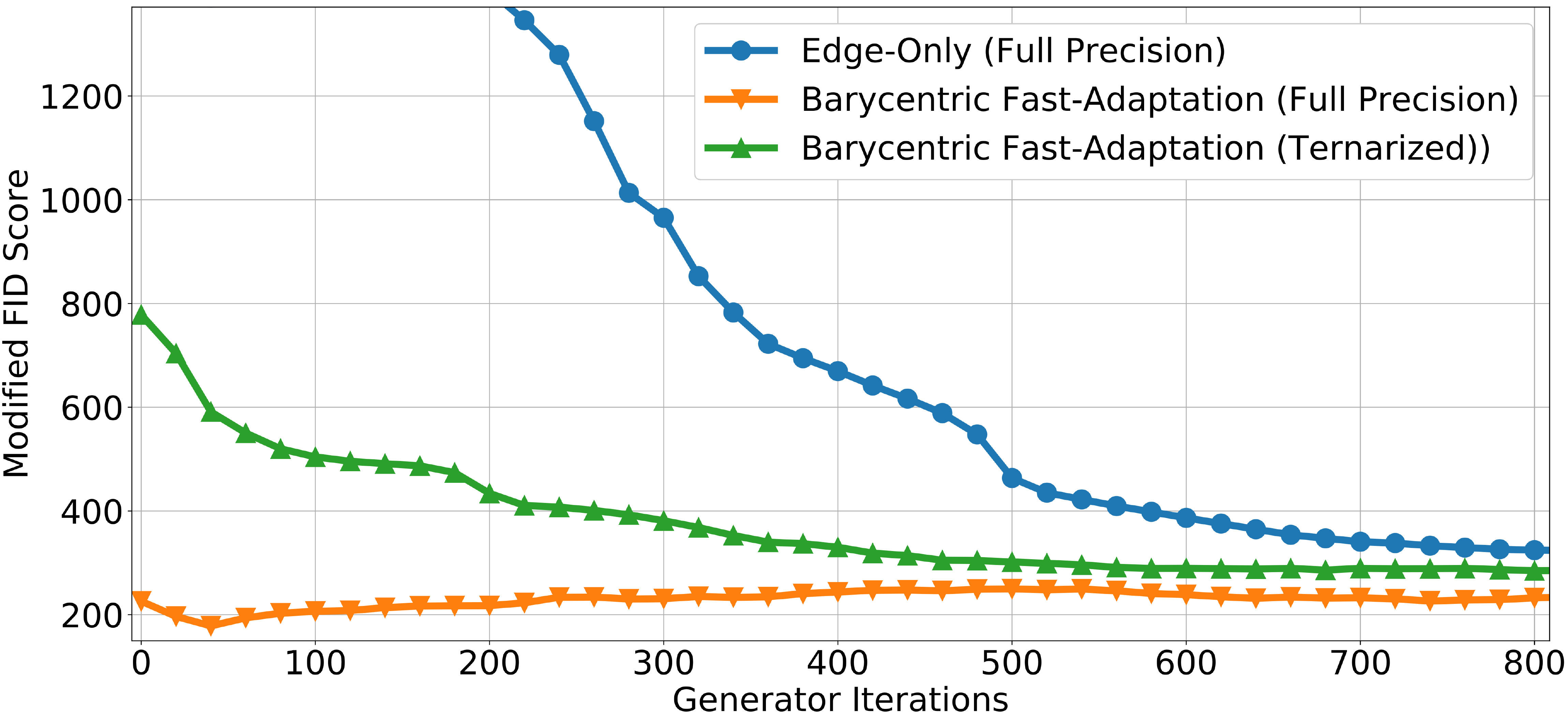} \label{exp6}}%
		\hspace{0.01\columnwidth}\\
		\hspace*{0.25\columnwidth}
		\subfigure[Convergence of ternarized and full precision fast adaptation methods on CIFAR10. ]{\includegraphics[width=0.5\columnwidth]{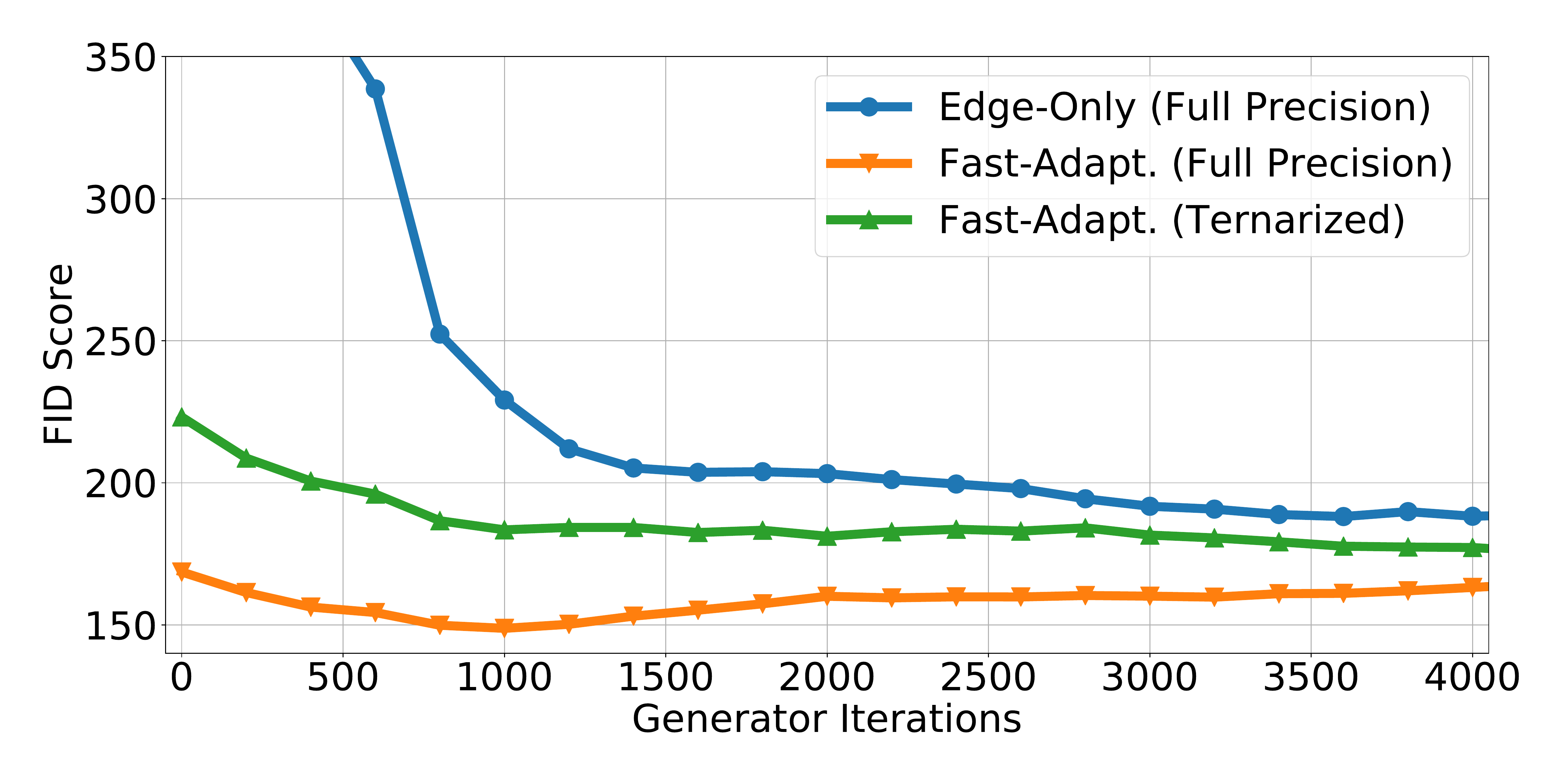}\label{ternaryperformance}}%
		\hspace{0.01\columnwidth}
		\caption{Image quality performance of two stage adaptive coalescence algorithm in various scenarios.}
\end{figure}

Figures \ref{sample2} and \ref{sample3} compare the performance of \emph{Barycentric Fast-Adaptation} on LSUN and CIFAR100 with additional 2 baselines \emph{Weight-Average} and \emph{Whole Data at Node \(0\)}. Again, \emph{Barycentric Fast-Adaptation} outperforms all baselines in the initial stages of training, but as expected,  \emph{Whole Data at Node \(0\)} achieves the best FID score upon convergence as it utilizes whole reference dataset.  Unsurprisingly, \emph{Weight-Average} performs poorly since weight averaging does not constitute a shape-preserving transformation of pre-trained models,  while \emph{Barycentric Fast-Adaptation} can by utilizing displacement interpolation in the Wasserstein space.

\begin{figure}[t]
        \hspace{0\columnwidth}
		\subfigure[Evolution of image quality on CIFAR100 for different Wasserstein ball radii values. ]{\includegraphics[width=0.49\columnwidth]{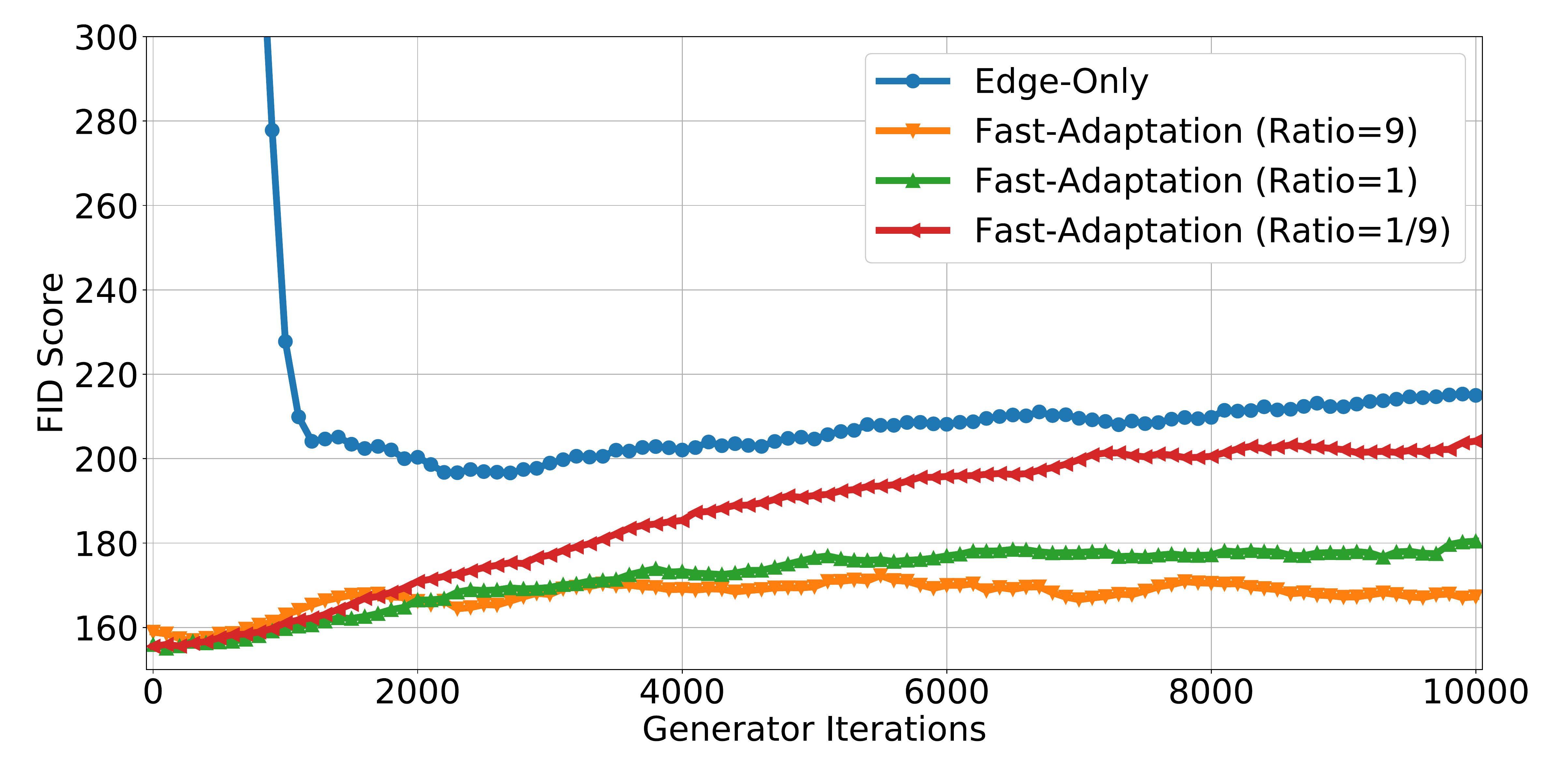} \label{CIFAR100Figure4}}%
		\hspace{0.01\columnwidth}
		\subfigure[Evolution of the quality of images generated by fast adaptation using pre-trained model or using few samples at target node.]{\includegraphics[width=0.49\columnwidth]{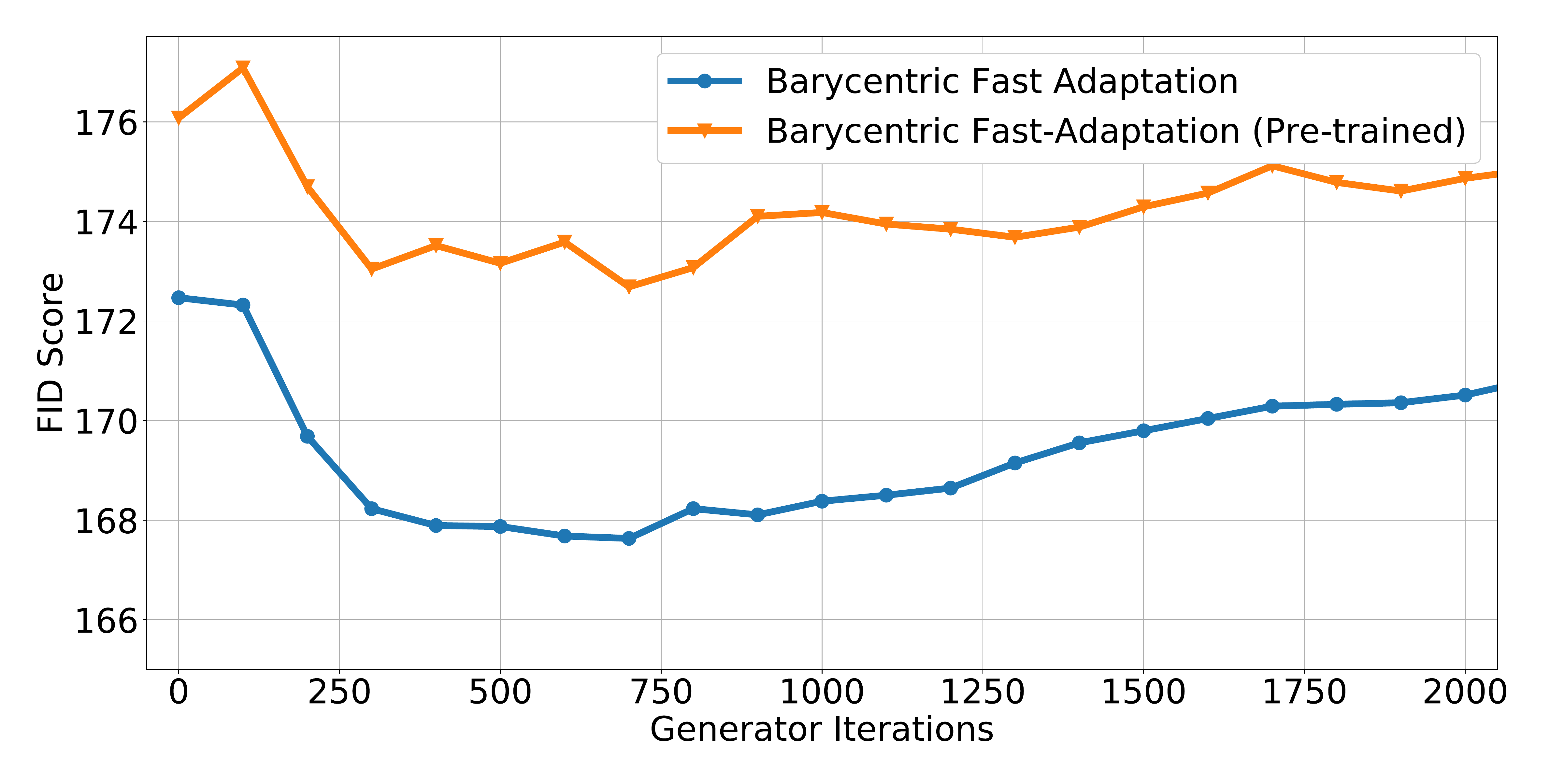} \label{CIFAR100Figure5}}%
		\hspace{0\columnwidth}
		\subfigure[Evolution of the quality of images generated by fast adaptation for different number of data samples with same data classes at edge nodes.]{\includegraphics[width=0.48\columnwidth]{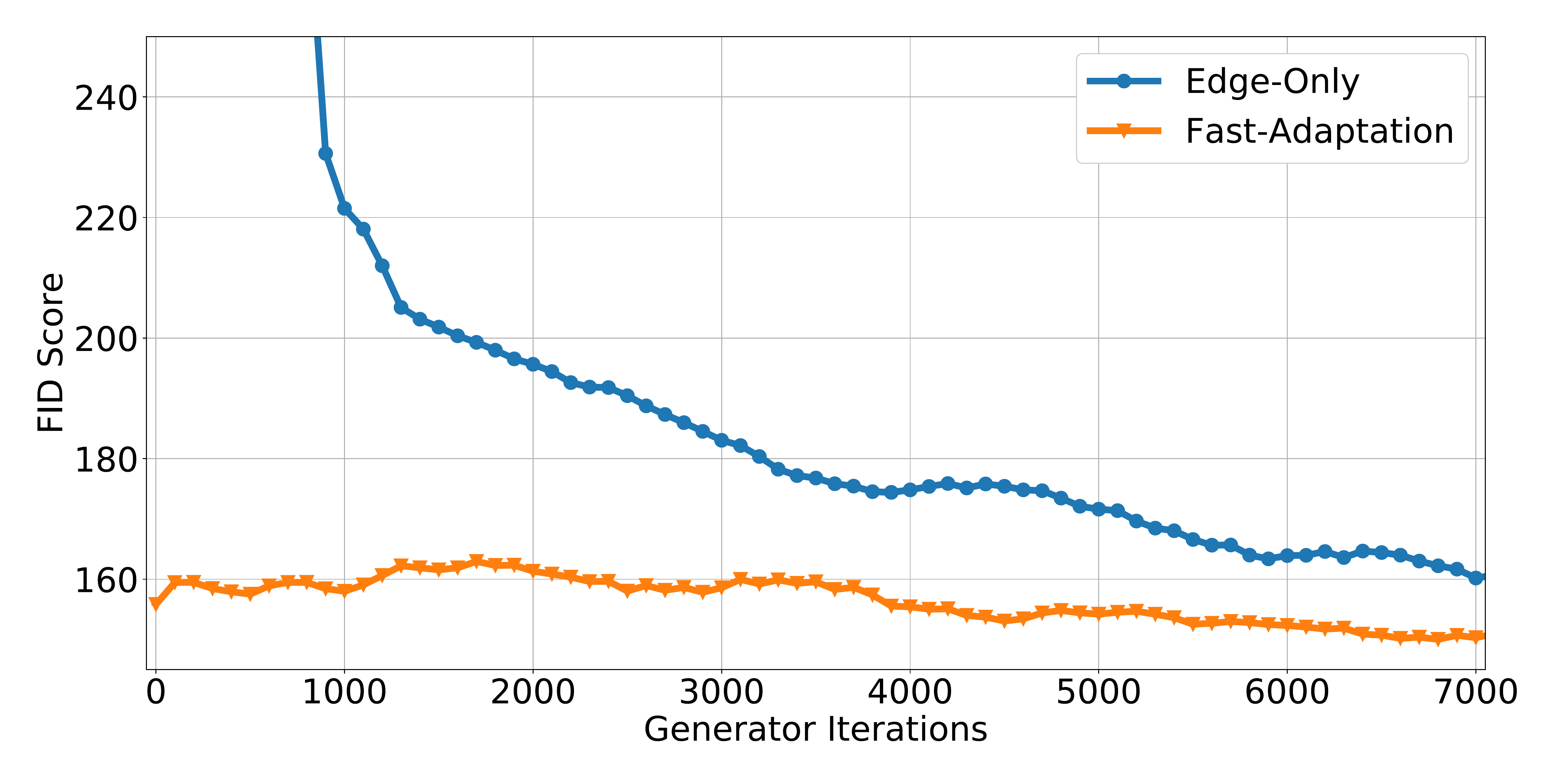} \label{CIFAR100Figure7}}%
		\hspace{0.01\columnwidth}
		\subfigure[Evolution of the quality of images generated by fast adaptation for disjoint dataset at target node.]{\includegraphics[width=0.49\columnwidth]{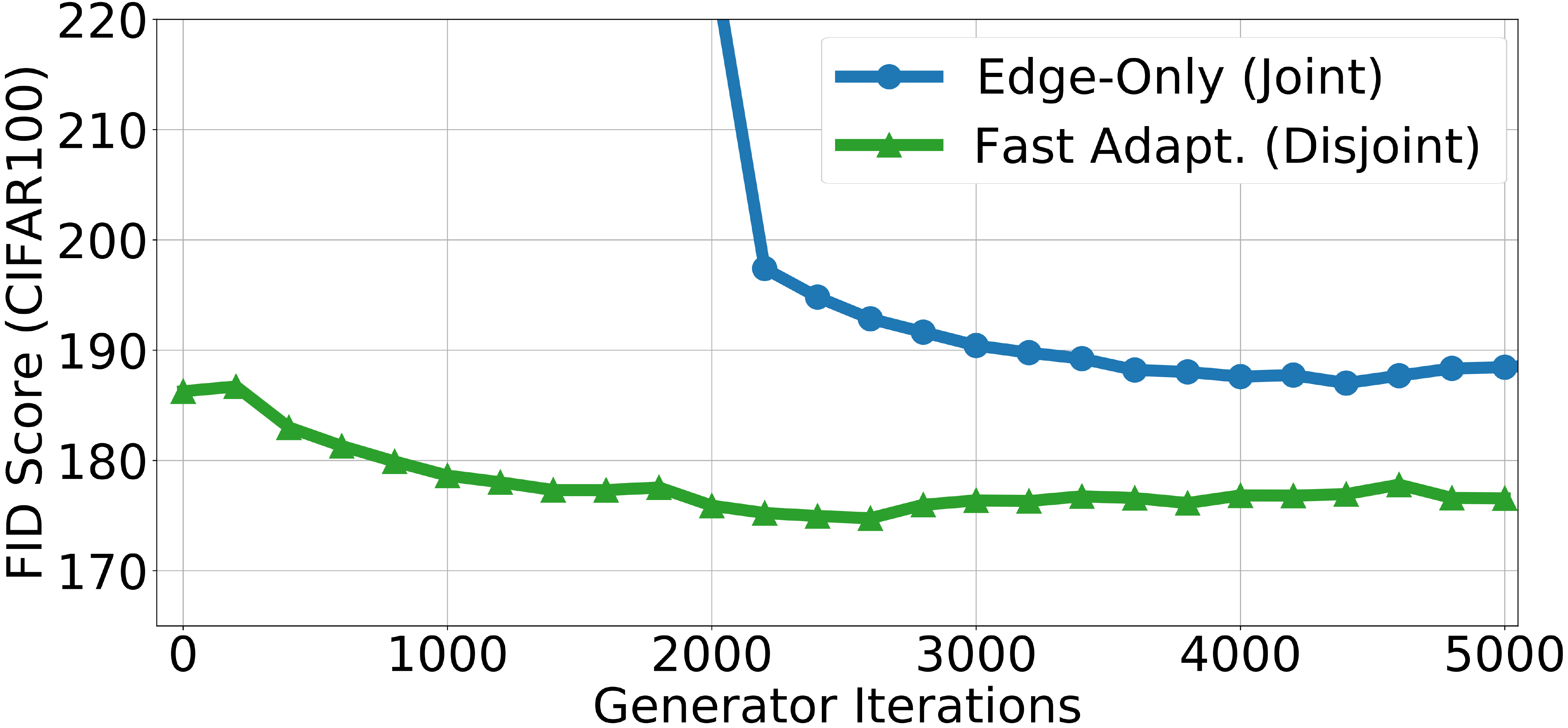}\label{CIFAR100Figure6}}%
		\hspace{0\columnwidth}
		\caption{Image quality performance of two stage adaptive coalescence algorithm in various scenarios. CIFAR100 dataset is used in all experiments.}\label{fig15}
\end{figure}

\noindent\textbf{Ternary WGAN based fast adaptation.} Following the same spirit of the experiment for LSUN, we compare the image quality obtained by ternary WGAN-based fast adaptation against both full precision counterpart and \emph{Edge-Only} for CIFAR100, CIFAR10 and MNIST datasets. It can be seen from the modified FID scores (Figure \ref{exp6}, \ref{ternaryperformance} and \ref{CIFAR100Figure3}) that the ternary WGAN-based fast adaptation facilitates image quality in between its full precision counterpart and the \emph{Edge-Only} approach, which indicates that the ternary WGAN-based fast adaptation provides negligible performance degradation compared to the full precision method.

\subsection{Additional Experiment Settings}
This subsection features additional experiment setups, which are not considered as primary use cases for the proposed \emph{Barycentric fast adaptation}, but might provide useful insights regarding the algorithm.

\noindent\textbf{The impact of Wasserstein ball radii.} To demonstrate the impact of the Wasserstein ball radii, we design an experiment with different radius values in the fast adaptation stage. The CIFAR100 dataset is equally split to 2 edge nodes and an offline barycenter is computed with equal Wasserstein ball radii. We trained 3 different models for fast adaptation with varying weights \( \lambda_k =\nicefrac{1}{\eta_k}\) . As noted in Section 1, 
radius \(\eta_k\) represents the relevance (hence utility) of the knowledge transfer, and the smaller it is, the more informative the corresponding Wasserstein ball is. 
As illustrated in Figure \ref{CIFAR100Figure4}, the performance of \emph{barycentric fast adaptation} improves as the weight \( \lambda_k \) increases, because the knowledge transfer from the offline barycenter is more informative. Consequently, the fast adaptation benefits from the coalesced model more, which mitigates the effects of catastrophic forgetting, leading to better image quality. 


\noindent\textbf{Pre-training WGAN at target edge node.} In this experiment, we explore the possible effects of using a pre-trained WGAN model, which is trained using the local samples at the target edge node, instead of using the samples at target edge node as in the proposed barycentric fast adaptation phase. Specifically, the CIFAR100 dataset is split into 2 equal size subsets and each subset is placed on one of two edge nodes, based on which an offline barycenter model is trained. In addition, another WGAN model is pre-trained using local samples  at the target edge node as in \emph{Edge-Only}. Subsequently, model fusion is applied using the  offline barycenter model and the pre-trained WGAN model at the target edge node. Figure \ref{CIFAR100Figure5} demonstrates that the performance of this approach is negatively impacted, when compared to
the proposed barycentric fast adaptation.

\noindent\textbf{Disjoint classes at the target edge node.} In this experiment, we investigate the performance degradation of fast adaptation when  the datasets in the source edge nodes and at the target edge node do not have data samples from the same class. To this end, two disjoint subsets from CIFAR100, \(50\) classes and \(40\) classes, are placed on 2 edge nodes, from which an offline barycenter is trained. A subset of samples from the remaining \(10\) classes are placed on the target edge node. Figure \ref{CIFAR100Figure6} shows the performance benefit of \emph{barycentric fast adaptation} compared to \emph{Edge-Only}. As expected, \emph{barycentric fast adaptation} with disjoint classes yield less knowledge transfer from offline training to fast adaptation (yet they still share common features), but perform better than its \emph{Edge-Only} counterpart.

\noindent\textbf{The impact of sample sizes.} Next, we explore if the offline barycenter model offers any benefit to fast adaptation when all the edge nodes possess the same dataset classes, but with different sample sizes. For this purpose, 250, 200 and 50 disjoint samples are sampled from each class in CIFAR100 and placed at two edge nodes and target node, respectively. We here notice that the offline barycenter is now just a barycenter of two close empirical distributions, which share the same underlying distributions. Therefore, this setup is more suitable to transfer learning rather than edge learning. Nonetheless, \emph{barycentric fast adaptation} utilizes the additional samples from offline training, in the same spirit to \emph{transfer learning} and improves FID score in comparison to \emph{Edge-Only}, which only has access to 5000 samples (Figure \ref{CIFAR100Figure7}).

\subsection{Additional Synthetic Images}

\begin{figure*}[t]
    \begin{center}
		\includegraphics[width=1\columnwidth]{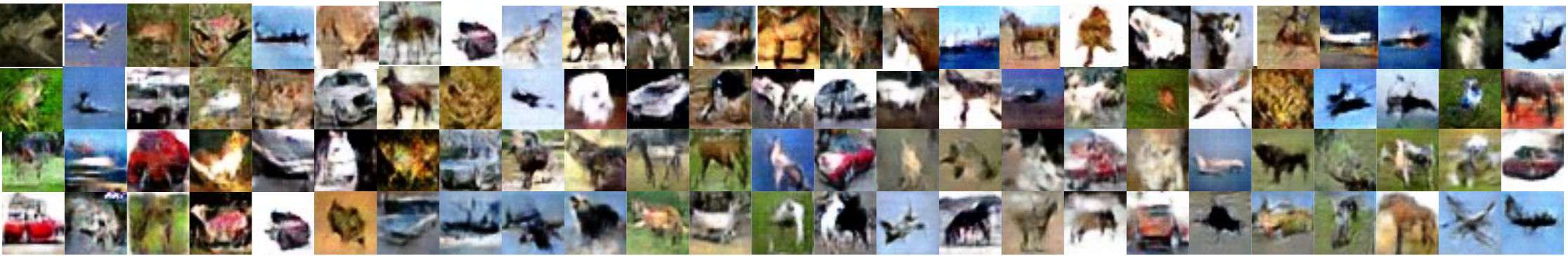}
	\end{center}
	\caption{Image samples generated at 1000th iteration using barycentric fast adaptation on CIFAR10 dataset. }\label{bary_images}
\end{figure*}

\begin{figure*}[t]
    \begin{center}
		\includegraphics[width=1\columnwidth]{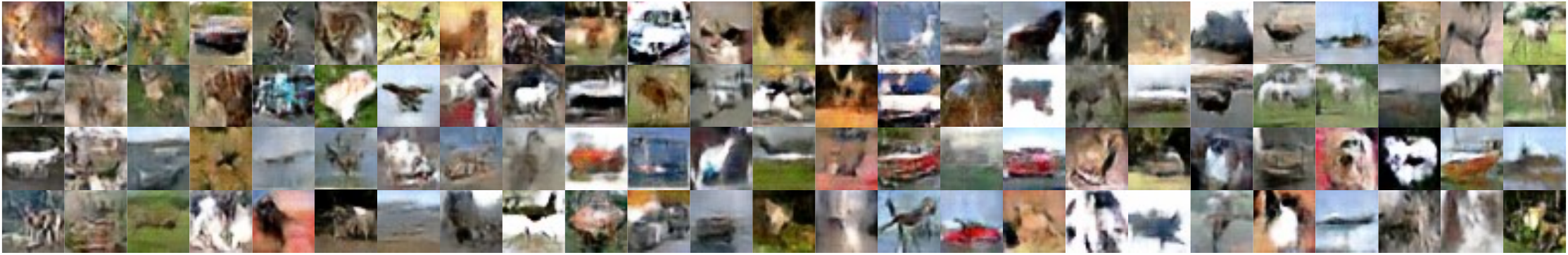}
	\end{center}
	\caption{Image samples generated at 1000th iteration using Transferring GANs on CIFAR10 dataset. }\label{transfer_images}
\end{figure*}

\begin{figure*}[t]
    \begin{center}
		\includegraphics[width=1\columnwidth]{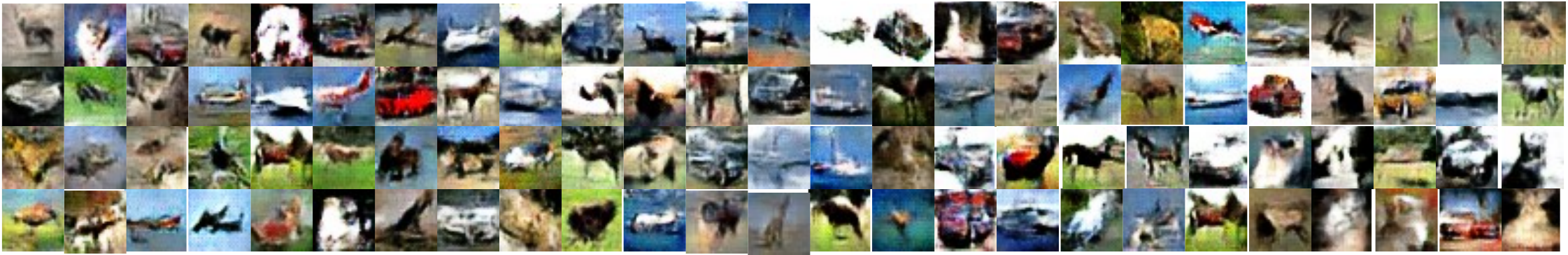}
	\end{center}
	\caption{Image samples generated at 5000th iteration using ternarized barycentric fast adaptation on CIFAR10 dataset. }\label{ternary_images}
\end{figure*}

In this section, we present more synthetic images generated using \emph{Edge-Only, transferring GANs, barycentric fast adaptation} and \emph{ternarized barycentric fast adaptation} techniques. Figure \ref{bary_images}, \ref{transfer_images} and \ref{ternary_images} illustrate 100 additional images generated by \emph{barycentric fast adaptation, transferring GANs} and \emph{ternarized barycentric fast adaptation} techniques, respectively. For barycentric fast adaptation and transferring GANs, the synthetic images are collected at iteration 1000, since both techniques attains a good FID score at early stages of training. However, transferring GANs suffers from catastrophic forgetting in latter stages of training, while barycentric fast adaptation can significantly prevent catastrophic forgetting, generating high quality synthetic images even at latter stages of training. We collected synthetic images from ternary barycentric fast adaptation at iteration 5000 since as expected it takes longer for this technique to converge to a good generative model. However, it saves significant memory in comparison to full precision barycentric fast adaptation at the expense of negligible performance degradation.

\begin{figure*}[t]
    \begin{center}
		\includegraphics[width=1\columnwidth]{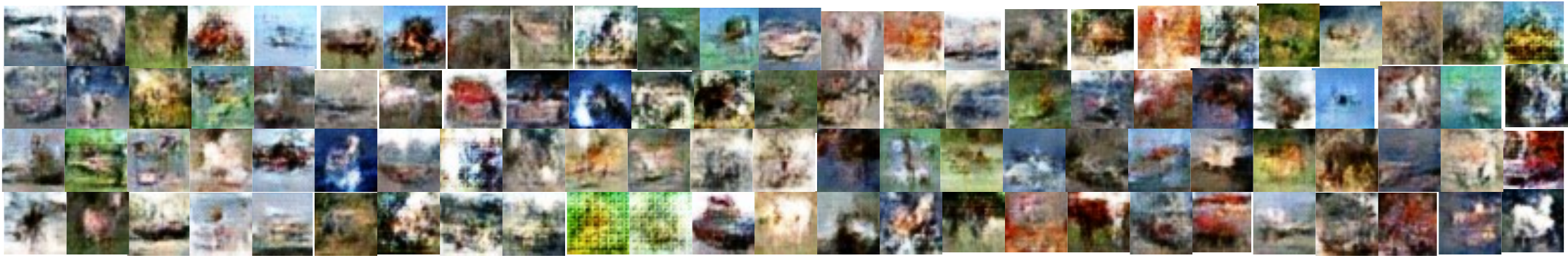}
	\end{center}
	\caption{Image samples generated at 5000th iteration using Edge-Only on CIFAR10 dataset. }\label{5000_images}
\end{figure*}

\begin{figure*}[t]
    \begin{center}
		\includegraphics[width=1\columnwidth]{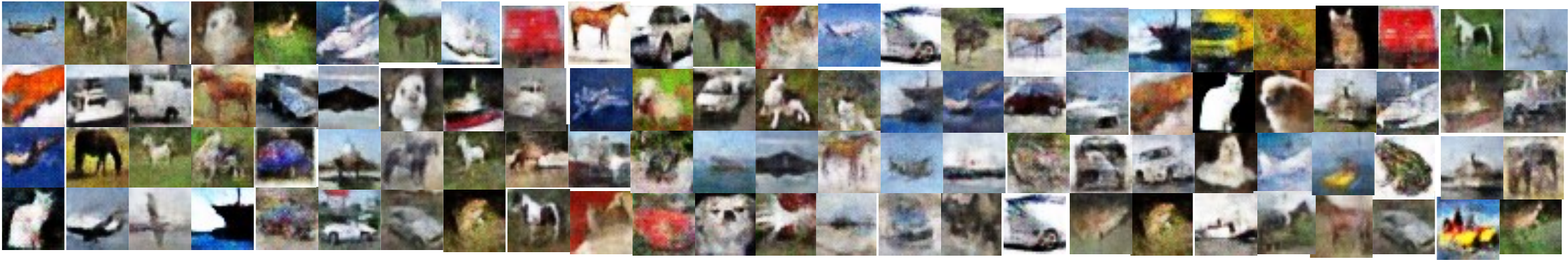}
	\end{center}
	\caption{Image samples generated at 90000th iteration using Edge-Only on CIFAR10 dataset. }\label{90000_images}
\end{figure*}

Finally, Figure \ref{5000_images} and \ref{90000_images} show images generated using Edge-Only at iterations 5000 and 90000 iterations, respectively. As it can be observed from the images in Figure \ref{5000_images}, Edge-Only has not converged to a good GAN model yet at iteration 5000.  Observe that the image quality at iteration 90000 in Figure \ref{90000_images} is significantly  better, since the Edge-Only has converged to the empirical distribution at Node 0, but it is still as not good as that generated by using barycentric fast adaptation.

\newpage

\end{document}